\pdfoutput=1
\documentclass[runningheads]{llncs}

\usepackage{eccv}

\usepackage{eccvabbrv}

\usepackage{graphicx}
\usepackage{booktabs}
\usepackage{wrapfig}
\usepackage{xcolor,colortbl}

\usepackage[accsupp]{axessibility}  %

\usepackage[pagebackref,breaklinks,colorlinks,citecolor=eccvblue]{hyperref}

\usepackage{orcidlink}
\usepackage{rotating}

\begin{document}

\title{Parameterization-driven Neural Surface Reconstruction for Object-oriented Editing in Neural Rendering} 

\titlerunning{NeuParam}

\author{Baixin Xu\inst{1}\orcidlink{0009-0002-8118-2725} \and
Jiangbei Hu\inst{2,3}\orcidlink{0000-0002-6774-6267} \and
Fei Hou\inst{4,5}\orcidlink{0000-0001-8226-6635} \and
Kwan-Yee Lin\inst{6}\orcidlink{0000-0003-0175-6398} \and 
Wayne Wu\inst{6}\orcidlink{0000-0002-1364-8151} \and
Chen Qian\inst{7}\orcidlink{0000-0002-8761-5563} \and
Ying He\inst{1,2}\orcidlink{0000-0002-6749-4485} }

\authorrunning{B.~Xu et al.}

\institute{$^{1}$S-Lab, Nanyang Technological University,
$^{2}$College of Computing and Data Science, Nanyang Technological University,
$^{3}$Dalian University of Technology,
$^{4}$Key Laboratory of System Software (Chinese Academy of Sciences) and State Key Laboratory of Computer Science, Institute of Software, Chinese Academy of Sciences,
$^{5}$University of Chinese Academy of Sciences,
$^{6}$Shanghai AI Lab,
$^{7}$SenseTime Research}

\maketitle
\renewcommand{\thefootnote}{} 
\footnotetext{Corresponding author: Ying He (email: yhe@ntu.edu.sg).}

\begin{abstract}
The advancements in neural rendering have increased the need for techniques that enable intuitive editing of 3D objects represented as neural implicit surfaces. This paper introduces a novel neural algorithm for parameterizing neural implicit surfaces to simple parametric domains like spheres and polycubes.  Our method allows users to specify the number of cubes in the parametric domain, learning a configuration that closely resembles the target 3D object's geometry. It computes bi-directional deformation between the object and the domain using a forward mapping from the object's zero level set and an inverse deformation for backward mapping. We ensure nearly bijective mapping with a cycle loss and optimize deformation smoothness. The parameterization quality, assessed by angle and area distortions, is guaranteed using a Laplacian regularizer and an optimized learned parametric domain. Our framework integrates with existing neural rendering pipelines, using multi-view images of a single object or multiple objects of similar geometries to reconstruct 3D geometry and compute texture maps automatically, eliminating the need for any prior information. We demonstrate the method's effectiveness on images of human heads and man-made objects. 
The source code is available at \url{https://xubaixinxbx.github.io/neuparam}.
\end{abstract}

\section{Introduction}
Neural radiance fields (NeRF)~\cite{mildenhall2020nerf} have garnered remarkable success in both the computer vision and computer graphics communities, redefining benchmarks for high-quality renderings and novel view synthesis. Building upon NeRF, a variety of methods leveraging implicit neural representations \cite{oechsle2021unisurf,wang2021neus,yariv2021volume} have emerged, delivering high-fidelity 3D reconstructions. The demand for intuitive 3D object editing techniques has increased with the growing capabilities of neural rendering, especially for neural implicit surfaces. Extensive explorations have been made in editing within the Neural Rendering framework to achieve diverse visual effects. By manipulating the latent codes of the network, one can implicitly alter the shape and color of 3D objects\cite{liu2021editing, yenamandra2021i3dmm, xu2023deformable}. In the work of \cite{tojo2022recolorable, kuang2023palettenerf}, palette-based methods have been employed to facilitate transformations in the color and style of rendered scenes. However, these methods suffer from limited controllability and disallow local pixel editing. Seal3D \cite{wang2023seal3d} proposed a teacher-student training strategy to edit the interactive scene at the pixel level. Yet, this method is tricky to handle shading, and each subsequent edit required a new training iteration. By employing neural networks to formulate a differentiable surface parameterization, it can be facilitated to edit textures or shapes within the neural rendering pipeline~\cite{xiang2021neutex, ma2022neural}. Existing neural parameterization techniques still face significant challenges, such as the need for appropriate distortion constraints to ensure accurate detail reconstruction or reducing the reliance on prior information from tracked mesh and UV mapping. The final quality of edited visual effects is significantly impacted by these issues.

This paper aims to develop an intuitive and easy-to-use tool for appearance editing in neural implicit surfaces within the neural rendering framework, utilizing only 2D images as input. Towards this goal, we propose a novel method to parameterize neural implicit surfaces to parametric domains where the radiance field is represented as a texture map. Distinguished from previous work that using 2D texture mapping, our neural parameterization algorithm supports the selection of different 3D parametric domains according to geometric shapes to avoid seams and reduce distortion, such as spheres, and polycubes. Technically, our parameterization algorithm utilizes learnable parameters to fit a parametric domain first and a neural network to learn a bi-directional deformation between 3D objects and the chosen parametric domains. This involves a forward deformation that maps points from the zero level set of the neural implicit surface to the parametric domain, followed by an inverse deformation, mapping points backward. Notably, we do not require any explicit prior information in learning both deformations. We employ a cycle loss to ensure the smoothness of the bi-directional deformation, and a Laplacian regularization to effectively control angle distortion. 
With the parameterization, 3D radiance field is mapped to a simple 3D parametric domain which can be further disassembled into a 2D domain, facilitating visualization and various editing tasks. Furthermore, we decompose the 2D radiance field into two components: view-independent material and view-dependent shading, streamlining both texture and shading editing. Our neural parameterization algorithm is fully compatible with existing neural rendering pipelines, allowing 3D reconstruction from multi-view images as input and the creation of texture maps simultaneously. It allows for the immediate rendering of edited textures through volume rendering, without network re-training. Moreover, it supports co-parameterization of objects of similar geometry and enables texture transfer between them. We validate the effectiveness of our method on scenes of human heads and man-made objects.
Our contributions are summarized as follows:
\begin{itemize}
\item We present a neural parameterization framework that automatically learns a parametric domain and computes bi-directional deformation between 3D objects and their parametric domains, eliminating the need for any prior information from tracked mesh or UV mapping. Our approach utilizes Laplacian loss to minimize angle distortion and allows the user to specify the domain complexity to control area distortion.

\item We introduce a simple yet effective technique for intrinsic radiance decomposition, facilitating both view-independent texture editing and view-dependent shading editing.

\item Our neural parameterization framework is  end-to-end and fully compatible with existing neural rendering architectures, accepting multi-view images as input to reconstruct 3D geometry and generate UV maps as output. Additionally, it enables the direct rendering of modified textures using the volume rendering pipeline.

\item Our approach supports co-parameterization of multiple objects and allows for texture transfer between different objects. 
\end{itemize}

\section{Related Work}

\textbf{Parameterization.} 
 Surface parameterization~\cite{sheffer2007mesh,floater2005surface} aims at computing a bijective mapping between the 3D surface and a suitable parametric domain, usually a 2D region or a simple 3D object, such as spheres~\cite{DBLP:journals/tog/GotsmanGS03} and polycubes~\cite{DBLP:journals/tvcg/GarciaXHXP13,tarini2004polycube}. Serving as an important computational tool in computer graphics and digital geometry processing~\cite{sheffer2007mesh}, surface parameterization facilitates various applications, including texture editing~\cite{fang2004textureshop}, surface painting~\cite{DBLP:conf/si3d/SunZZYXXH13}, details transfer~\cite{biermann2002cut}, remeshing~\cite{praun2003spherical}, among others. Neural networks are now commonly used in digital geometry processing, resulting in the creation of parameterization algorithms based on deep learning techniques~\cite{groueix2018papier,williams2019deep,bednarik2020shape,guo2022complexgen}.
 Unlike classical methods that yield global and seamless parametrization, most neural parameterization methods are computing local parametrization~\cite{groueix2018papier, DBLP:journals/pami/ZhangHQZZH23, DBLP:journals/ijcv/ZhangHQCZH22}. These methods typically partition the 3D model into multiple patches and parameterize each to a 2D region, possibly followed by a post-processing step to stitch the patches together. 
 Although these methods are efficient and can handle surfaces of arbitrary geometry and topology, they frequently encounter problems including non-bijectivity, a lack of smoothness, the presence of seams and overlaps, and large distortions. There are also works on neural parametric surfaces, defined in rectangular domains~\cite{low2022minimal} and $n$-sided patches~\cite{yang2023neural}. They can model complex surface geometries with high precision; however, such representations are not applicable to neural implicit surfaces and thereby incompatible with neural rendering.

\textbf{Neural Implicit Functions.} 
Recent years have seen a surge in the successful application of neural implicit functions in 3D deep learning. Compared to explicit representations such as point clouds~\cite{qi2017pointnet}, voxels~\cite{qi2016volumetric}, and meshes~\cite{wang2018pixel2mesh}, neural implicit representations offer unique advantages including flexibility, continuity, and robustness~\cite{mescheder2019occupancy, saito2019pifu,park2019deepsdf}.  When integrated within the neural rendering pipeline, neural implicit surfaces exhibit exceptional quality in 3D reconstructions from multi-view images~\cite{li2021d2im, wang2022geometry, wang2022hf, xu2023deformable, muller2022instant, li2023neuralangelo, rosu2023permutosdf}.
However, as geometries and radiance fields are encoded as network parameters, editing them becomes complex and non-intuitive, compared to explicit representations. This paper aims at tackling this challenge by explicitly representing radiance fields as texture mapping that encompasses both view-dependent and view-independent components. 

\textbf{NeRF Editing.}
Numerous studies have focused on editing neural radiance fields, including relighting~\cite{nerv2021, li2022physically}, composition~\cite{perez2023poisson, yang2021objectnerf}, content generation~\cite{Niemeyer2020GIRAFFE}, and shape editing~\cite{yuan2023interactive,lin2022neuform}. 
Editing tools operating at the scene- or object-level, such as \cite{yang2021objectnerf, yenamandra2021i3dmm, Huang22StylizedNeRF}, utilize latent codes to stylize the appearance of an object or entire scenes. Conversely, pixel-level editing tools, such as NeuMesh~\cite{yang2022neumesh} and Seal3D~\cite{wang2023seal3d}, utilize training-based approaches to edit fine-grained details. However, NeuMesh relies on explicit meshes which lack efficiency and robustness. Although Seal3D achieves interactive editing, it still suffers from poor geometry reconstruction and cannot change the illumination scene.
Parameterization-based methods, such as NeuTex~\cite{xiang2021neutex}, ISO-UVField~\cite{DasISOUVfield} and NeP~\cite{ma2022neural}, integrate the learning of UV mapping into neural rendering, enabling intuitive editing within 2D domains. However, these methods often suffer from large distortions in the parametrization. Furthermore, they require 3D prior information from tracked meshes and UV mapping - conditions that may not be readily met in practical, real-world scenarios.
IntrinsicNeRF~\cite{Ye2023IntrinsicNeRF} enhances standard neural radiance fields by generating additional outputs, including reflectance, shading, and a residual term. Incorporating a semantic branch, it facilitates real-time scene editing. However, it often results in aggregated colors, akin to palette-based methods~\cite{tojo2022recolorable, kuang2023palettenerf}, hence cannot offer fine-grained editing results at the pixel level. We decompose the radiance field into view-dependent shadings and view-independent materials. Both components are represented as textures, enabling intuitive material modifications while preserving consistent shading.

\section{Method}
\begin{figure*}[!htbp]
    \centering
    \includegraphics[width=\textwidth]{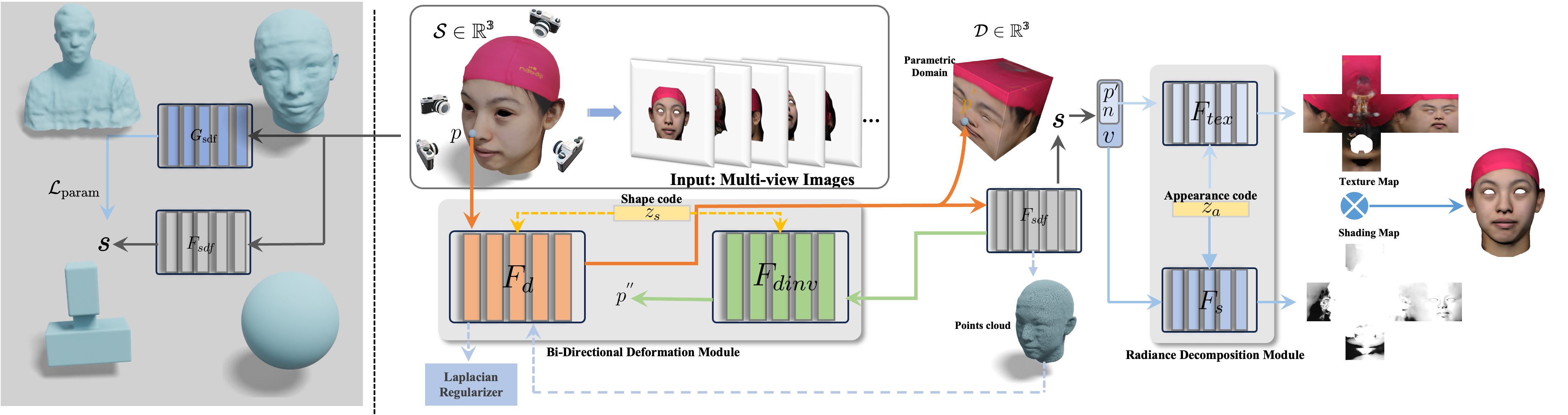}
    \caption{\textbf{Algorithmic pipeline.} Initially, we learn the parametric domain $F_\text{sdf}$ using a coarse SDF $G_\text{sdf}$ of the target surface $\mathcal{S}$ during the early phase of neural rendering (in the left grey box). Once the parametric domain $\mathcal{D}$ is determined, our pipeline subsequently utilizes the bi-directional deformation $\left(F_{\text{def}} 
    \text{~and~} F_{\text{inv-def}}\right)$, the parametric domain SDF $\left(F_{\text{sdf}}\right)$ and the radiance decomposition $\left(F_{\text{mat}} \text{~and~} F_\text{shd}\right)$. The output includes the reconstructed 3D surface $\mathcal{S}$, the decomposed radiance fields (featuring both view-dependent and view-independent components) and the map between $\mathcal{S}$ and $\mathcal{D}$.
    It is worth mentioning that while our framework involves two SDFs - one for the 3D surface $\mathcal{S}$ and the other for the parametric domain $\mathcal{D}$ - our network design only requires one of them. This is because one SDF can be derived from the other via either the forward or the backward deformation. To reduce the network complexity, we only adopt one geometry sub-network $F_{\text{sdf}}$, which is used to represent the parametric domain $\mathcal{D}$.}
    \label{fig:pipeline}
\end{figure*}

The input of our method is a collection of RGB images, denoted $\mathcal{I}=\{\mathbf{I}_i\}$, representing  \textbf{multiple objects of similar geometry or a single object}, captured from different viewpoints. To differentiate among objects, we assign a latent shape code $\mathbf{z}_s$ and an appearance code $\mathbf{z}_a$ to each object. This facilitates the co-parameterization of multiple objects and enables the transfer of material and shading among them. As illustrated in Fig.~\ref{fig:pipeline}, our method reconstructs 3D geometry as a neural implicit surface and encodes the 3D radiance field in a simple parametric domain, such as spheres and polycubes, which we learn from an initial phase in our training (Sec.~\ref{subsec: param_domain}). For each input image, we randomly select a set of pixels. Then, for each pixel, we shoot a ray originating from the location of camera and traversing through the pixel. If the ray intersects with the surface, at a point, denoted as $\mathbf{p}$, we compute its corresponding position in the parametric domain $\mathcal{D}$, denoted as $\mathbf{p}'$, via a bi-directional deformation (Sec.~\ref{subsec: bi-deform}): the forward deformation (depicted in orange) generates a displacement vector so that the updated position $\mathbf{p}'$ is in the parametric domain; subsequently, the inverse deformation (depicted in green) maps $\mathbf{p}'$ back to a point $\mathbf{p}''$, which is anticipated to be in close proximity to the surface. We incorporate a cycle loss to penalize occurrences where $\mathbf{p}''$ deviates from $\mathbf{p}$. In addition, we employ a Laplacian loss (Sec.~\ref{subsec: laplace}) to effectively reduce the angle distortion between the parametric domain and the original surface. Furthermore, we decompose the radiance into a view-independent material field and a view-dependent shading field (Sec.~\ref{subsec: app_model}), each of which can be edited independently, thus augmenting the editing capability of our framework. See Fig.~\ref{fig:pipeline} for the algorithmic pipeline. 

\subsection{Parametric Domain Learning}
\label{subsec: param_domain}
In our framework, we consider two types of parametric domains: spheres and polycubes, both represented as signed distance fields. Spheres are characterized by four learnable parameters: the center position and radius. Polycubes, a generalization of cubes, are simple structures yet offer sufficient degrees of freedom to capture the salient geometric features of target surfaces. We define a \textbf{learnable} polycube with $k$ elements as follows: For the $i$-th element, which is an axis-aligned box, its SDF $\phi_i$ is given by  
\begin{equation}
\phi_i = \max(|x-x_i|-a_i, |y-y_i|-b_i, |z-z_i|-c_i),
\end{equation}
where $(x_i, y_i, z_i)$ represents the box center, and $(a_i, b_i, c_i)$ is its dimensions. By combining the $k$ boxes, the SDF of the polycube is obtained as  
\begin{equation}
    f_\text{pc} = \max(-\phi_1, -\phi_2, ..., -\phi_k). \label{eq: polycube}
\end{equation} To ensure the differentiability of Eq.~(\ref{eq: polycube}), we approximate it using Kreisselmeier-Steinhauser (KS) function~\cite{kreisselmeier1980systematic} as $$f_\text{pc} \approx  -1/ \lambda * \log\left(\sum_i \exp(- \lambda * \phi_i)\right),$$
where $\lambda$ is a typically large value, set to 100 in our experiments. 

Since determining the configuration of boxes does not require a highly accurate geometry of the target object $\mathcal{S}$, we design the parametric domain learning as a \textbf{one-time, intermediate} step in the neural rendering pipeline. Specifically, after an initial training phase of approximately 100 epochs, the coarse geometry of the surface $\mathcal{S}$ is obtained. Subsequently, we activate the domain learning module, which takes the SDF $G_\text{sdf}$ of the coarse surface $\mathcal{S}$ and the user-specified number of boxes $k$ as input. It computes the SDF $F_\text{sdf}$ of the parametric domain $\mathcal{D}$ via minimizing the loss function: \begin{equation}
\label{eqn:paradomain}\mathcal{L}_\text{param} = \mathcal{L}_\text{sdf} + \lambda_s \mathcal{L}_\text{lapsdf},\end{equation} 
where
\begin{align}
    \mathcal{L}_\text{sdf} = \left| F_\text{sdf}(p) - G_\text{sdf}(p) \right|,
    \mathcal{L}_\text{lapsdf} = \left|\frac{\partial^2 F_\text{sdf}}{\partial x^2}  + \frac{\partial^2 F_\text{sdf}}{\partial y^2} + \frac{\partial^2 F_\text{sdf}}{\partial z^2} \right|,  
\end{align}
and the weight $\lambda_s$ is set to 0.01. The term $\mathcal{L}_\text{lapsdf}$, which is the Laplacian of the signed distance field of the polycube $\mathcal{D}$, plays a critical role in controlling its geometric complexity. 

\begin{wrapfigure}{r}{0.3\linewidth}
    \centering
    \includegraphics[width=0.3\linewidth]{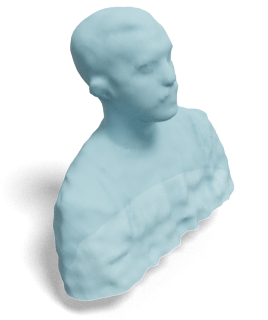}
\includegraphics[width=0.3\linewidth]{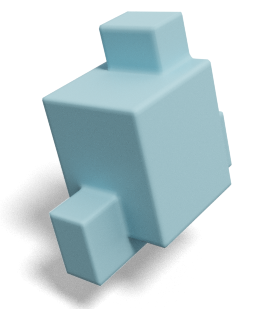}   \includegraphics[width=0.3\linewidth]{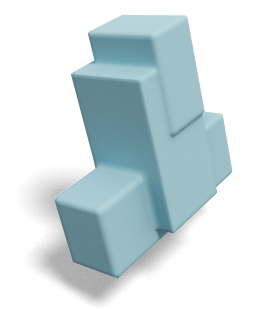}\\
\makebox[0.3\linewidth]{\footnotesize $\mathcal{S}$}
\makebox[0.3\linewidth]{\footnotesize Iter 0}
\makebox[0.3\linewidth]{\footnotesize Iter 5}\\
\includegraphics[width=0.3\linewidth]{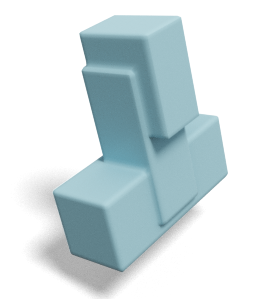} \includegraphics[width=0.3\linewidth]{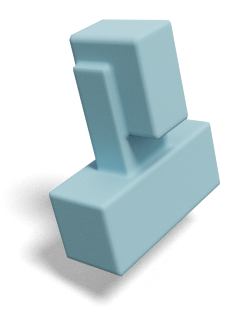} \includegraphics[width=0.3\linewidth]{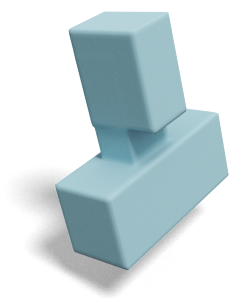}\\
\makebox[0.3\linewidth]{\footnotesize Iter 10}
\makebox[0.3\linewidth]{\footnotesize Iter 45}
\makebox[0.3\linewidth]{\footnotesize Iter 55}
\end{wrapfigure} 
After obtaining the SDF $F_{\text{sdf}}$ of the polycube $\mathcal{D}$, we deprecate the SDF network $G_\text{sdf}$ and instead use the deformation network $F_\text{def}$ and parametric domain SDF $F_\text{sdf}$ network to represent the source surface $\mathcal{S}$. 
The right inset showcases a polycube learning example for a human head model. Given a user-specified parameter $k=3$, our method initializes three randomly placed boxes. It proceeds to minimize the loss function
$\mathcal{L}_{\text{param}}$, iteratively refining the size and positioning of each box to ensure the polycube closely approximates the surface $\mathcal{S}$. This parametric domain learning phase typically converges within 30 to 60 iterations, taking under 10 minutes to complete. For more details, see the supplementary video.

\subsection{Bi-directional Deformation}\label{subsec: bi-deform}

The bi-directional deformation module serves as the pivotal component within our neural parameterization framework. As illustrated in Fig.~\ref{fig:pipeline}, this module consists of two sub-networks: $F_{\text{def}}$, which is responsible for the forward mapping from the original model to the parameter domain, and $F_{\text{inv-def}}$, which establishes the inverse mapping.

Let $\mathcal{S}$ be the 3D surface and $\mathcal{D}$ be the parametric domain; both are represented by the zero level-set of neural signed distance fields. 
For an arbitrary point $\mathbf{p}\in \mathcal{S}$, the forward deformation $F_\text{def}(\mathbf{p}, \mathbf{z}_s)$ computes a displacement vector, mapping the point $\mathbf{p}$ to a corresponding point $\mathbf{p}'\in\mathcal{D}$ in the parametric domain as follows:
\begin{equation}
    \mathbf{p}'=\mathbf{p}+F_\text{def}(\mathbf{p}, \mathbf{z}_s).
\end{equation}
Conversely, we employ $F_{\text{inv-def}}$ to map points inversely from the  parametric domain $\mathcal{D}$ back to the original shape $\mathcal{S}$:
\begin{equation}
    \mathbf{p}'' = \mathbf{p}' + F_{\text{inv-def}}(\mathbf{p}', \mathbf{z}_s),
\end{equation}
where the point $\mathbf{p}''$ is expected to be in close proximity to $p$. 
To encourage the smoothness and stability of the deformations, we impose an $L_2$ loss on the magnitude of the displacement vectors:
\begin{equation}
    \mathcal{L}_\text{smooth} = \| F_\text{def}(\mathbf{p}, \mathbf{z}_s) \|_2 + \| F_{\text{inv-def}}(\mathbf{p}', \mathbf{z}_s)\|_2.
\end{equation}
This loss helps minimize abrupt and irregular deformations, leading to more coherent and gradual changes between $\mathcal{S}$ and $\mathcal{D}$.

In training the networks $F_\text{def}$ and $F_{\text{inv-def}}$, our objective is to achieve a bijective mapping between the 3D surface $\mathcal{S}$ and the parametric domain $\mathcal{D}$. 
To achieve this goal, we adopt a cycle loss, which is similar to NeuTex~\cite{xiang2021neutex}: 
\begin{equation}
\label{eq: cyc}
    \mathcal{L}_{\text{cycle}} = \sum\nolimits_{\mathbf{p}\in\mathcal{S}} \lambda(\mathbf{p})\| \mathbf{p} - \mathbf{p}''\|_2,
\end{equation}
where the weight $\lambda$ characterizes the importance of sample $\bf p$ to the loss. The reason that we do not consider all sample points equally is that a sample point may not be exactly on the zero level-set of the SDF. To tolerate such an inaccuracy, we define $\lambda(\mathbf{p}) = T(\mathbf{p})(1-\exp(-\sigma(s) \delta(\textbf{p})))$ as the color weight used in volume rendering~\cite{mildenhall2020nerf}. Here, $T(\mathbf{p})$ is the transparency of sample $\mathbf{p}$ in the viewing direction, $\delta$ is the length of the sample interval along the ray, and $s$ is the signed distance value of the point $\mathbf{p}$. Clearly, when $\bf p$ is away from the surface, the coefficient $\lambda(\mathbf{p})$ has few effects on this loss. 
We refer the readers to~\cite{mildenhall2020nerf} for details about volume rendering and the computation of transparency $T$. 

By explicitly minimizing the distance between $\mathbf{p}$ and $\mathbf{p}''$, 
the cycle loss effectively prevents scenarios where two distinct points $\mathbf{p}_1,\mathbf{p}_2\in\mathcal{S}$ map to an identical point $\mathbf{p}'\in\mathcal{D}$. In such cases, the inverse deformation $F_{\text{inv-def}}$ would be unable to map the single point $\mathbf{p}'$ back to the respective distinct points $\mathbf{p}_1''$ and $\mathbf{p}_2''$.

\subsection{Laplacian Regularization}
\label{subsec: laplace}
 \begin{wrapfigure}{r}{0.3\linewidth}
    \centering
    \includegraphics[width=0.95\linewidth]{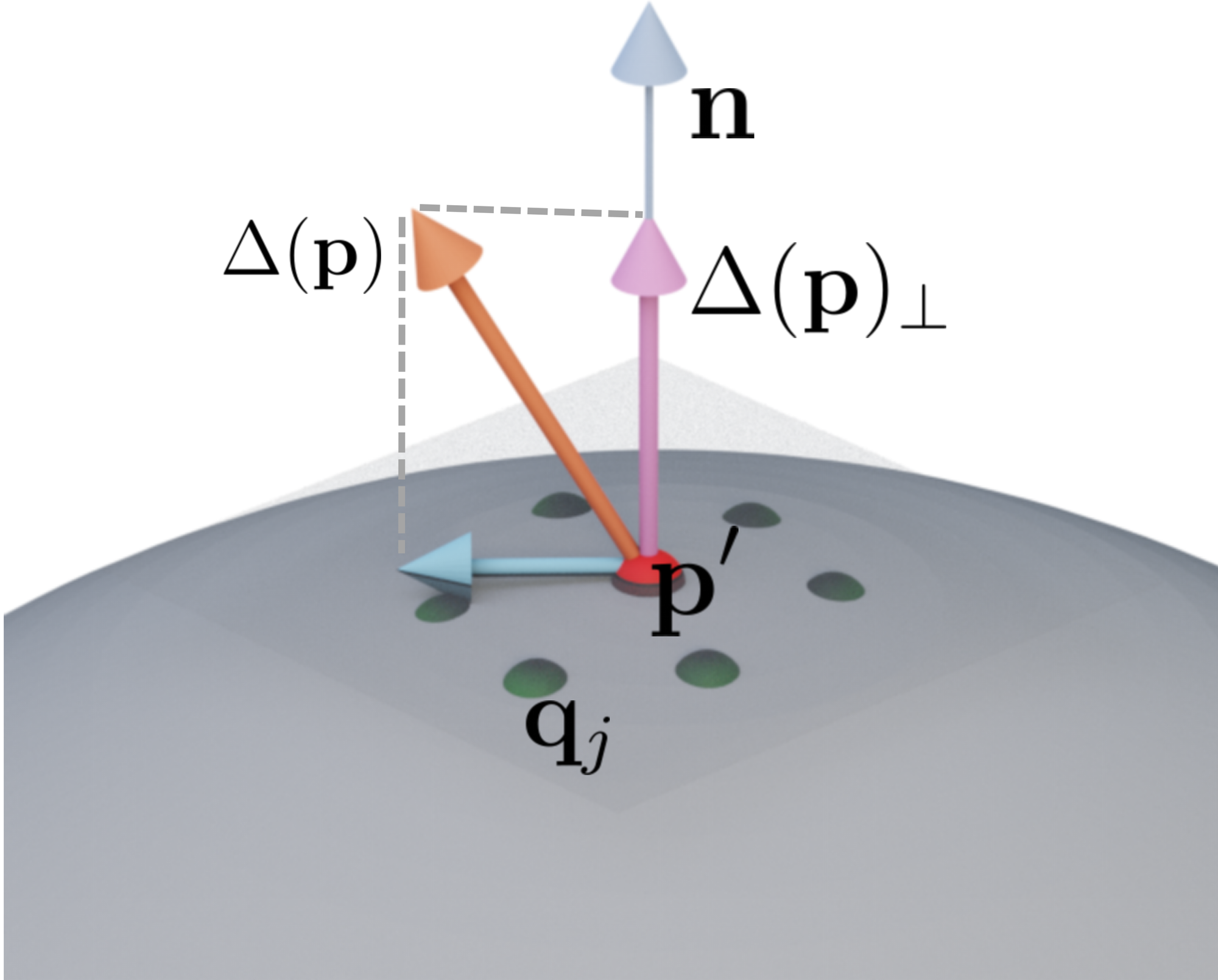}   
\end{wrapfigure} 
 Mitigating angle distortion is crucial in ensuring the quality of parameterization. One approach to reduce angle distortion involves minimizing the ratio of the two singular values of the Jacobian matrix associated with the parameterization~\cite{floater2005surface}. 
However, considering the representation of the surface $\mathcal{S}$, the parametric domain $\mathcal{D}$, and the parameterization $F_{\text{def}}$ as neural implicit functions in our approach, directly computing the Jacobian matrix is a non-trivial task. To circumvent this, we employ a Laplacian regularizer in our implementation to control angle distortion. Achieving the minimizer of the Laplacian of the map results in a harmonic map. For genus-0 surfaces, harmonic maps are also conformal, which preserves angles~\cite{DBLP:journals/tmi/GuWCTY04}.
 Although for surfaces of higher genus, a harmonic map is not necessarily conformal, it still serves an effective tool for reducing angle distortion~\cite{DBLP:conf/sgp/GuY03}. 
 Specifically, for a sample point $\mathbf{p}\in\mathcal{S}$ and its neighboring points $\mathbf{q}_j\in\mathcal{S}$, the forward deformation maps them to $\mathbf{p}'\in\mathcal{D}$ and $\mathbf{q}_j'\in\mathcal{D}$, respectively.
 We compute the Laplacian of the sample $\mathbf{p}$ as follows:
\begin{equation}
    \Delta(\mathbf{p}) = \sum\nolimits_{j=1}^m \omega_{j}(\mathbf{p}' - \mathbf{q}'_{j}).
    \label{eq: lap}
\end{equation}
Here, the weights $w_j$ are defined using the metric from the original surface $\mathcal{S}$ as:
\begin{equation}
    \omega_j = \exp\left(-\|\mathbf{p}-\mathbf{q}_j\|_2/l\right),
\end{equation}
where $l$ is the average distance from point $\mathbf{p}$ to its neighboring points. Subsequently, the weights are normalized $\omega_j=\omega_j/\sum_{k=1}^m\omega_k$.
Notice that $\Delta(\mathbf{p})\in\mathbb{R}^3$ is a 3D vector, which may not necessarily lie on the tangent plane. Therefore, we define the normal component of the Laplacian as:
\begin{equation}
    \Delta(\mathbf{p})_{\perp} = \langle \Delta(\mathbf{p}) , \mathbf{n}\rangle \mathbf{n},
\end{equation}
where $\mathbf{n}$ is the normal at $\mathbf{p}'$, and $\langle, \rangle$ represents the inner product. Summing the tangential components yields the Laplacian loss:
\begin{equation}
\label{eq:la_regularizer}
    \mathcal{L}_{\text{Lap}}=\sum\nolimits_{\mathbf{p}\in\mathcal{S}}\|\Delta(\mathbf{p})-\Delta(\mathbf{p})_{\perp}\|_2.
\end{equation}
In our implementation, we set $m=6$ neighboring points for each sample $\mathbf{p}$. 

\subsection{Appearance Decomposition}
\label{subsec: app_model}
Radiance represents the directional emission of color, illustrating how the appearance of an object varies in response to changes in the observational viewpoint~\cite{mildenhall2020nerf}. Drawing inspiration from previous works~\cite{ma2022neural,Ye2023IntrinsicNeRF} and applying the assumptions of Lambertian and grayscale shading~\cite{fan2018revisiting}, we decompose the radiance field on the parametric domain into two components: a view-independent material field and a view-dependent shading field (see Fig.~\ref{fig:shade_tranfer}(a)). This decomposition allows for independent editing of material and shading. More precisely, for a sample point $\mathbf{p}\in\mathcal{S}$, we employ two sub-networks, $F_{\text{mat}}$ and $F_{\text{shd}}$, to decompose the radiance $\mathbf{r}\in \mathbb{R}^3$ at point $\mathbf{p}$ as follows:
\begin{equation}
        \mathbf{r}(\mathbf{p}) = F_{\text{mat}}(\mathbf{p}', \mathbf{n}, \mathbf{z}_a) \cdot \exp({F_{\text{shd}}(\mathbf{p}', \mathbf{n}, \mathbf{v}, \mathbf{z}_a)}),\label{eq: rad}
\end{equation}
where $\mathbf{p}'\in\mathcal{D}$ is the image of $\bf p$ and $\mathbf{n}$ is the unit normal of $\bf p$ on the original surface. The material network $F_{\text{mat}}$ takes the mapped point $\mathbf{p}'$, the surface normal $\bf n$, and the appearance latent code $\mathbf{z}_a$ as input, generating the view-independent material field in response. The shading network $F_{\text{shd}}$, which is designed for computing the view-dependent shading field, takes an additional input: the viewpoint $\bf v$. 
To ensure the shading sub-network $F_{\text{shd}}$ concentrates on \textbf{local} shading features, we employ an $L_1$ loss to the output of $F_{\text{shd}}$, encouraging sparsity,
\begin{equation}
    \mathcal{L}_{\text{shd}} = \| F_{\text{shd}}(\mathbf{p}', \mathbf{n}, \mathbf{v}, \mathbf{z}_a) \|_1 .
\end{equation}

To compute the color, we need the signed distance values for the samples $\mathbf{p}$. As mentioned above, our network adopts only one SDF network $F_{\text{sdf}}$, which represents the geometry of the parametric domain $\mathcal{D}$. We use the forward deformation $F_{\text{def}}$ to compute $s(\mathbf{p})$ as 
\begin{equation}
 \label{eq: sdf}
    s(\mathbf{p}) = F_{\text{sdf}}\left(\mathbf{p} + F_{\text{def}}(\mathbf{p}, \mathbf{z}_s)\right).
\end{equation}
Then we employ the SDF-based volume rendering~\cite{yariv2021volume} to integrate the radiances along the ray to compute color $\mathbf{c}_{\text{pred}}= \sum_i \lambda_i\mathbf{r}_i$, where $\mathbf{r}_i$ is the radiance of a sample point $\mathbf{p}_i$ and $\lambda_i$ is the same color weight as used in Eq.~(\ref{eq: cyc}).

\subsection{Training Losses}
\label{subsec:loss}

In contrast to NeP~\cite{ma2022neural}, which necessitate prior information such as tracked mesh and UV mapping for supervision, our method is capable of reconstructing 3D geometry and computing the corresponding texture map directly from multi-view images. This process is supervised by the image loss
\begin{align}
    \mathcal{L}_{\text{rgb}} = \| \mathbf{c}_{\text{gt}} - \mathbf{c}_{\text{pred}} \|_1,
\end{align}
where $\mathbf{c}_{\text{gt}}$ and $\mathbf{c}_{\text{pred}}$ are the ground-truth and predicted colors, respectively. 
To ensure the shape and appearance latent codes, $\mathbf{z}_s$ and $\mathbf{z}_a$, conform to Gaussian distributions, we incorporate a regularization term, expressed as $\mathcal{L}_{\text{code}} = \| \mathbf{z}_s \|_2 + \| \mathbf{z}_a \|_2$~\cite{park2019deepsdf}.
Additionally, with geometry represented as signed distance fields,  we employ the Eikonal loss~\cite{yariv2020multiview, gropp2020implicit}, defined as $\mathcal{L}_{\text{Eik}} = (\| \nabla s \|_2- 1)^2$. Putting it all together, we define the total loss function as follows:
\begin{equation}
    \mathcal{L} = \mathcal{L}_{\text{rgb}} + \lambda_1 \mathcal{L}_{\text{Eik}} + \lambda_2 \mathcal{L}_{\text{cycle}} + \lambda_3 \mathcal{L}_{\text{smooth}} 
    + \lambda_4 \mathcal{L}_{\text{Lap}} + \lambda_5 \mathcal{L}_{\text{shd}} + \lambda_6 \mathcal{L}_{\text{code}}.
    \label{eq: loss}
\end{equation}
In our implementation, we empirically set the coefficients as $\lambda_1=\lambda_2=\lambda_5=\lambda_6=0.01$ and $\lambda_3=\lambda_4=0.001$.

\section{Experiments}
\label{sec:exp}
\textbf{Datasets.} 
For our experiments, we used a customized dataset named FS-Syn, derived from the Facescape dataset~\cite{yang2020facescape}, selecting 10 human head models and configuring 30 fixed viewpoints and lighting conditions to obtain synthesized multi-view images. We also evaluated our method on several man-made objects from the OmniObject dataset~\cite{wu2023omniobject3d}, and the H3DS dataset~\cite{ramon2021h3d}.
\begin{figure}[t]
    \centering
    \rotatebox[origin=t]{90}{GT}
    \includegraphics[width=0.12\linewidth]{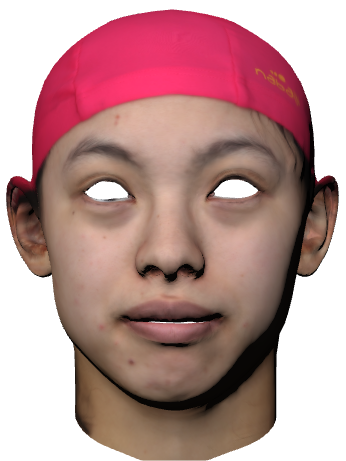} 
    \includegraphics[width=0.12\linewidth]{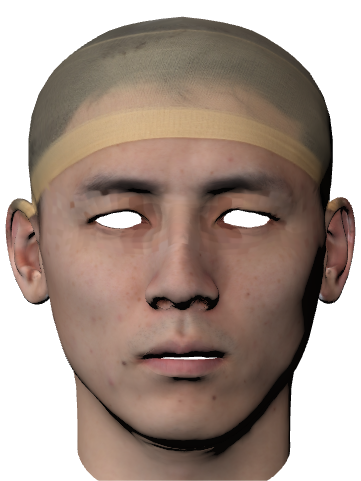} 
    \includegraphics[width=0.12\linewidth]{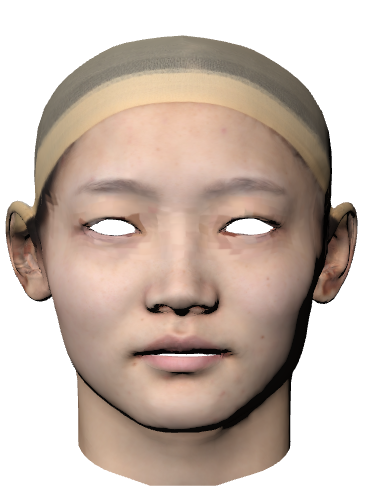} 
    \includegraphics[width=0.12\linewidth]{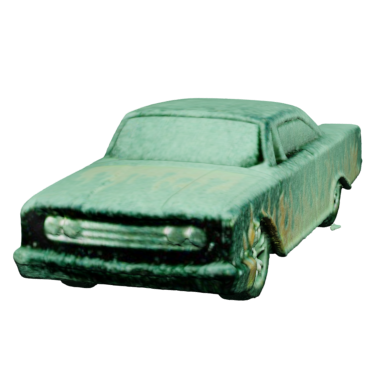} 
    \includegraphics[width=0.12\linewidth]{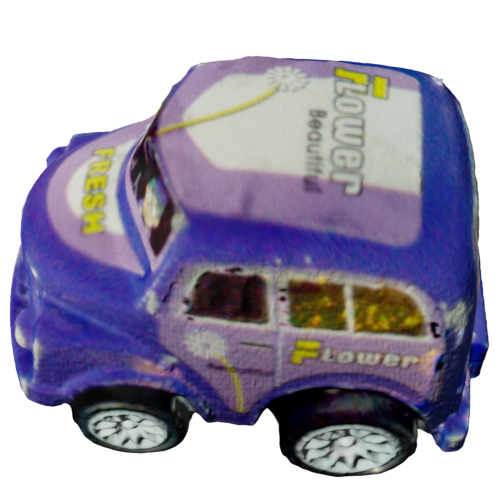} 
    \includegraphics[width=0.12\linewidth]{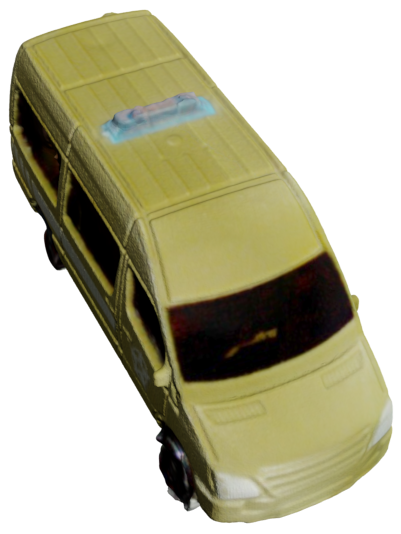}
    \\
    \rotatebox[origin=t]{90}{Normal}
    \includegraphics[width=0.12\linewidth]{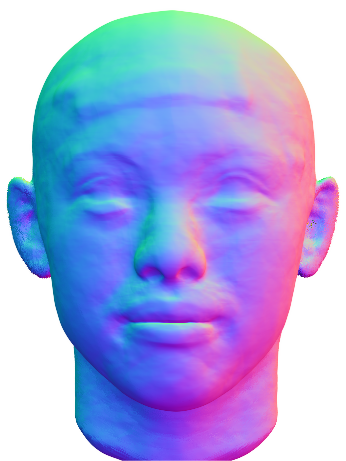} 
    \includegraphics[width=0.12\linewidth]{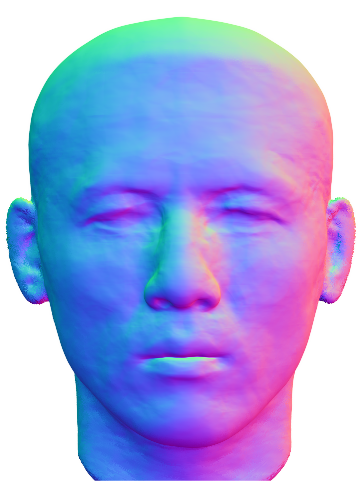} 
    \includegraphics[width=0.12\linewidth]{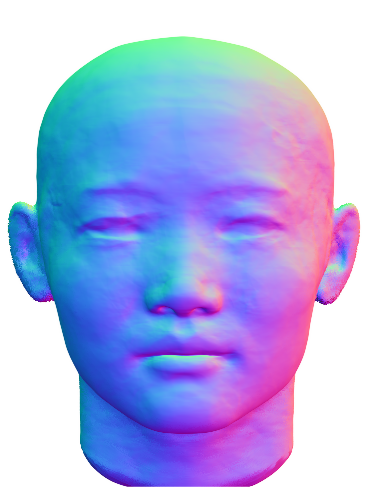} 
    \includegraphics[width=0.12\linewidth]{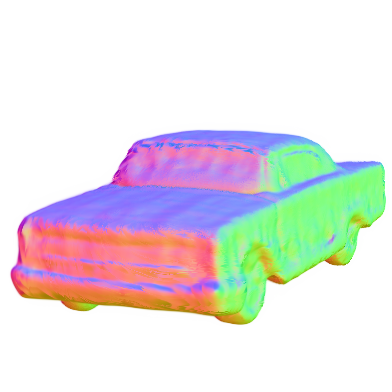} 
    \includegraphics[width=0.12\linewidth]{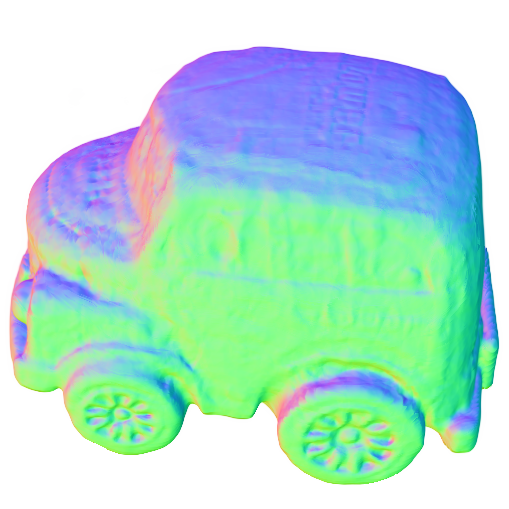} 
    \includegraphics[width=0.12\linewidth]{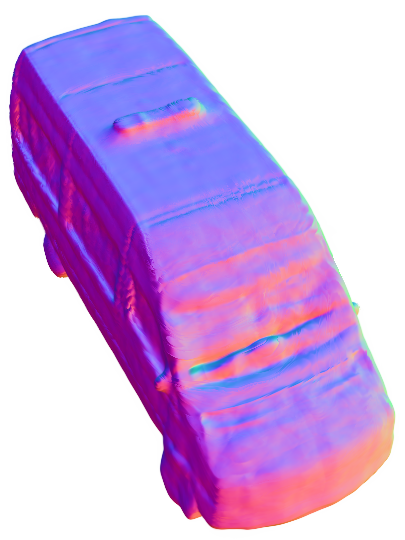}
    \\
    \rotatebox[origin=t]{90}{\footnotesize$\mathcal{D}$}
    \includegraphics[width=0.12\linewidth]{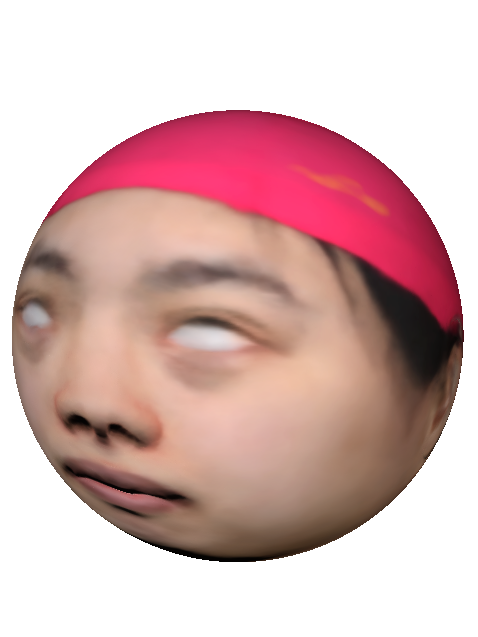} 
    \includegraphics[width=0.12\linewidth]{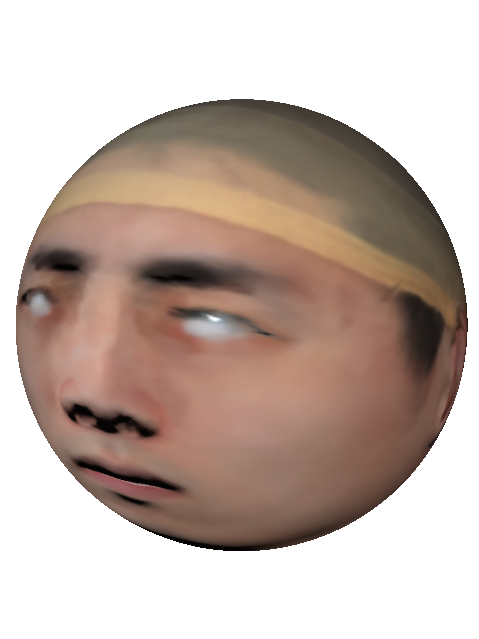} 
    \includegraphics[width=0.12\linewidth]{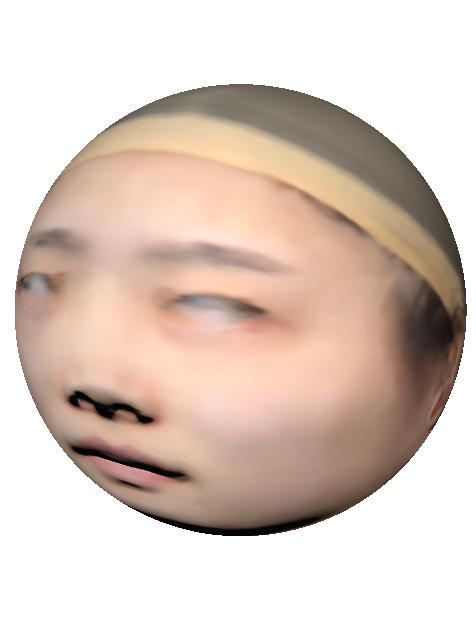} 
    \includegraphics[width=0.12\linewidth]{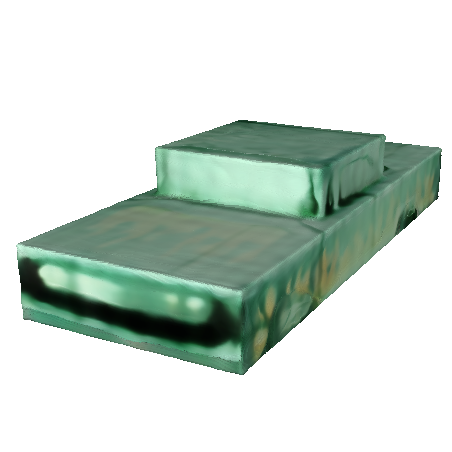} 
    \includegraphics[width=0.12\linewidth]{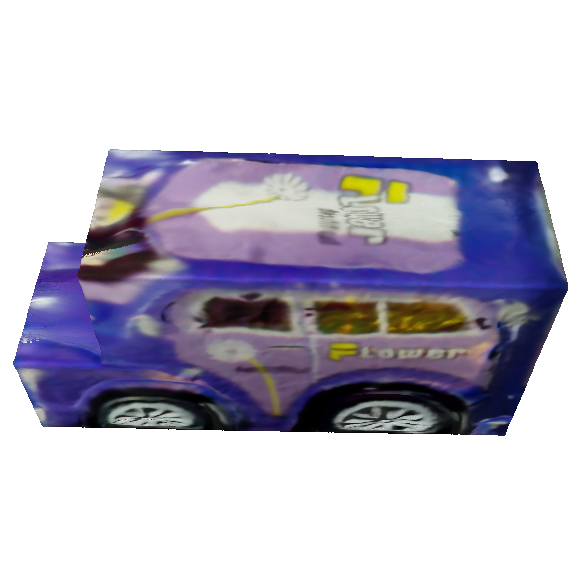} 
    \includegraphics[width=0.12\linewidth]{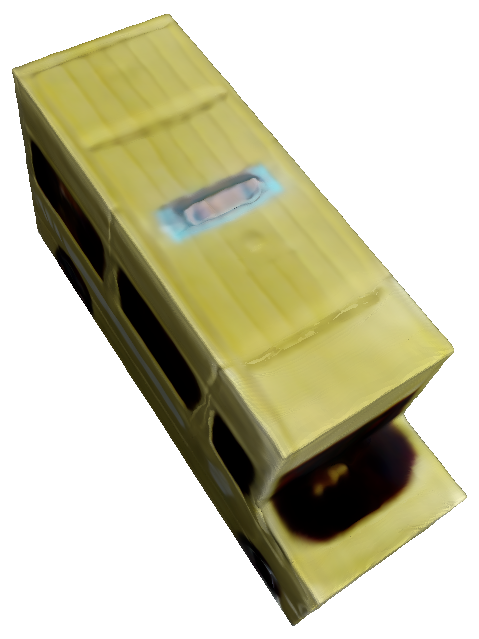}
    \\
    \rotatebox[origin=t]{90}{\footnotesize$\mathcal{S}$}
    \includegraphics[width=0.12\linewidth]{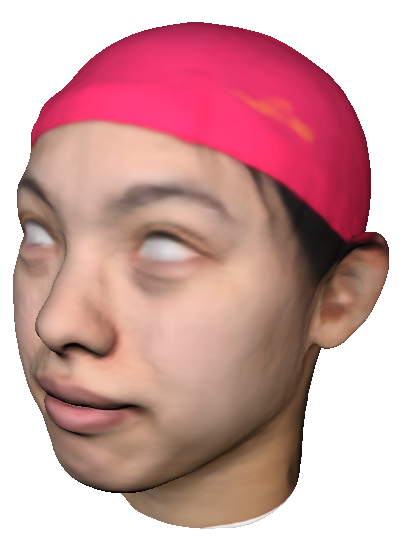} 
    \includegraphics[width=0.12\linewidth]{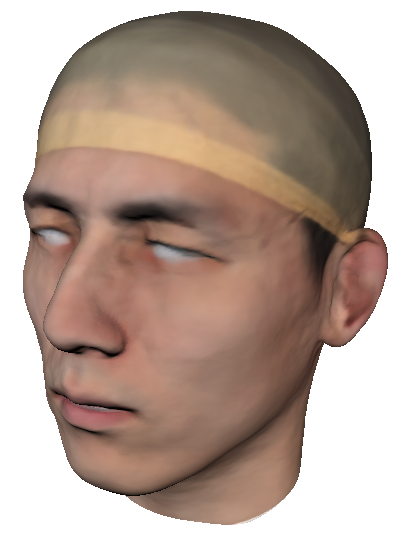} 
    \includegraphics[width=0.12\linewidth]{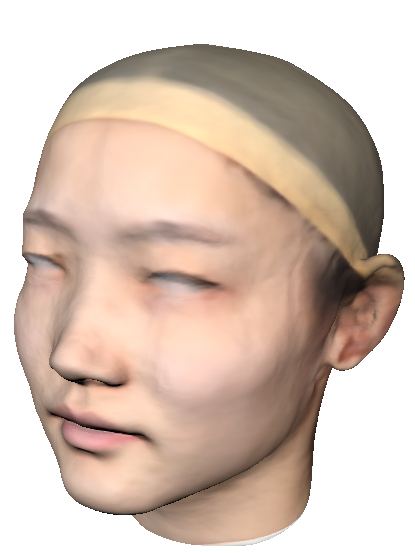} 
    \includegraphics[width=0.12\linewidth]{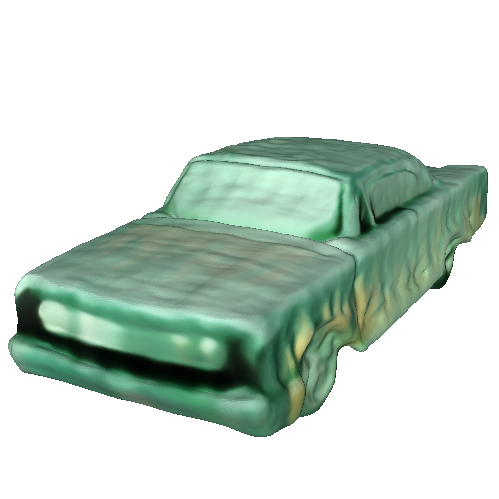} 
    \includegraphics[width=0.12\linewidth]{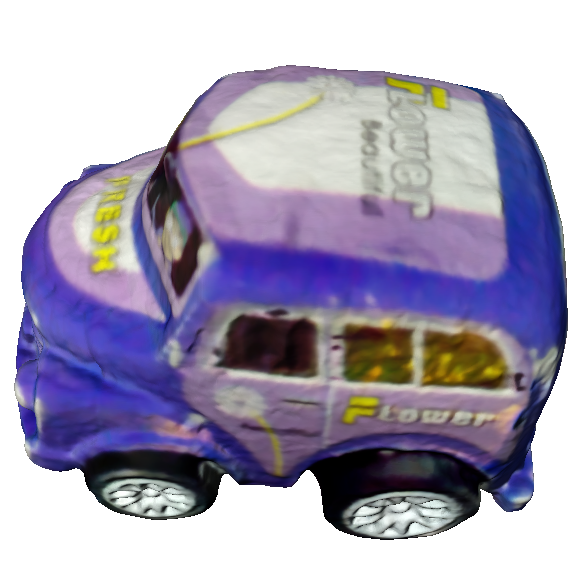} 
    \includegraphics[width=0.12\linewidth]{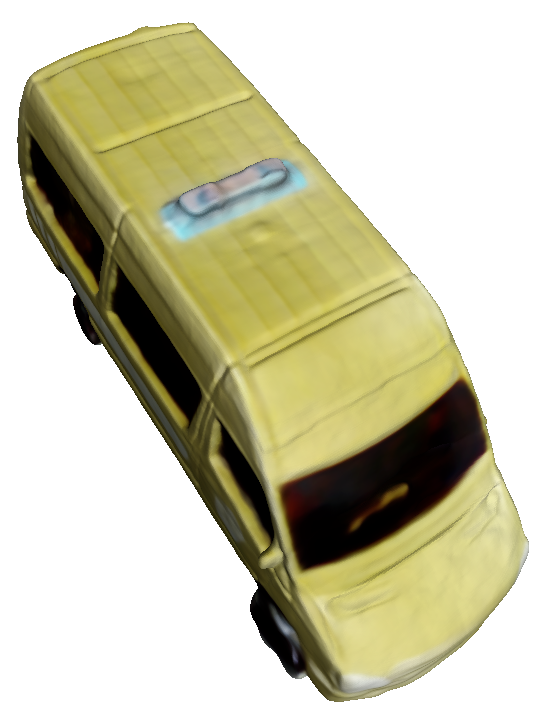}
\caption{Parameterization results. Human heads are co-parameterized to a sphere owing to the clear geometric resemblance. Given the diverse geometry of cars, each is parameterized to a polycube domain (bottom three rows). From top to bottom: the input image, the normal map of the reconstructed geometry $\mathcal{S}$, the parametric domain $\mathcal{D}$, and the reconstructed surface $\mathcal{S}$. See supplementary material for additional results.}
    \label{fig:para-result}
\end{figure}

\textbf{Training details and metrics.}
All the sub-networks $F_{\text{def}}$, $F_{\text{inv-def}}$, $F_{\text{sdf}}$, $F_{\text{mat}}$, and $F_{\text{shd}}$ are MLPs. See the supplementary material for other detailed network hyperparameters. Once the parametric domain is chosen, 
we fixed $F_{\text{sdf}}$ and trained other sub-networks for 2,000 epochs driven by the loss function as Eq.~(\ref{eq: loss}). We applied the standard Marching Cubes algorithm to extract triangle meshes for the zero level-set of the source surface $\mathcal{S}$ and then used the forward map to obtain the mesh representing the parametric surface. With triangle meshes, we measured the quality of the computed parameterization by the angle distortion $\mathcal{E}_{\text{angle}}$ and area distortion~\cite{degener2003adaptable,he2009divide} $\mathcal{E}_{\text{area}}$ as follows:
\begin{equation}
\mathcal{E}_{\text{angle}} = \frac{\cot \alpha|a|^{2}+\cot \beta|b|^{2}+\cot \gamma|c|^{2}}{4\text{area}(\Delta_i)}, ~~~
    \mathcal{E}_{\text{area}} = \frac{1}{2} \left( \frac{\text{area}(\Delta_i)}{\text{area}(\hat{\Delta}_i)}+\frac{\text{area}(\hat{\Delta}_i)}{\text{area}(\Delta_i)} \right),\nonumber
\end{equation}
where $a, b, c$ represent the side lengths of the triangle $\Delta_i$ of the original surface. $\alpha, \beta, \gamma$ are angles of the corresponding triangle $\hat{\Delta}_i$ in parametric domain.In the case of conformal mapping, $\mathcal{E}_{\text{angle}}$ is equal to 1, and $\mathcal{E}_{\text{area}}$ is also equal to 1 for an isometric map.

\begin{figure}[htbp]
    \centering
    \includegraphics[width=0.17\linewidth]{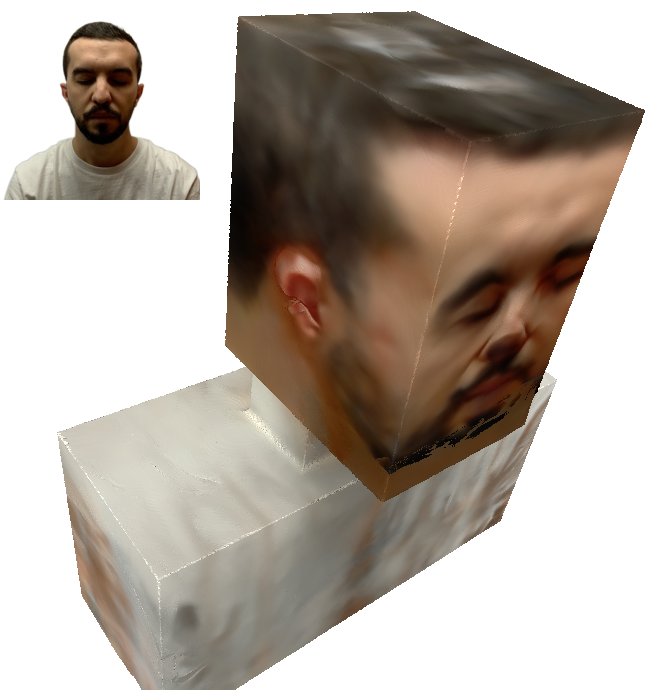}
    \includegraphics[width=0.17\linewidth]{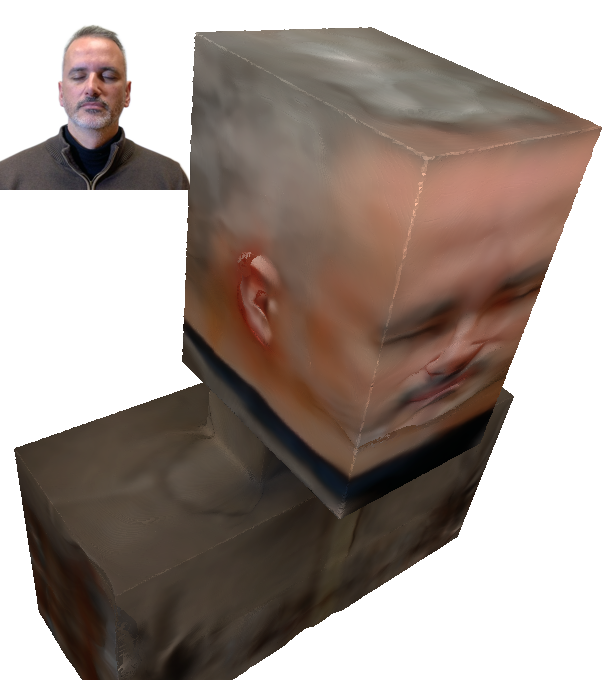}
    \includegraphics[width=0.17\linewidth]{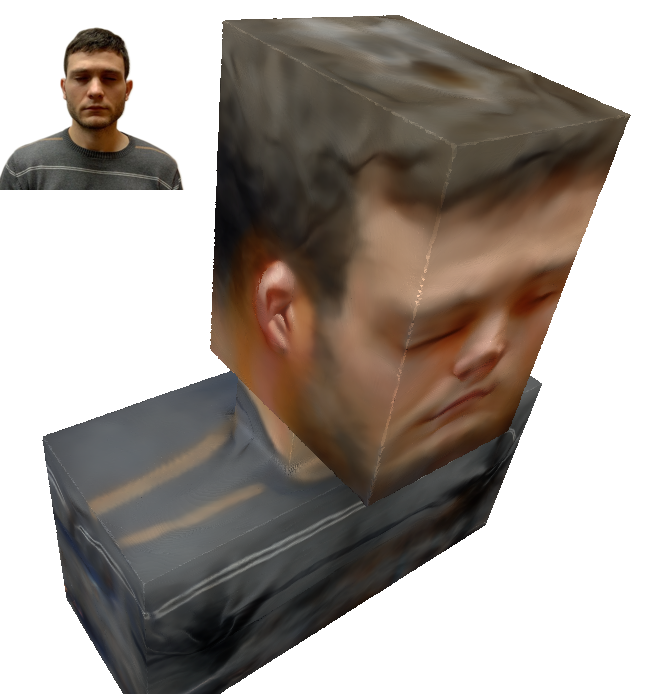}
    \includegraphics[width=0.17\linewidth]{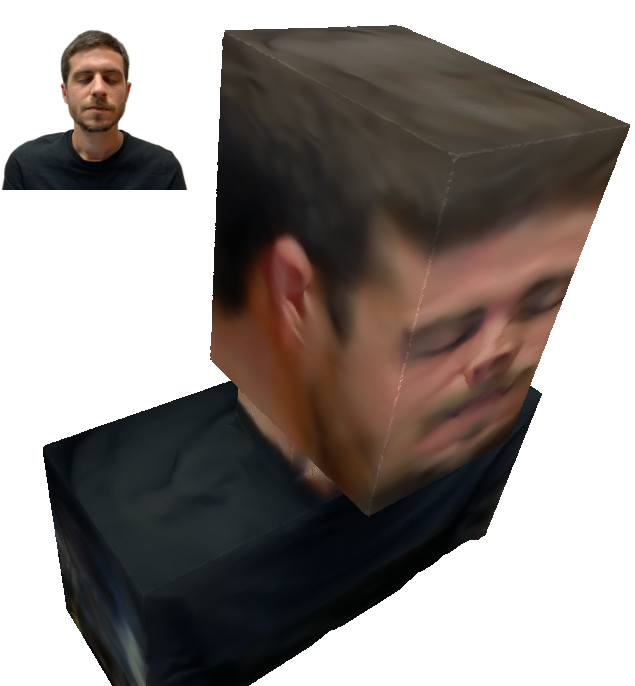}
    \includegraphics[width=0.17\linewidth]{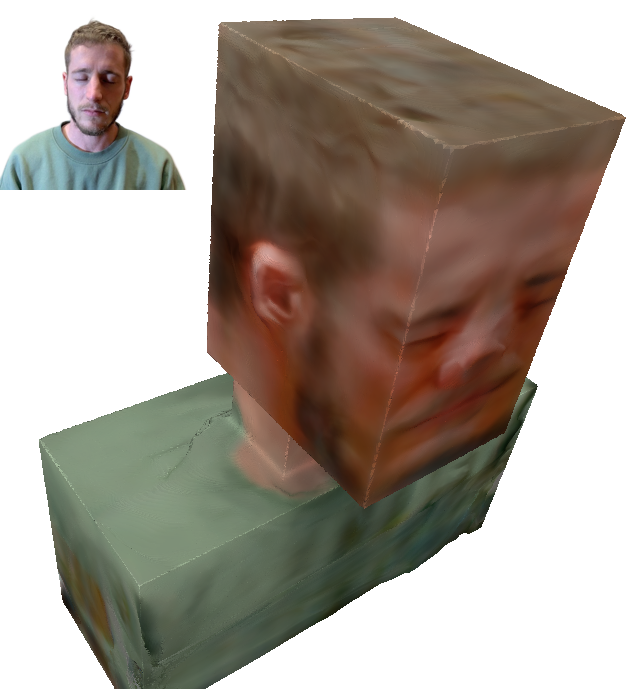}
\\
    \makebox[0.17\linewidth]{\footnotesize 1.545/1.396}
    \makebox[0.17\linewidth]{\footnotesize 1.755/1.446}
    \makebox[0.17\linewidth]{\footnotesize 1.394/1.302}
    \makebox[0.17\linewidth]{\footnotesize 1.271/1.216}
    \makebox[0.17\linewidth]{\footnotesize 1.351/1.275}
    \caption{
Results of cross-polycube parameterization on the H3DS dataset~\cite{ramon2021h3d}. We learned a common polycube domain for all human heads with the parameter $k=3$ (i.e., the polycube is formed by 3 boxes) and then computed the polycube parameterization for all the human heads simultaneously. Since human heads share similar geometries, the computed polycube parameterizations are also highly consistent. The values below each figure represent the angle distortion and the area distortion, respectively.}
    \label{fig:complex}
\end{figure}
\textbf{Results.}
Our framework constructs a nearly bijective map between objects and multiple parametric domains such as spheres and polycubes, as shown in Fig.~\ref{fig:para-result} and Fig.~\ref{fig:complex}. It supports co-parameterization of multiple objects and facilitates material and shading transfer. Fully compatible with existing neural rendering pipelines, our neural parameterization algorithm uses multi-view images to simultaneously reconstruct 3D geometry and create texture maps. As shown in Fig.~\ref{fig:para-result}, we achieve high-fidelity results without prior information.

\begin{figure}[htbp]
    \centering
    \includegraphics[width=0.18\linewidth]{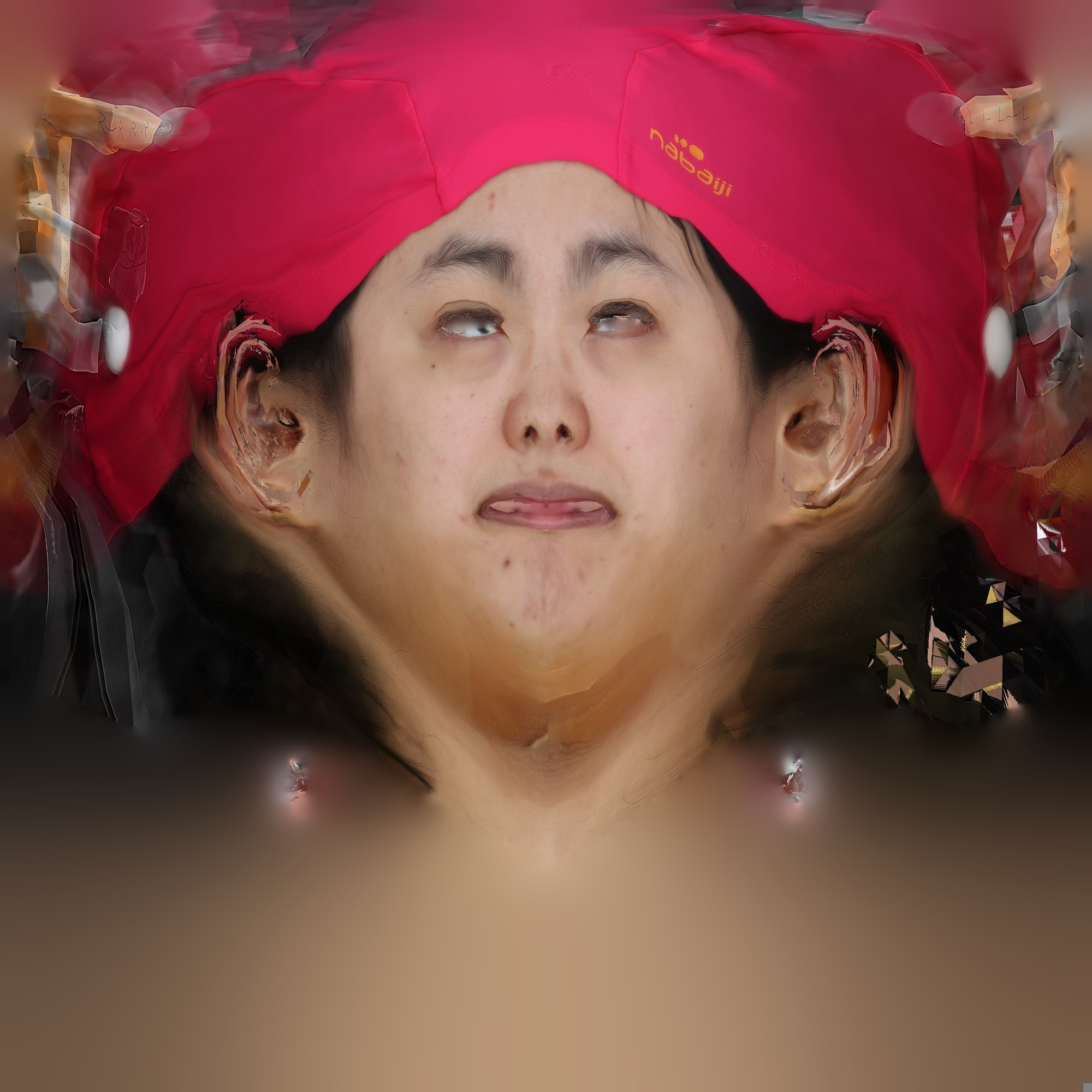}\hfill
    \includegraphics[width=0.18\linewidth]{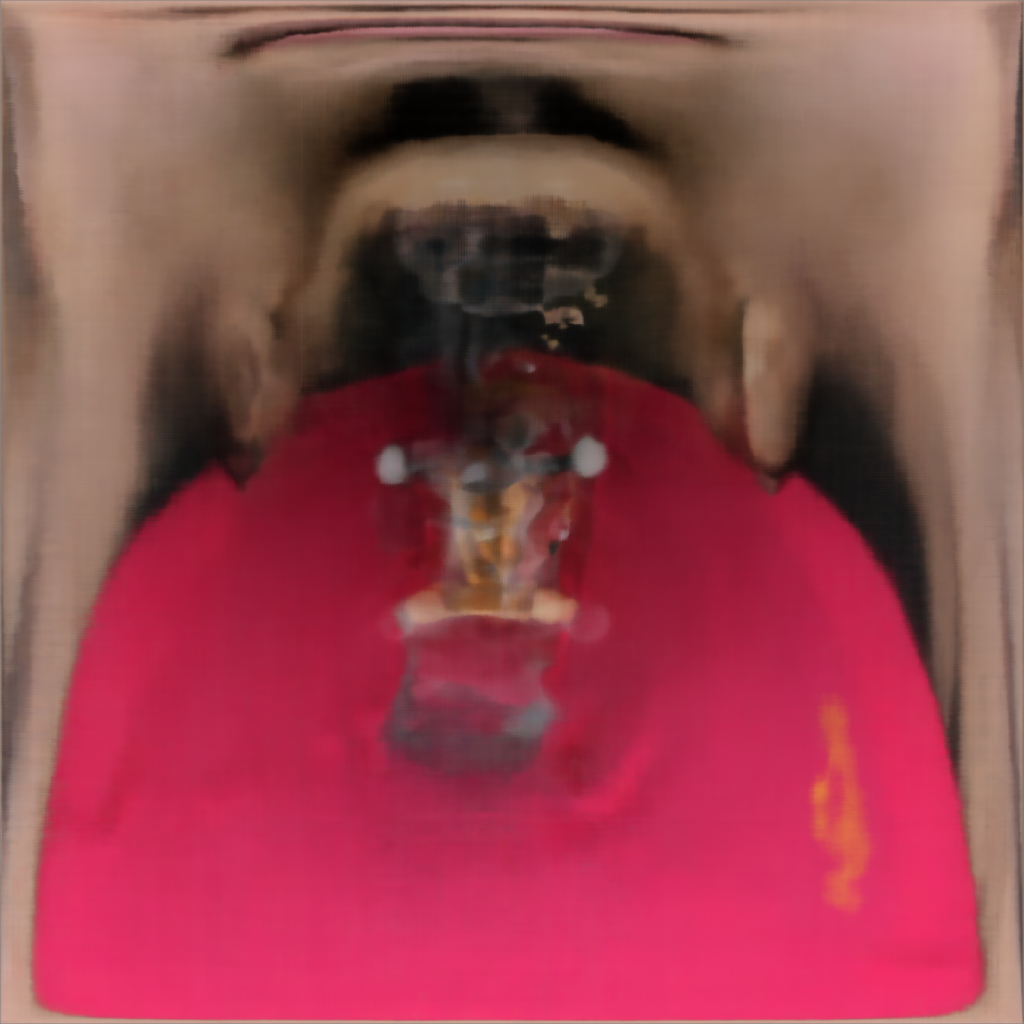}\hfill
    \includegraphics[width=0.18\linewidth,angle=90]{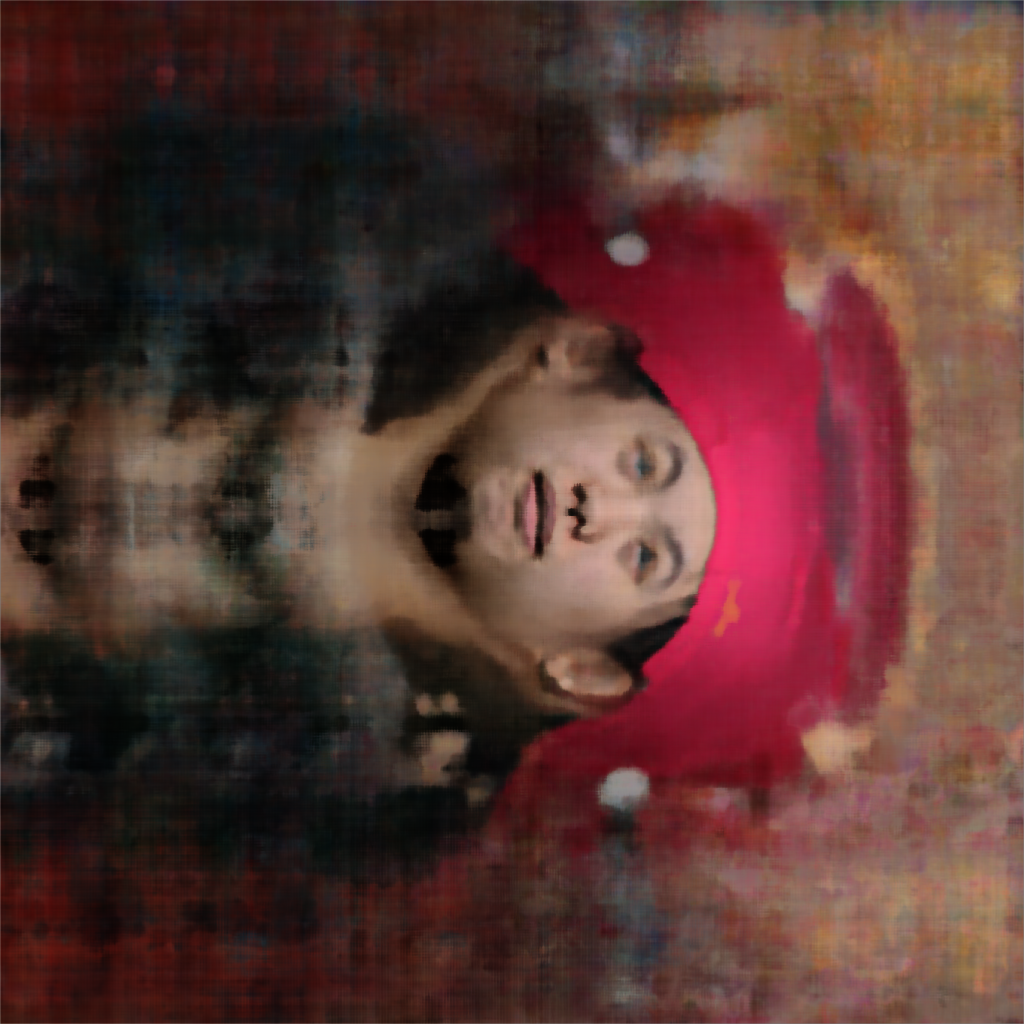}
    \includegraphics[width=0.16\linewidth]{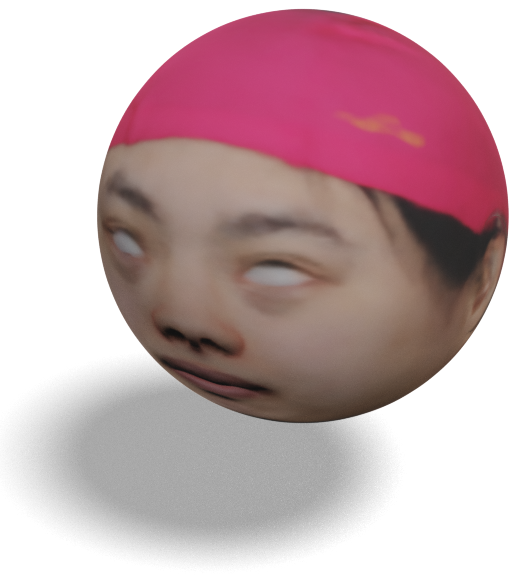}\hfill
    \includegraphics[width=0.2\linewidth]{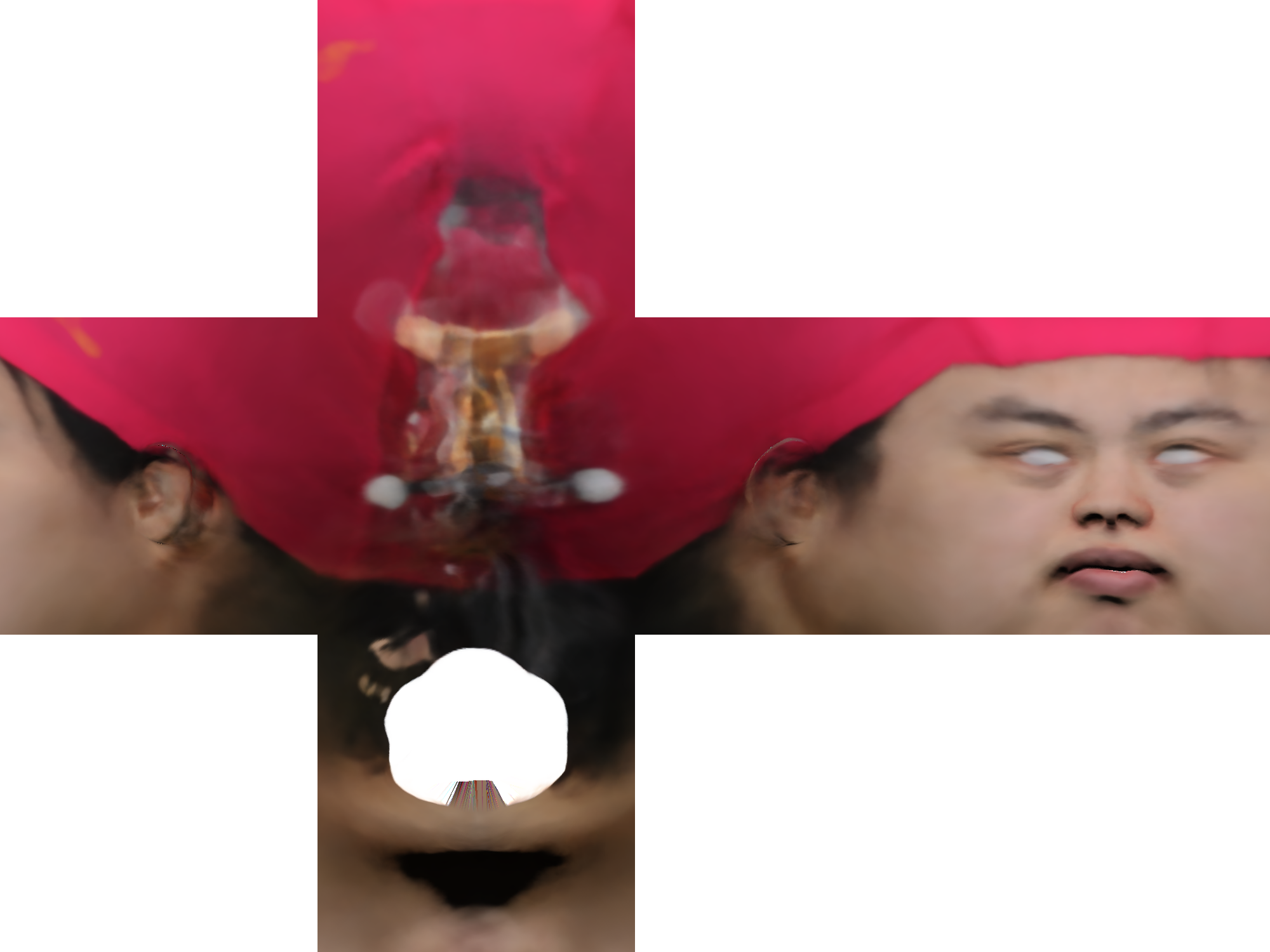}
    \\
    \makebox[0.18\linewidth]{\footnotesize UV prior }\hfill
    \makebox[0.18\linewidth]{\footnotesize NeP w/o prior}\hfill
    \makebox[0.18\linewidth]{\footnotesize  w/ prior}
    \makebox[0.36\linewidth]{\footnotesize Ours}
    \caption{Parametrization results of NeP~\cite{ma2022neural} in comparison with our method. NeP heavily relies on the initial UV prior, leading to poor texture maps when such prior information is lacking. Our method learns texture maps from a set of multi-view images without any 3D tracked mesh or UV prior as input.}
    \label{fig:nep-param}
\end{figure}

Compared with NeP~\cite{ma2022neural}, which is the latest neural surface editing work using neural parameterization, our method eliminates the need for prior information from UV mapping. 
Unlike our algorithm, NeP learns a map between surfaces and the UV plane. 
However, it heavily relies on an existing UV mapping as initialization, and yields rather poor texture mapping results without the UV mapping as prior (see Fig.\ref{fig:nep-param}). Moreover, the results from NeP still exhibit significant distortions even with UV prior.

\begin{figure*}[htbp]
    \centering
    \includegraphics[width=0.95\linewidth]{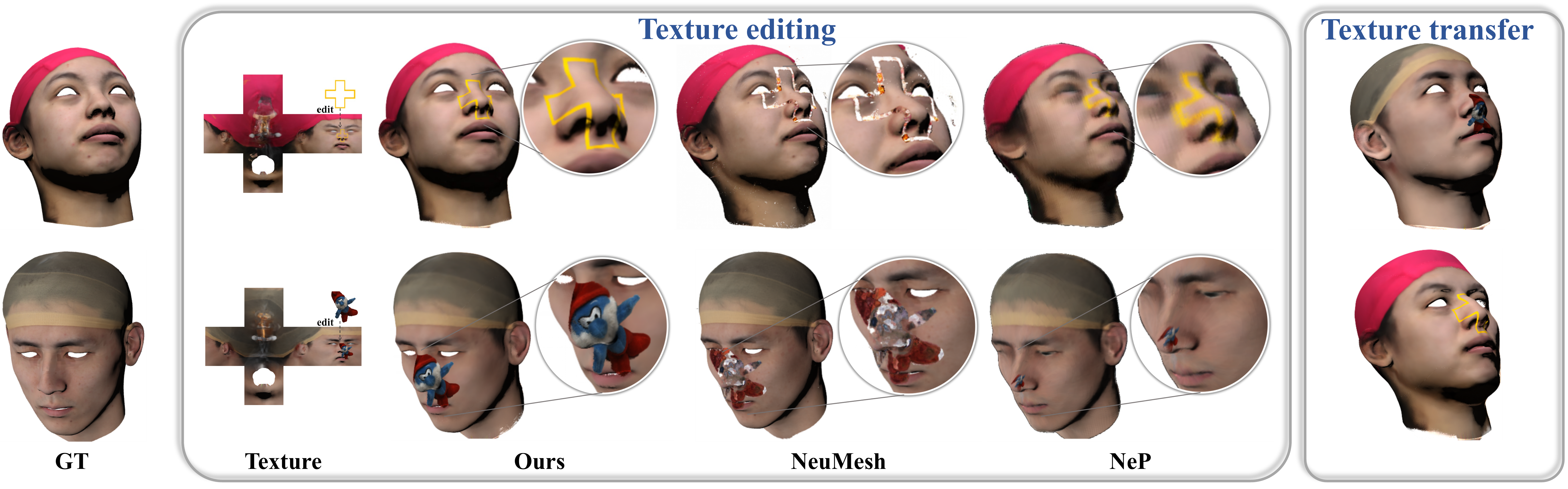}
    \caption{Representing the materials as textures defined in a simple parametric domain enables easy editing. Also, our co-parameterization naturally supports texture transfer between different objects.}
    \label{fig:edit-trans}
\end{figure*}
\textbf{Texture editing.}
Our parameterization results enable flexible and pixel-level editing as shown in Fig.~\ref{fig:edit-trans}. We re-paint the extracted 2D texture map using cubemaps~\cite{greene1986environment} and retrieve the modified color using the established mapping when volume rendering novel images after each editing.
Both NeuMesh~\cite{yang2022neumesh} and NeP~\cite{ma2022neural} can edit the object's texture at a pixel level. We compare the editing and rendering results with NeuMesh and NeP, as shown in Fig.~\ref{fig:edit-trans}. NeuMesh requires a mask of the editing region. It also re-trains the network after each edit, which is time-consuming and cannot guarantee the stability of the training results. Our method outperforms NeuMesh in terms of the visual quality. NeP, which is also neural parameterization-based method, requires a UV prior for initialization. It often yields poor parameterization results without such prior.  Even with the UV prior, its parameterization results exhibit large distortion, making the edit in the 2D domain challenging. Our method is free of prior, and can yield parameterization with low distortion, thereby facilitating pixel-level editing and material transferring without network re-training.

\begin{figure}[htbp]
    \centering
    \includegraphics[width=0.075\linewidth]{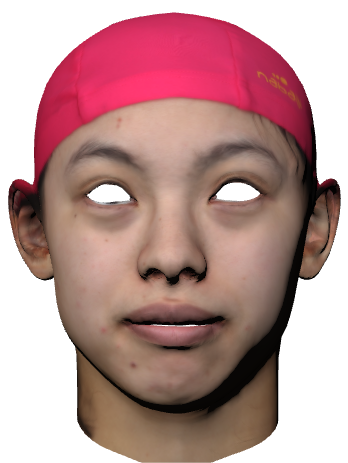}
    \includegraphics[width=0.175\linewidth]{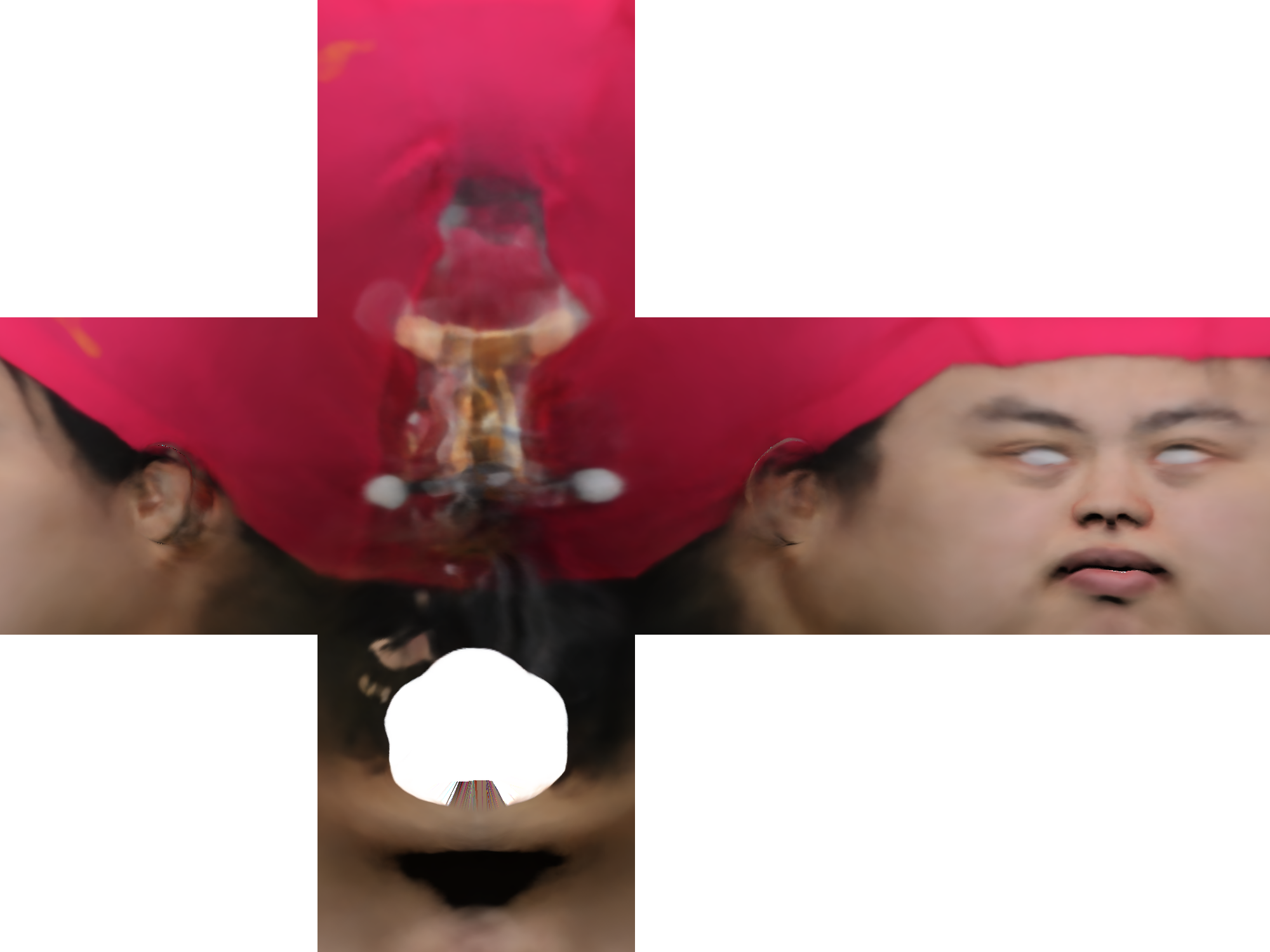}
    \includegraphics[width=0.175\linewidth]{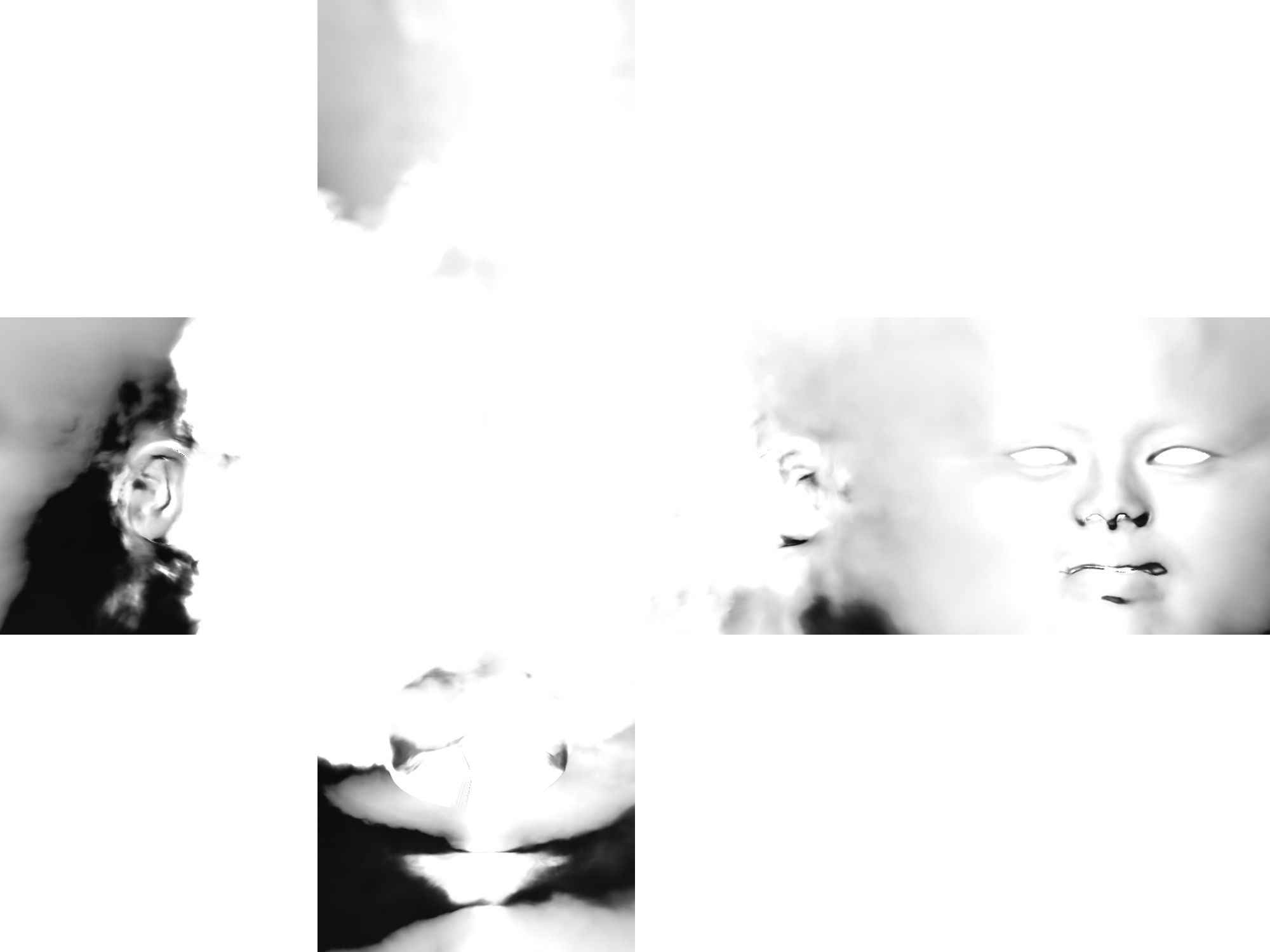}
    \includegraphics[width=0.075\linewidth]{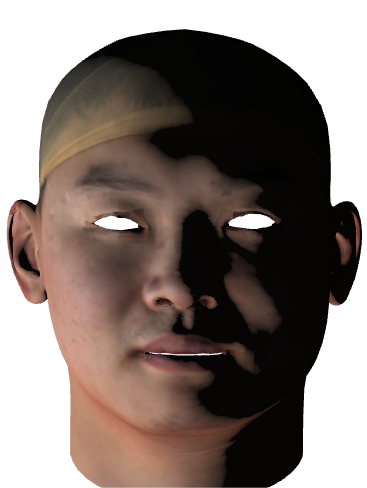}
    \includegraphics[width=0.175\linewidth]{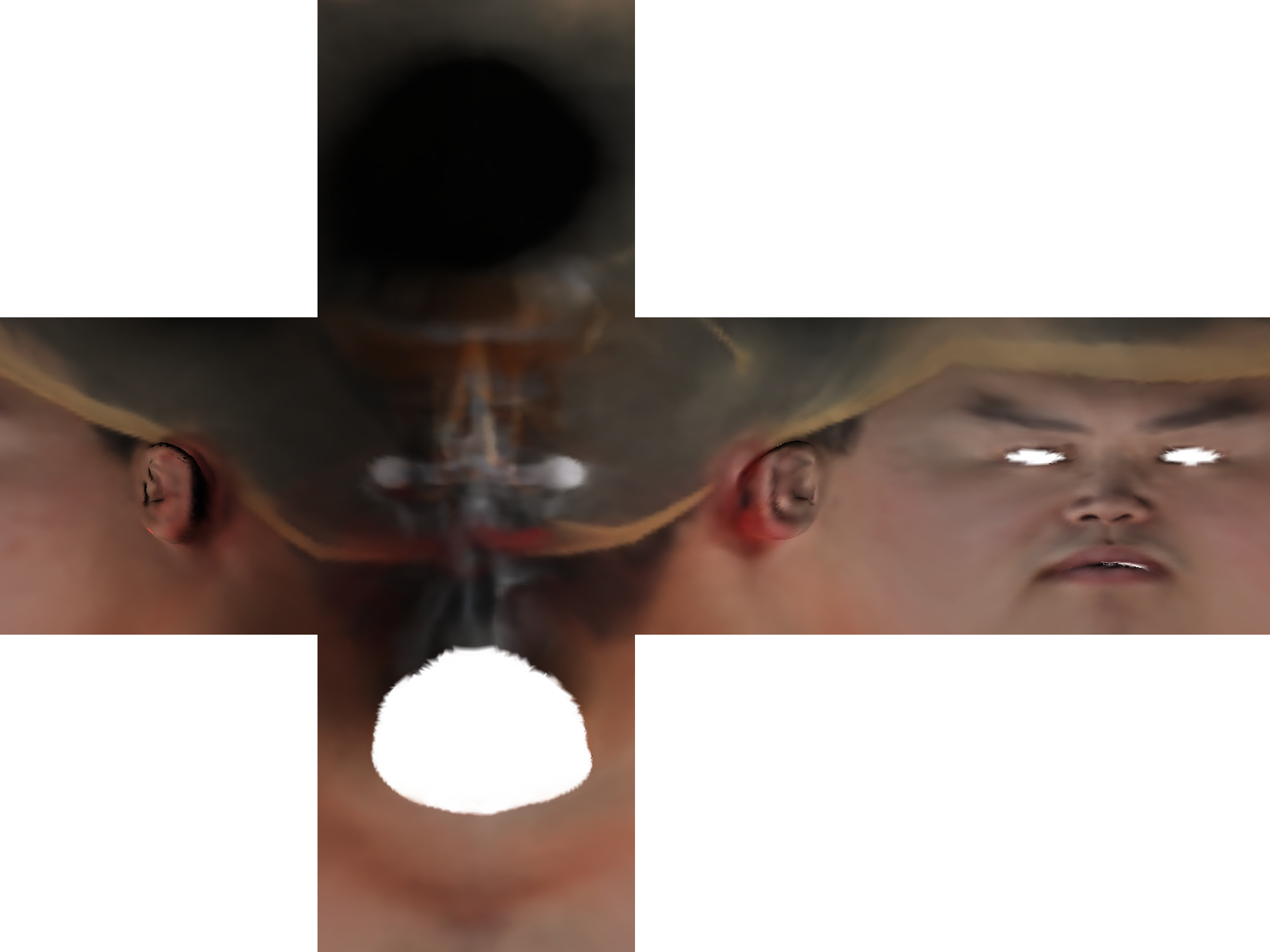}
    \includegraphics[width=0.175\linewidth]{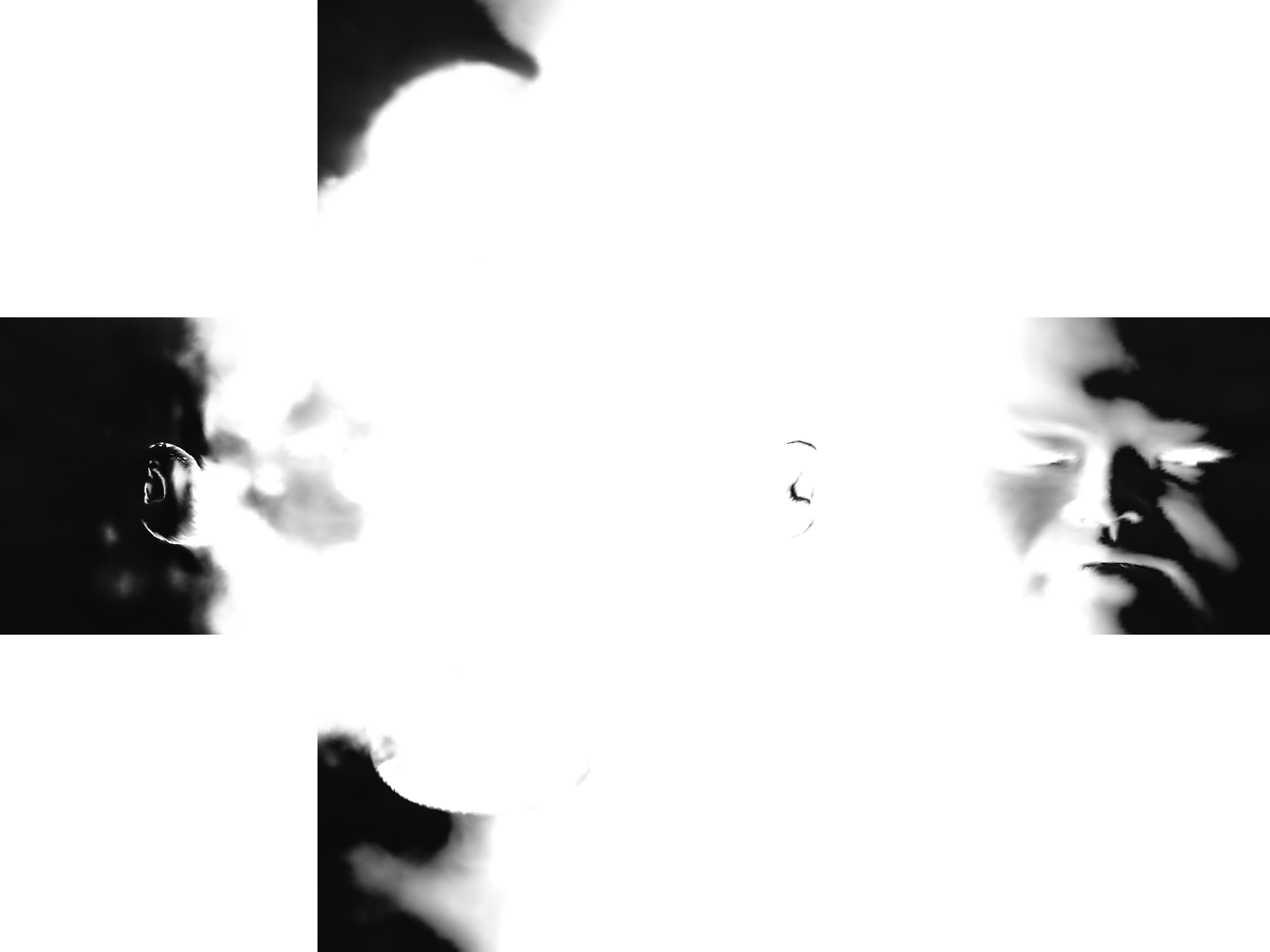}
    \\
    \makebox[0.075\linewidth]{\footnotesize GT}
    \makebox[0.175\linewidth]{\footnotesize Material}
    \makebox[0.175\linewidth]{\footnotesize Shading}
    \makebox[0.075\linewidth]{\footnotesize GT}
    \makebox[0.175\linewidth]{\footnotesize Material}
    \makebox[0.175\linewidth]{\footnotesize Shading}
    \\
    (a) We decompose the radiance into view-dependent shadings and view-independent materials. The shading map is visualized from the front view.
    \\
    \includegraphics[width=0.2\linewidth]{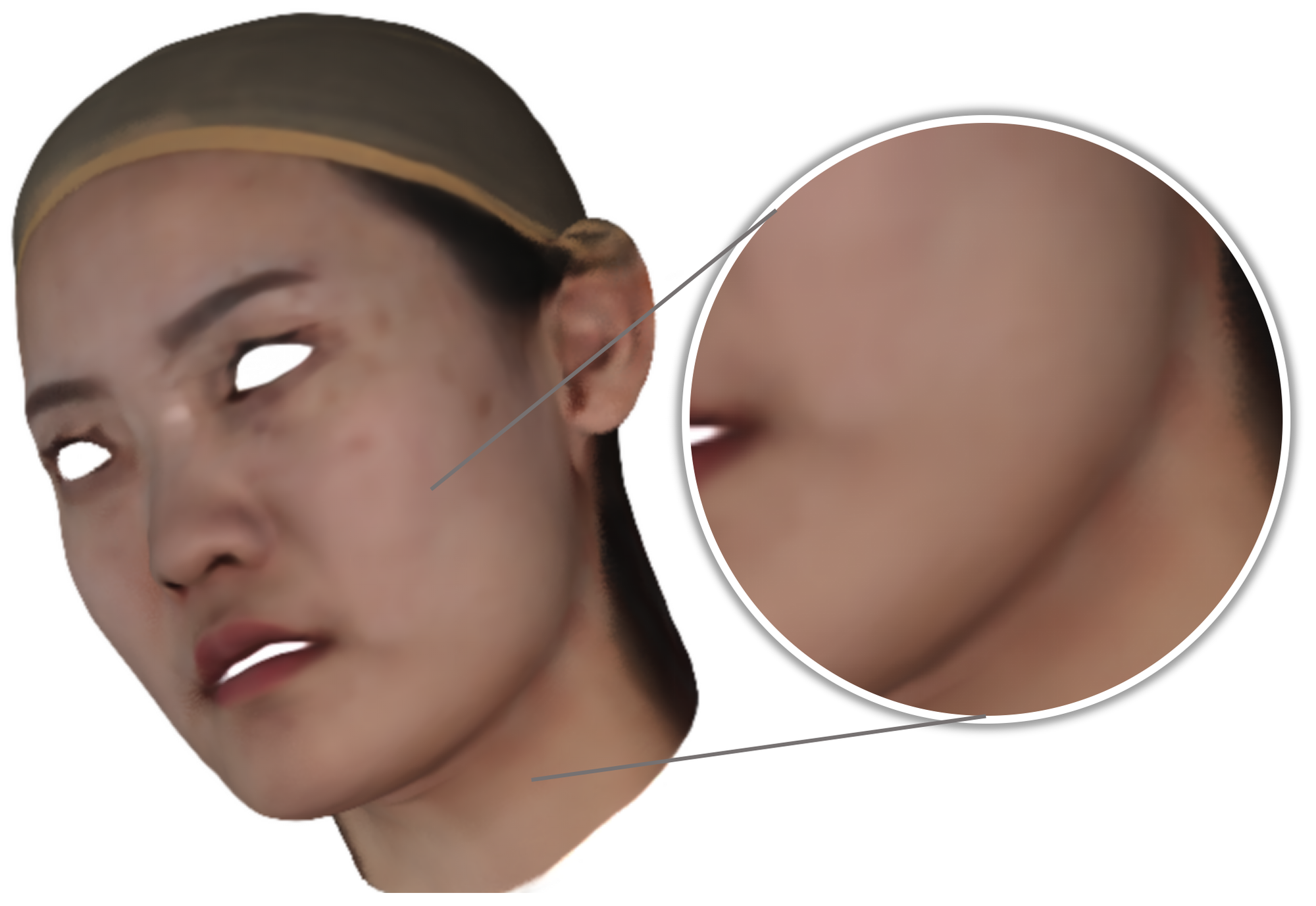}
    \includegraphics[width=0.05\linewidth]{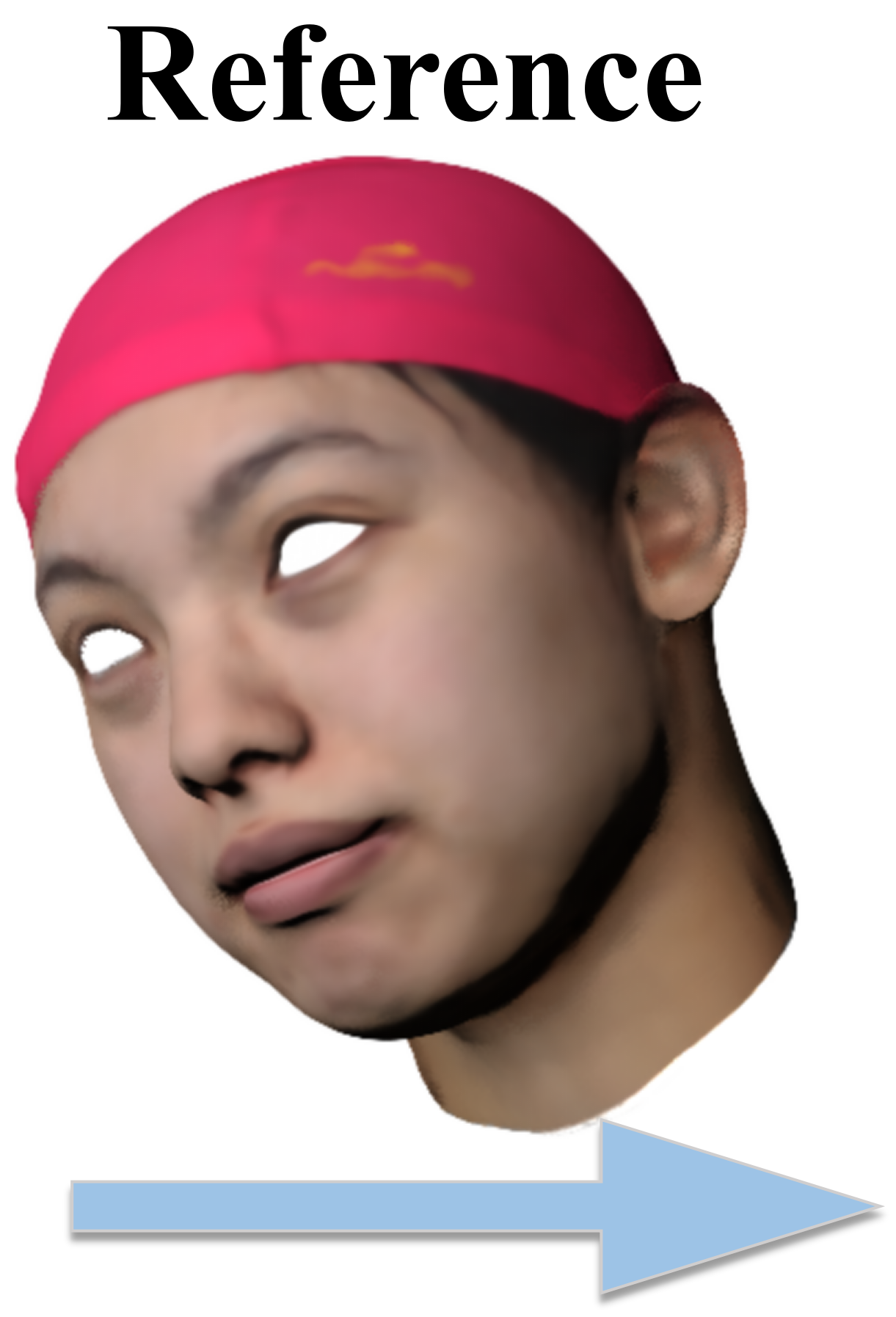}
    \includegraphics[width=0.2\linewidth]{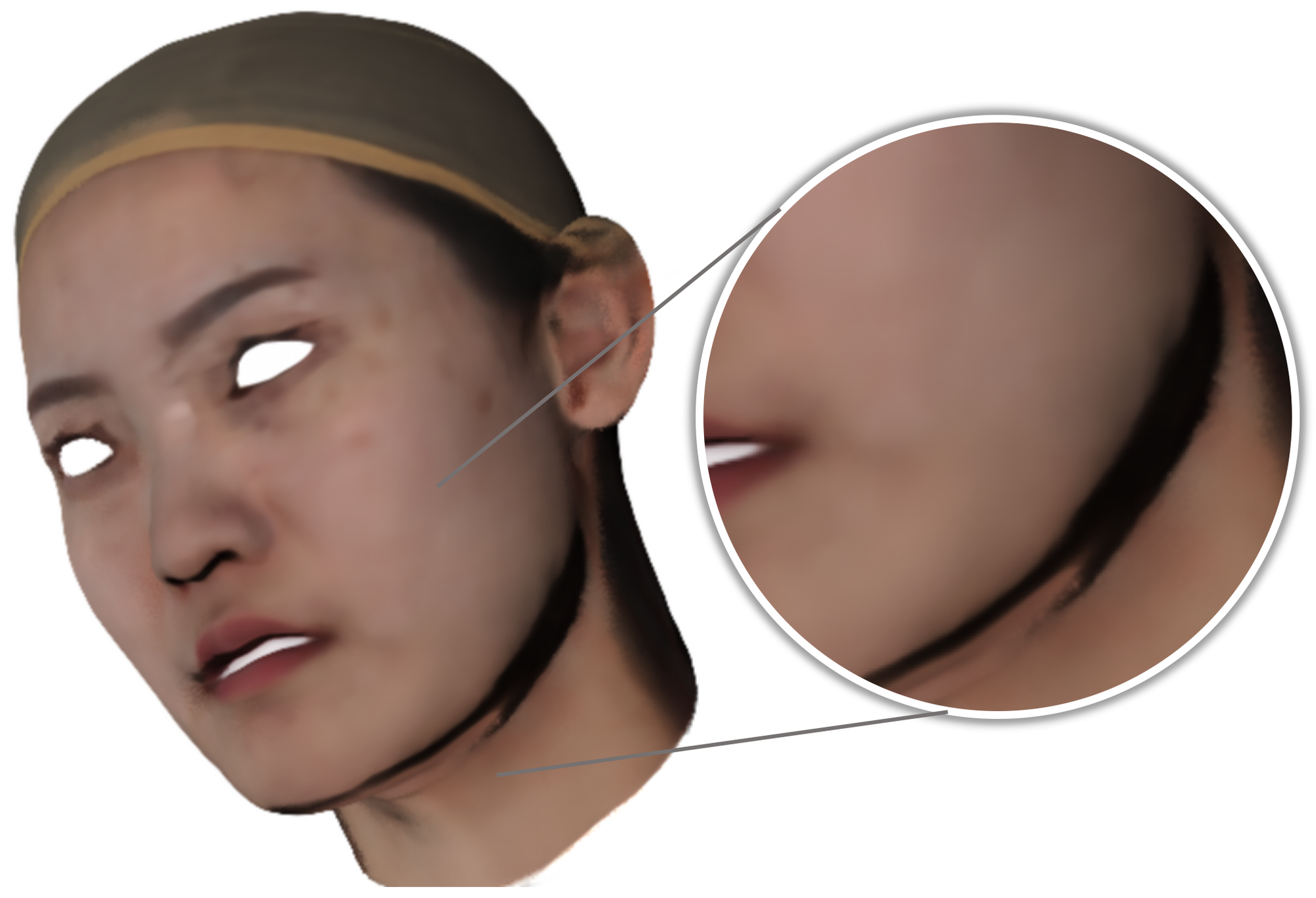}
    \includegraphics[width=0.2\linewidth]{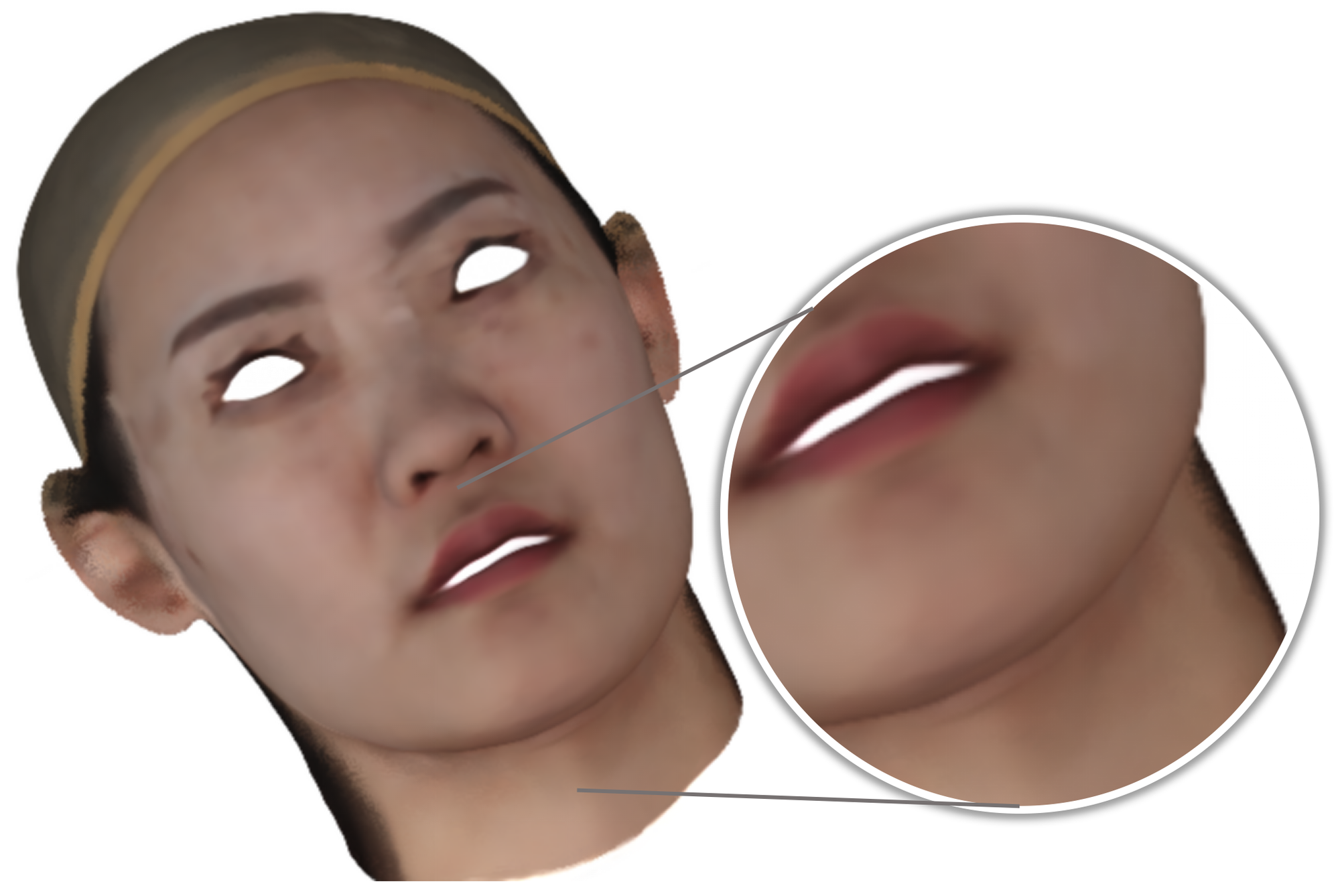}
    \includegraphics[width=0.05\linewidth]{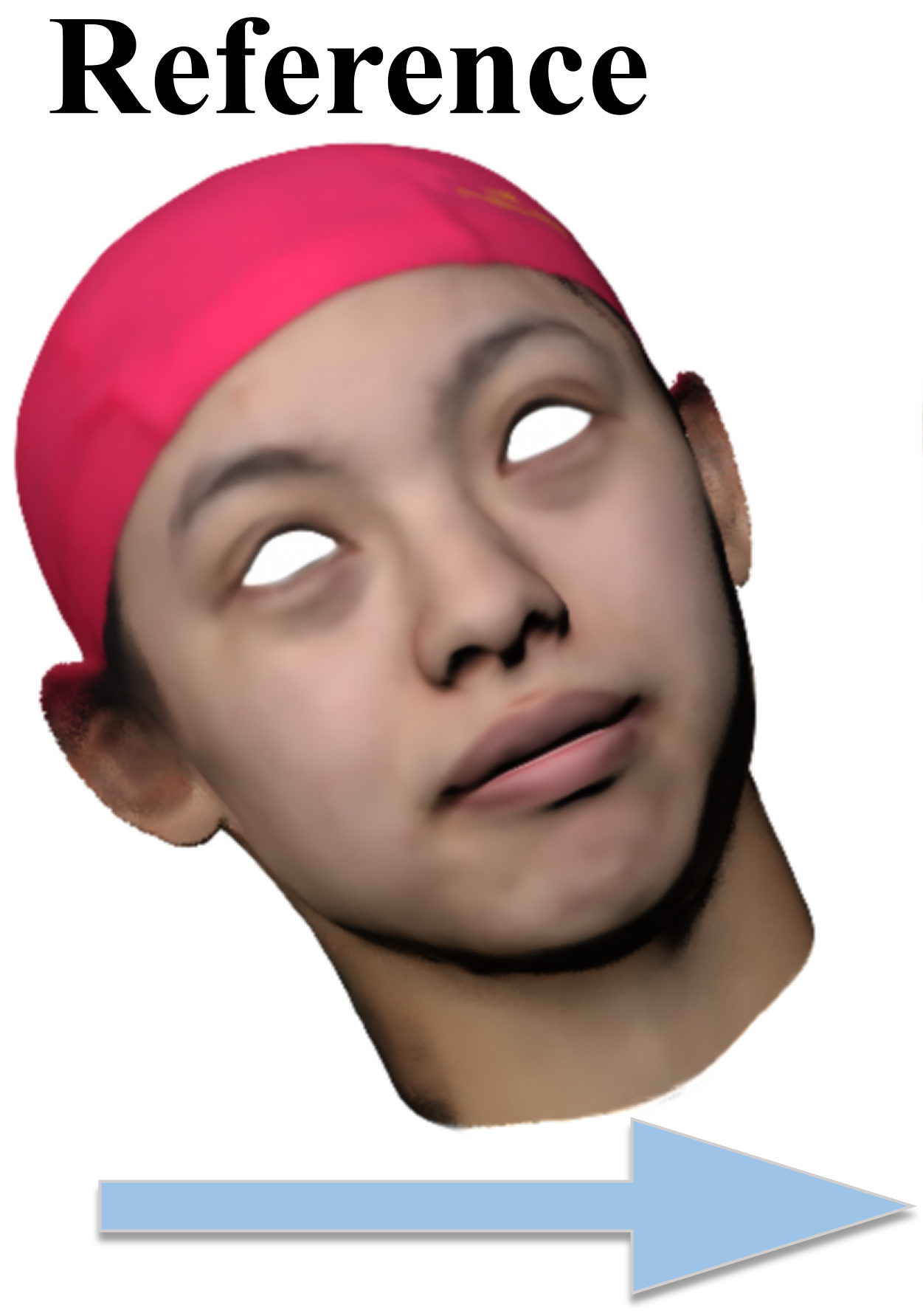}\
    \includegraphics[width=0.2\linewidth]{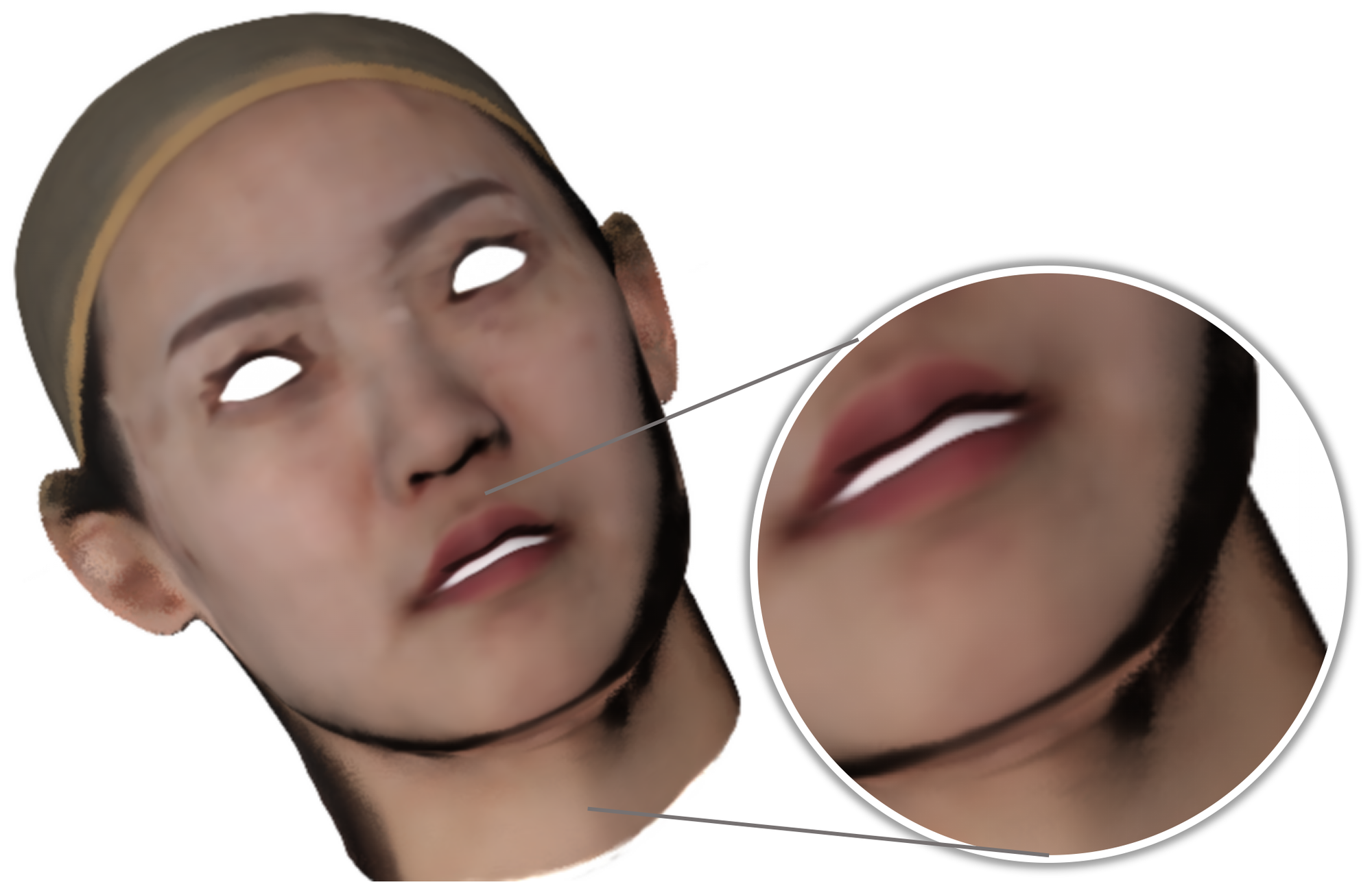}
    \\
    \includegraphics[width=0.2\linewidth]{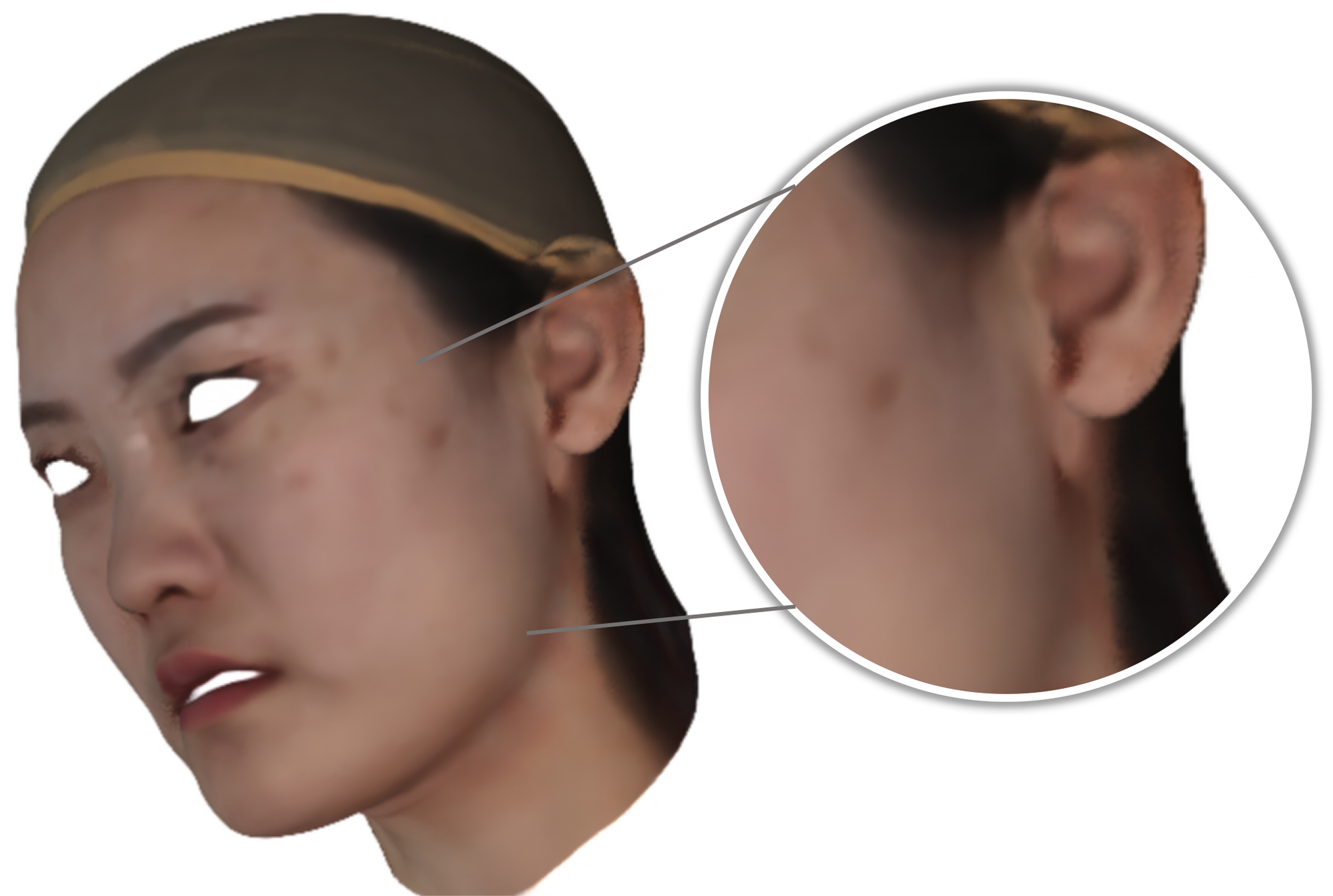}
    \includegraphics[width=0.05\linewidth]{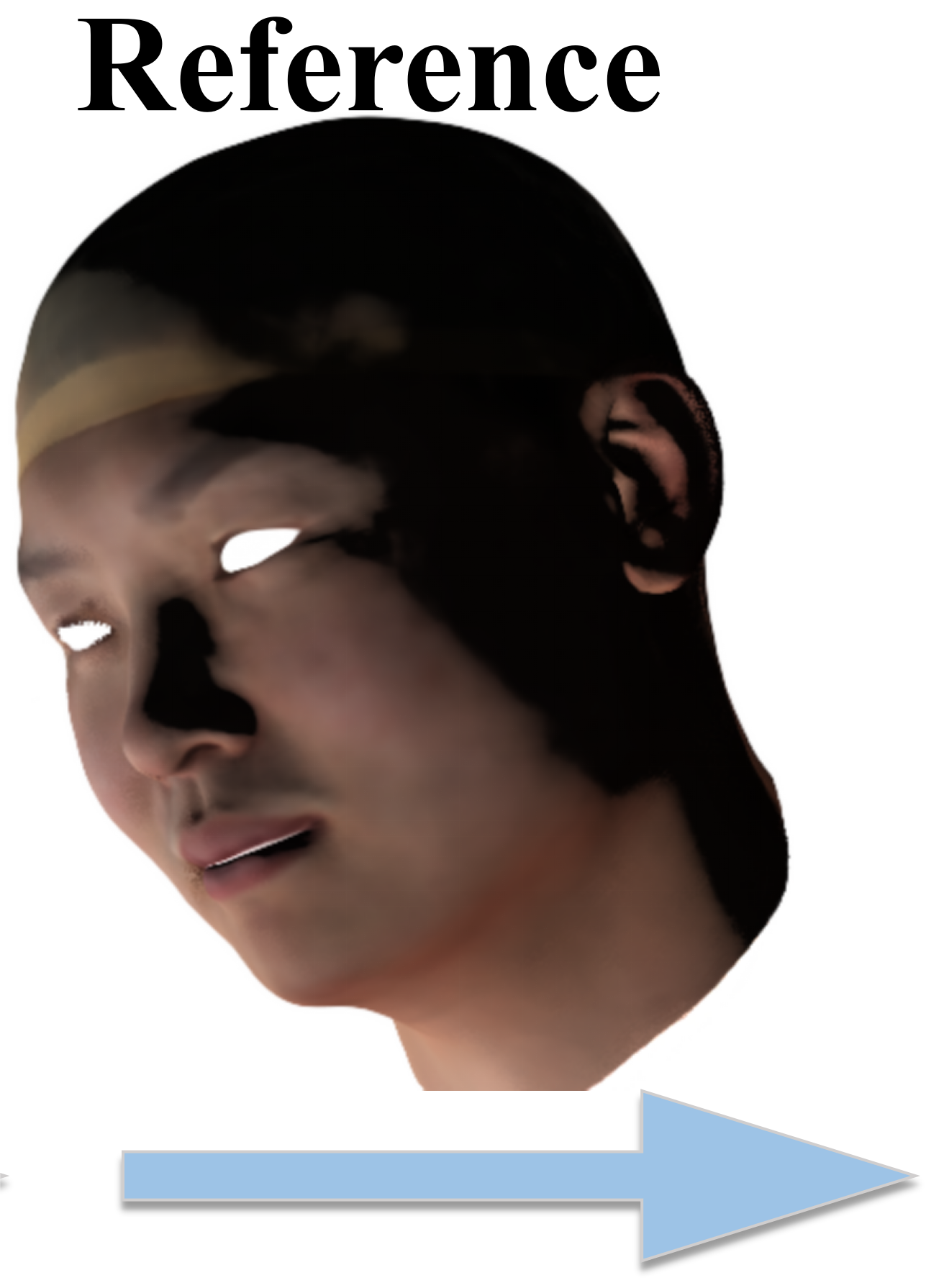}
    \includegraphics[width=0.2\linewidth]{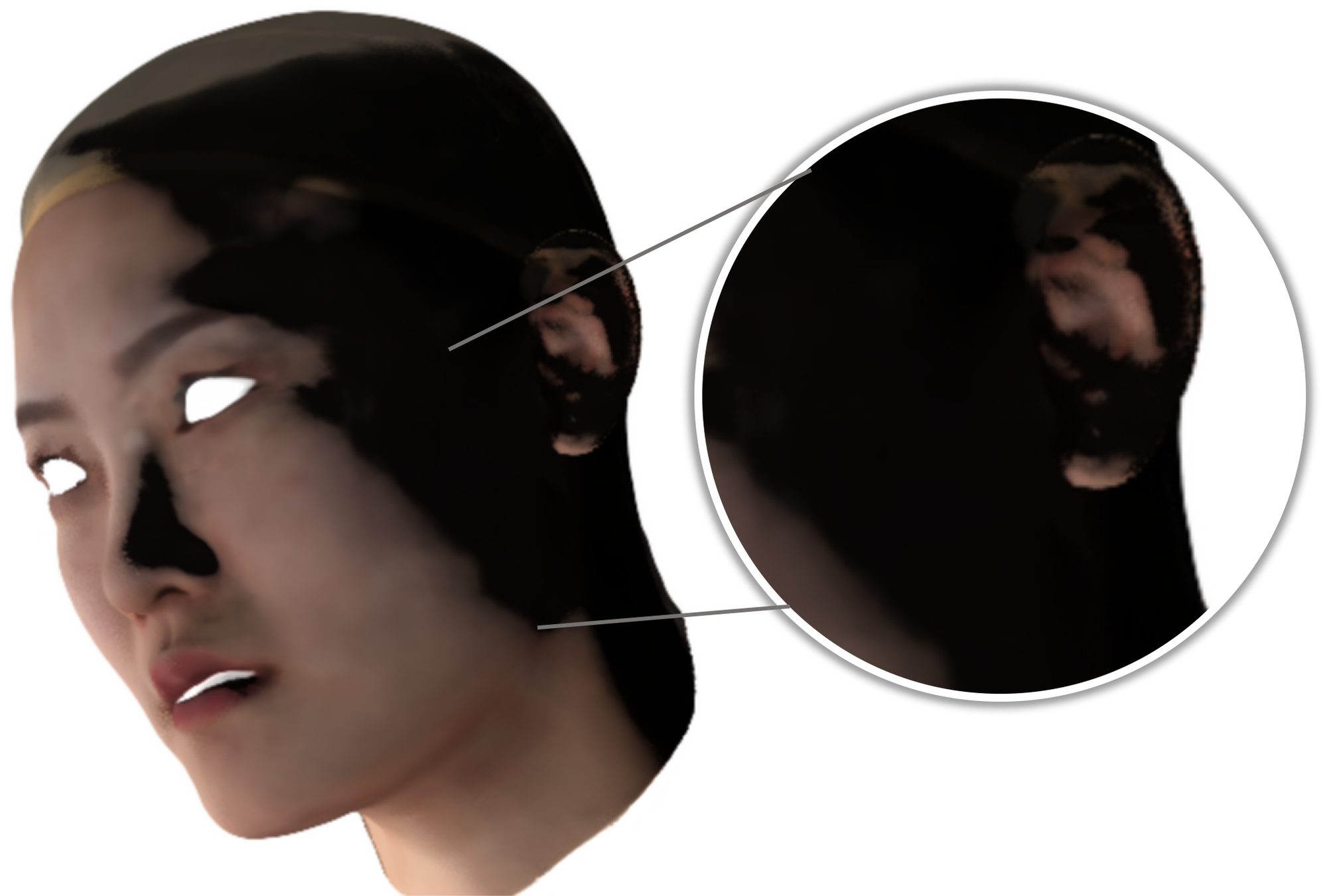}
    \includegraphics[width=0.2\linewidth]{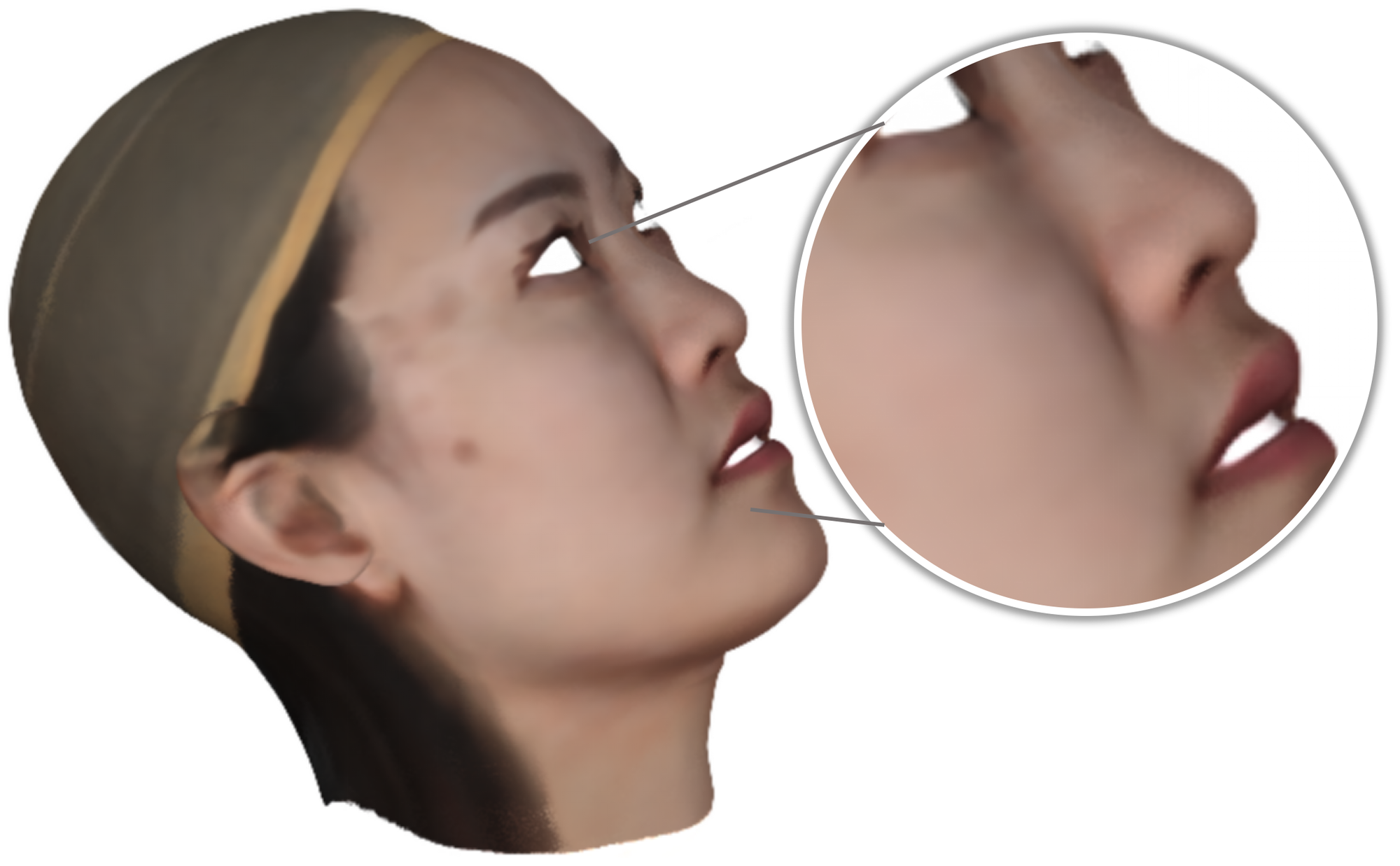}
    \includegraphics[width=0.05\linewidth]{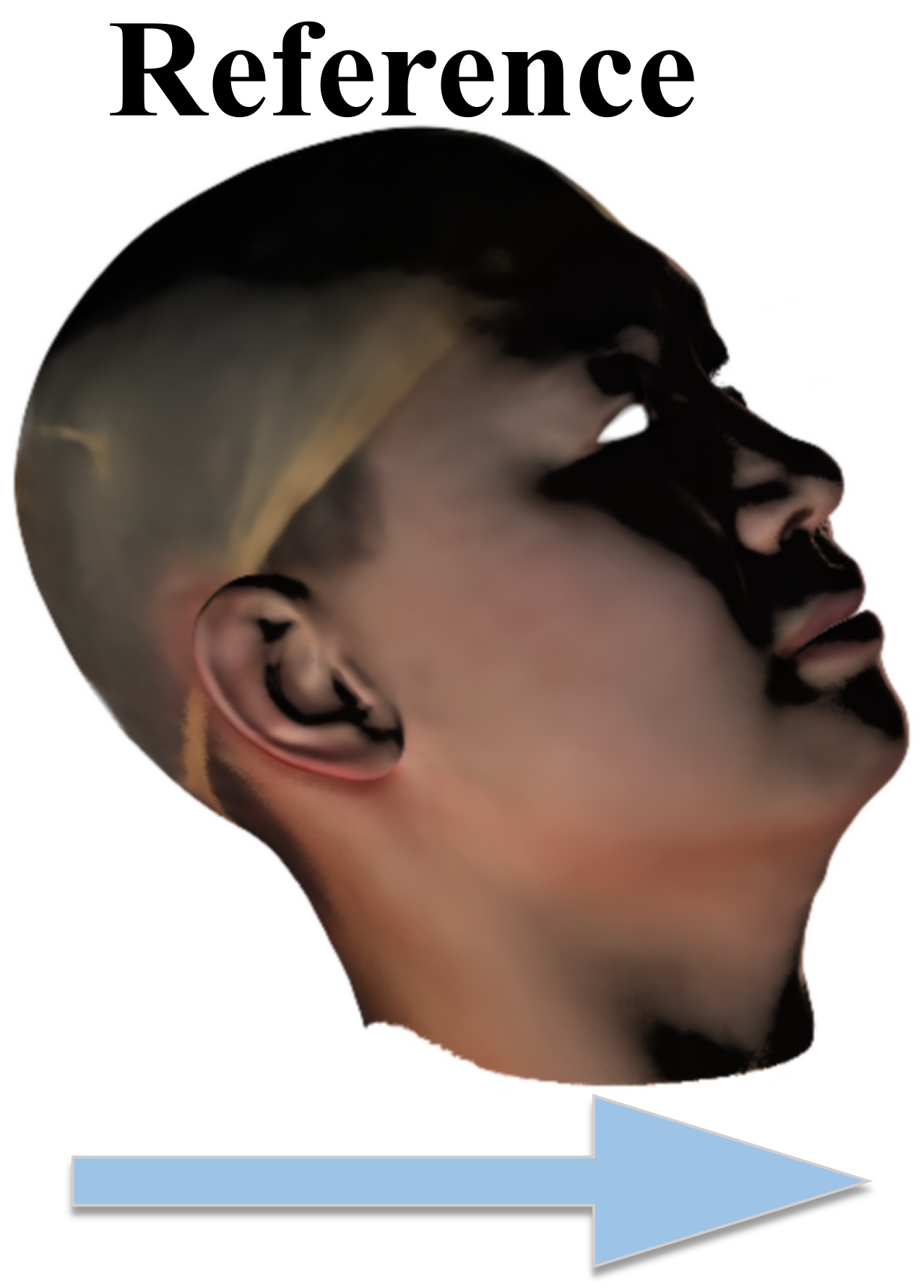}
    \includegraphics[width=0.2\linewidth]{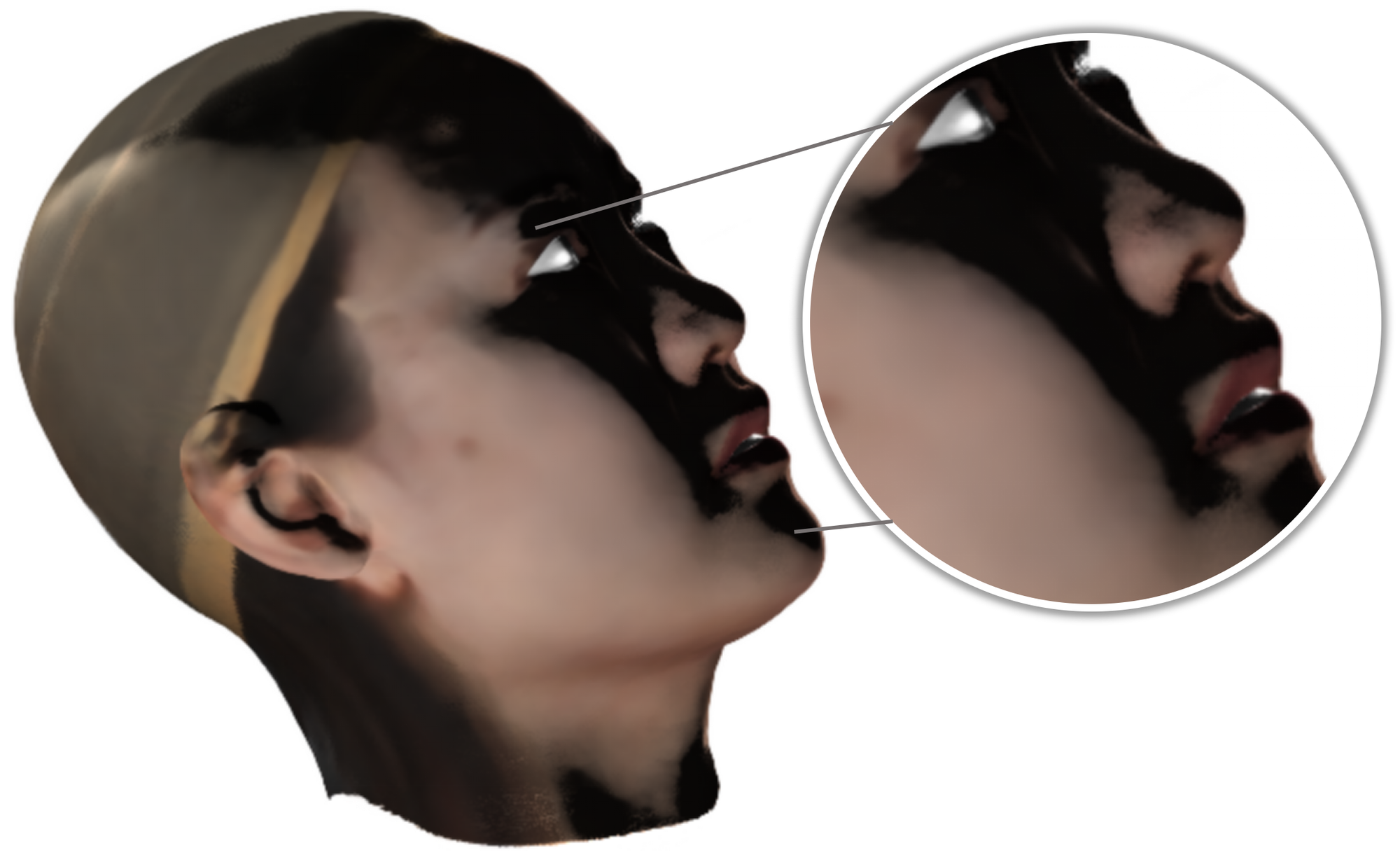}
    \\
    \makebox[0.35\linewidth]{\footnotesize{Before}}
    \makebox[0.1\linewidth]{}
    \makebox[0.35\linewidth]{\footnotesize{After}}
    \\
    (b) We transfer the decomposed shading from one model
to another. See also the accompanying video.
    \caption{We transfer the decomposed shading from one model to another. For further illustration, please refer to the accompanying video.}
    \label{fig:shade_tranfer}
\end{figure}

\textbf{Shading editing.}
We decompose the 3D radiance field into view-independent material and view-dependent shading defined in the parametric domain. Taking advantage of this decomposition and the embedding into a simple domain, we can edit both the material and shading of the 3D surfaces in an intuitive manner. As illustrated in Fig.~\ref{fig:shade_tranfer}(b), we modify the shading of one model by transferring the shading from another model under different lighting conditions, while keeping the material unchanged.

\textbf{Inverse deformation.}
To validate the cycle loss and illustrate inverse deformation, we sample points on the polycube parametric surface $\mathcal{D}$. Subsequently, we inversely deform these points back to the surface $\mathcal{S}$. 
The point cloud is visualized in the right inset, and we quantify the CD ($\times 10^{-3}$) with respect to the ground truth. The obtained low CD results confirm the efficacy of inverse deformation and cycle loss.

\begin{figure}
    \centering
    \begin{minipage}[b]{0.55\linewidth}
    \includegraphics[width=0.23\linewidth]{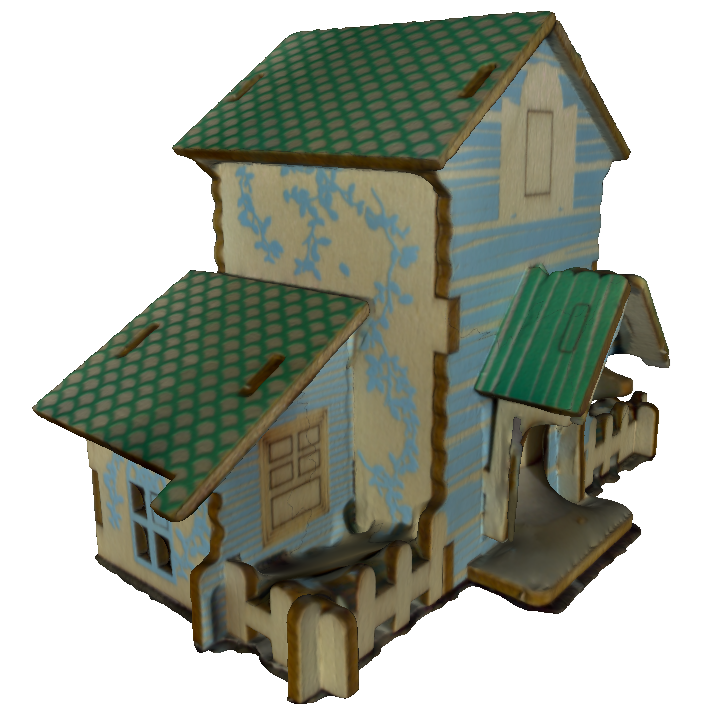}
    \includegraphics[width=0.23\linewidth]{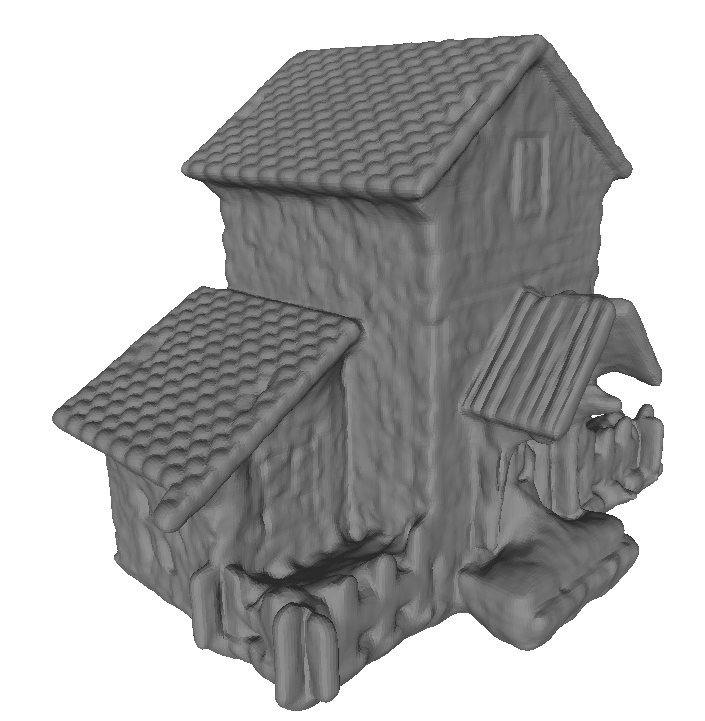}
    \includegraphics[width=0.23\linewidth]{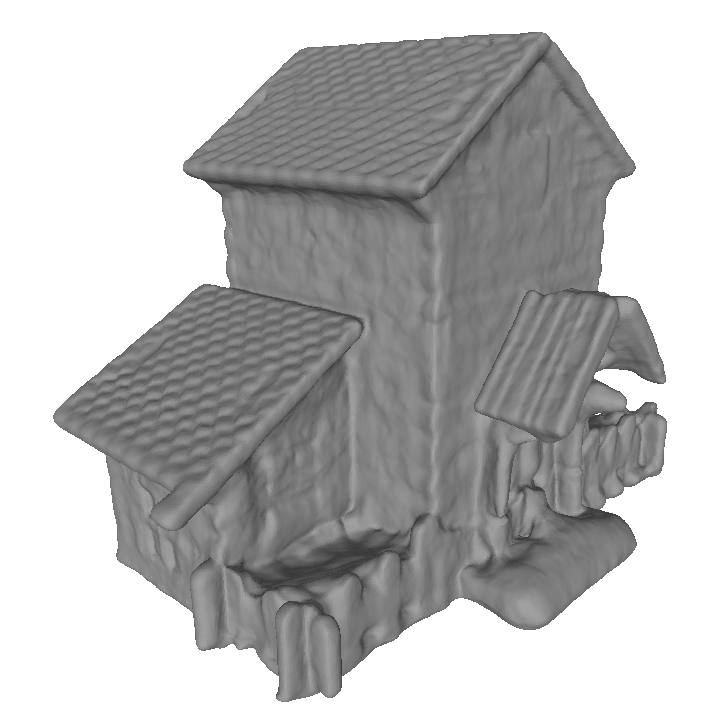} \includegraphics[width=0.23\linewidth]{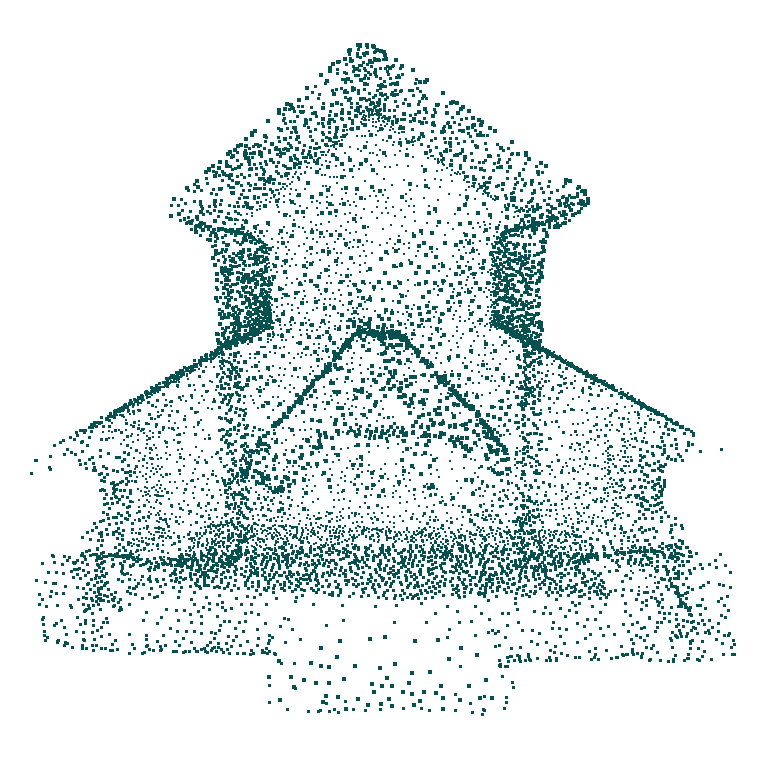}\\
    \makebox[0.23\linewidth]{\tiny GT }
    \makebox[0.23\linewidth]{\tiny VolSDF 4.65}
    \makebox[0.23\linewidth]{\tiny Ours 4.87}
    \makebox[0.23\linewidth]{\tiny Inverse 4.89}
    \\
    \includegraphics[width=0.23\linewidth]{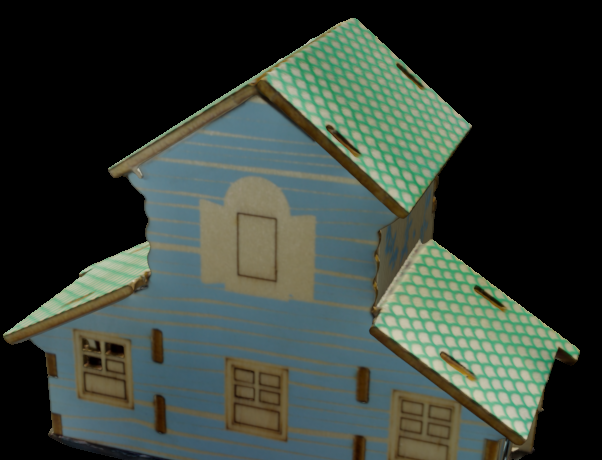}
    \includegraphics[width=0.27\linewidth]{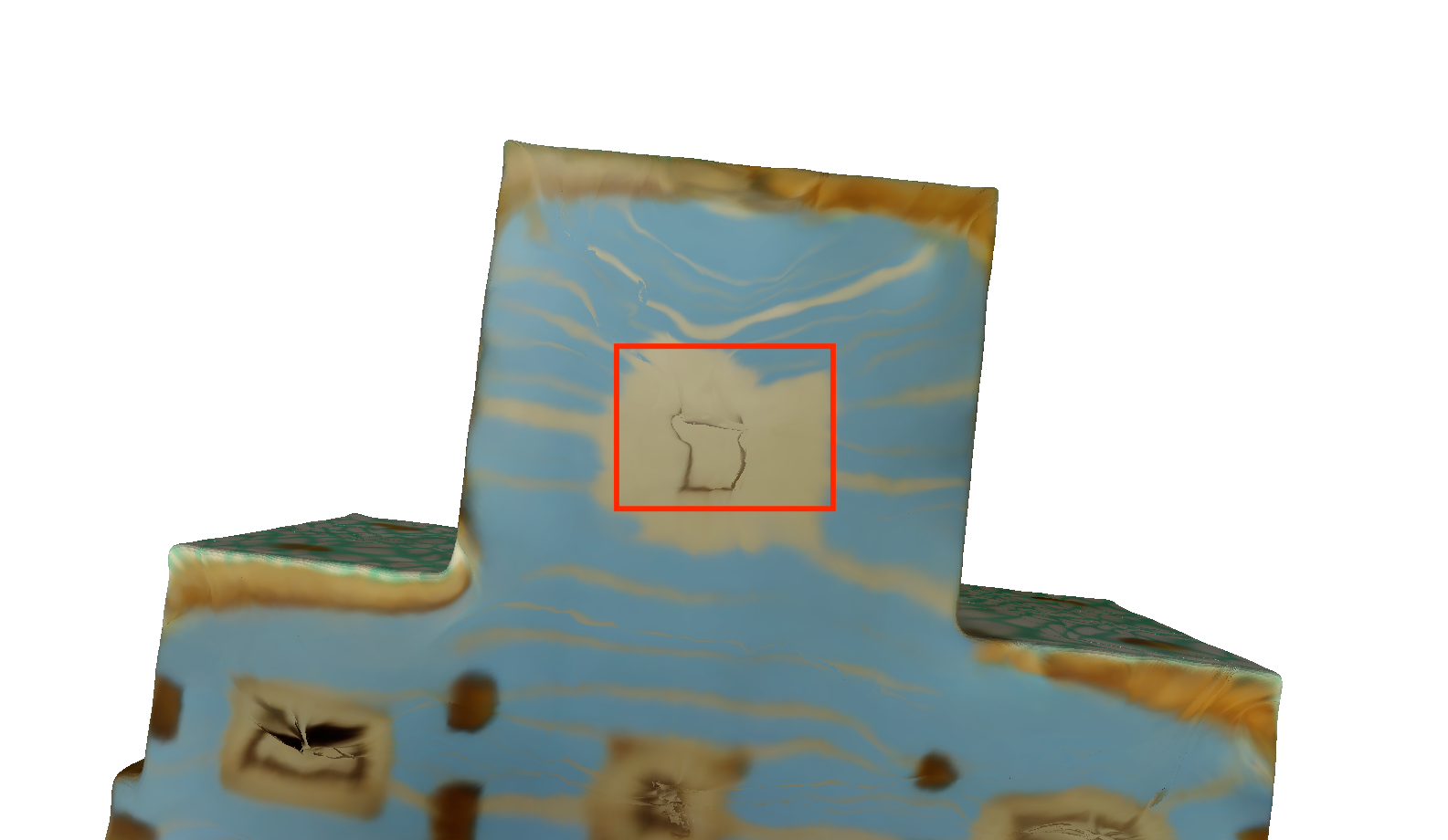}
    \includegraphics[width=0.23\linewidth]{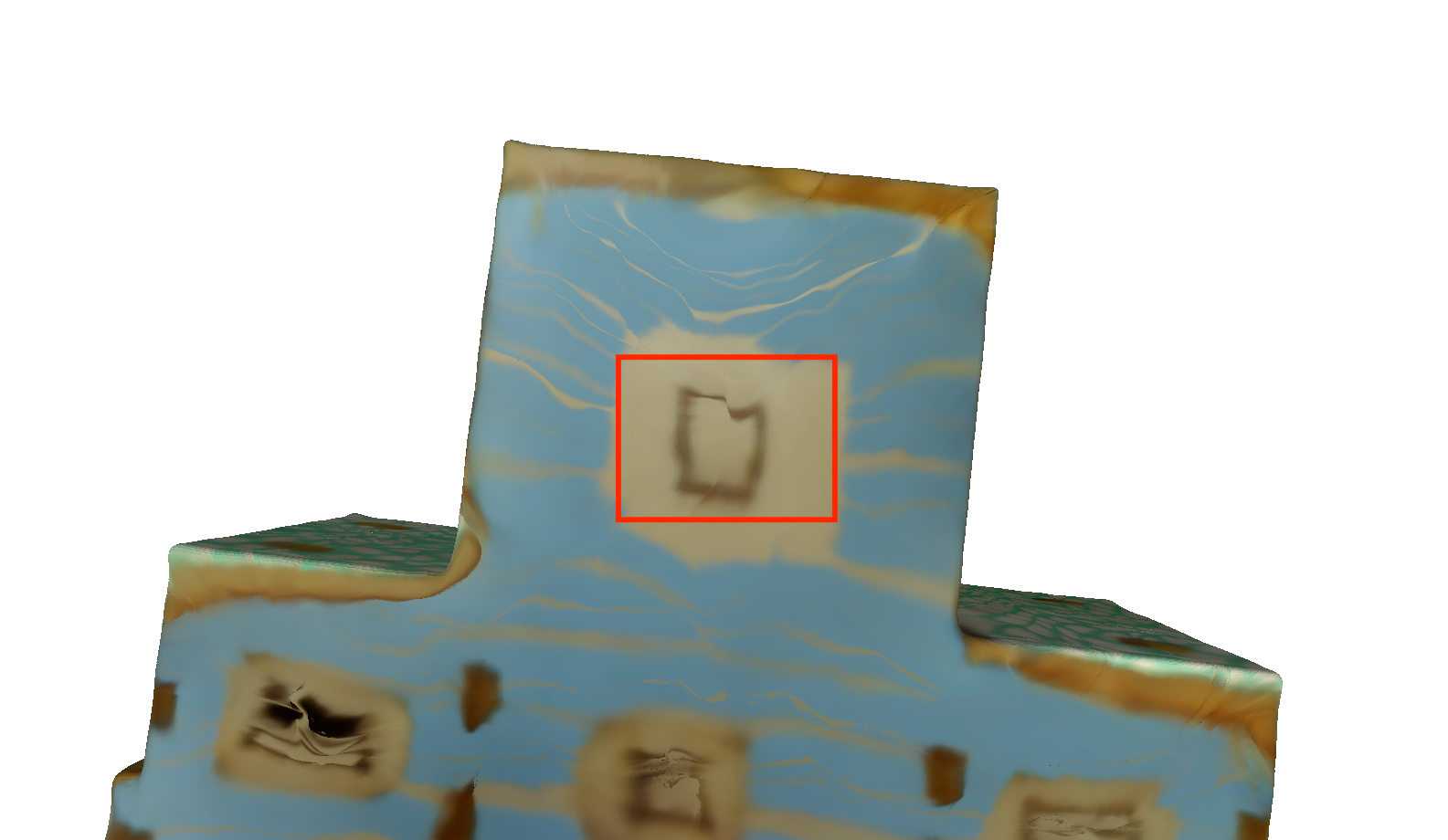}
    \includegraphics[width=0.23\linewidth]{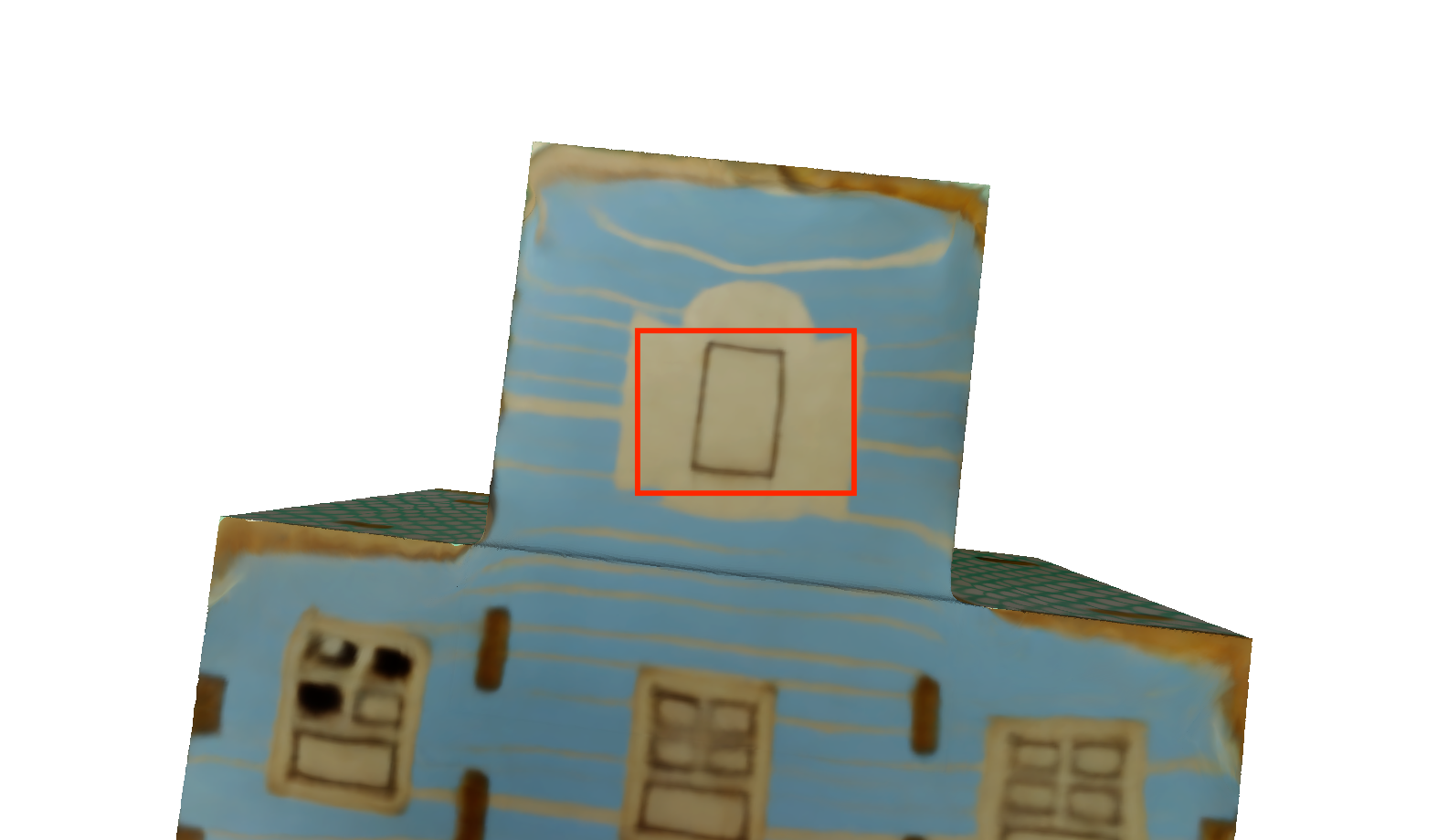} 
    \\
    \makebox[0.23\linewidth]{\scriptsize Input}
    \makebox[0.27\linewidth]{\tiny 
    w/o $\mathcal{L}_\text{smooth} \& \mathcal{L}_\text{Lap}$}
    \makebox[0.23\linewidth]{\scriptsize
    $+\mathcal{L}_\text{smooth}$}
    \makebox[0.23\linewidth]{\scriptsize 
    $+\mathcal{L}_\text{Lap}$}
    \end{minipage}
    \begin{minipage}[b]{0.4\linewidth}
        \includegraphics[width=0.92\linewidth]{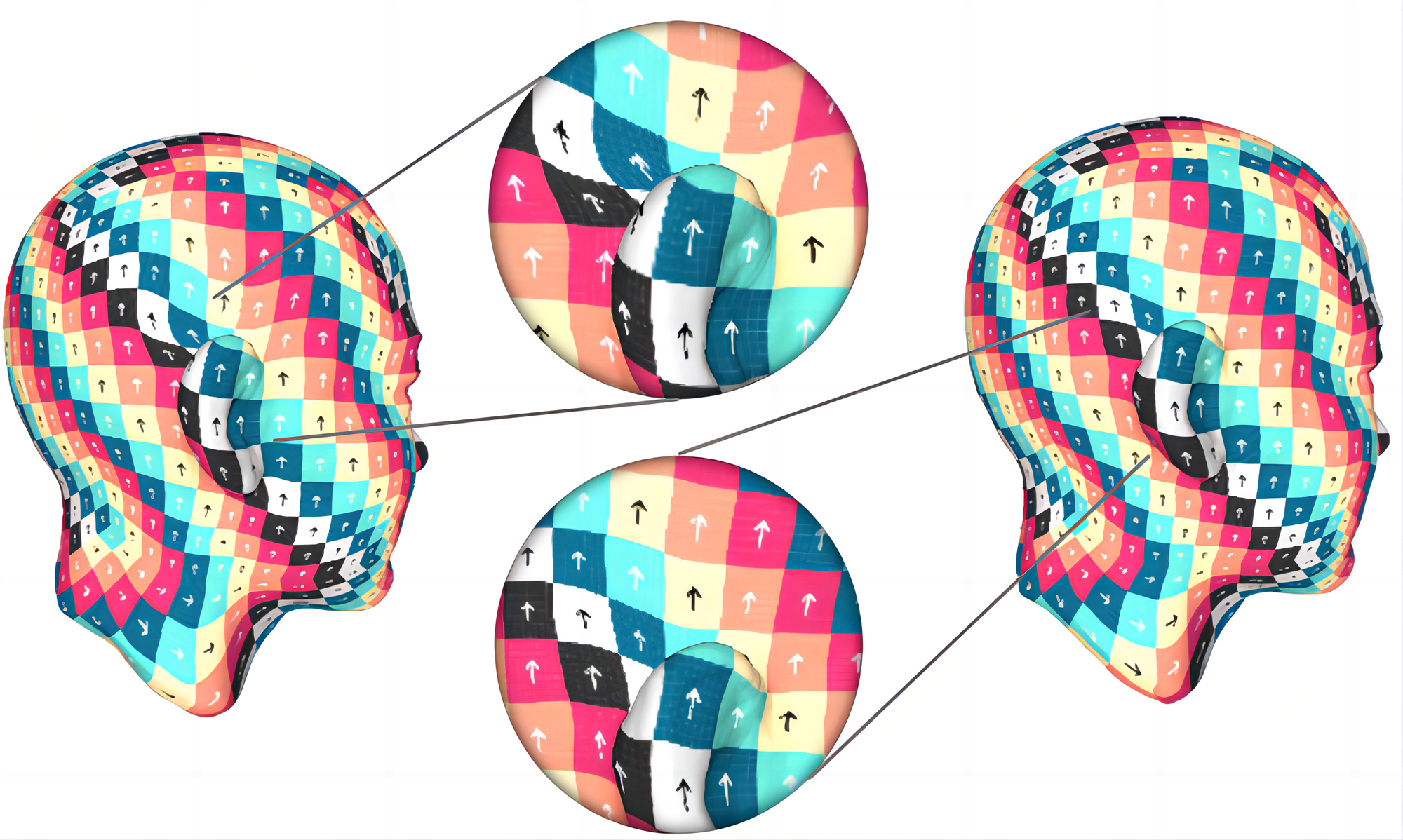}
        \\
        \makebox[0.4\linewidth]{\scriptsize{$\mathcal{E}_{\text{angle}}=1.147$}}\hspace{10pt}
        \makebox[0.4\linewidth]{\scriptsize{$\mathcal{E}_{\text{angle}}=\mathbf{1.092}$}}
        \\
        \makebox[0.4\linewidth]{\footnotesize w/o $\mathcal{L}_{\text{Lap}}$}\hspace{10pt}
        \makebox[0.4\linewidth]{\footnotesize w/ $\mathcal{L}_{\text{Lap}}$}
        
    \end{minipage}
    \caption{Ablation studies. On the left, we calculate the CD ($\times 10^{-3}$) based on the point cloud extracted from the inverse deformation module and present the ablation studies for $\mathcal{L}_\text{smooth}$ and $\mathcal{L}_\text{Lap}$. On the right, utilization of the Laplacian loss has proven effective in reducing angle distortion ($\mathcal{E}_\text{angle}$), especially in regions with high curvature.}
    \label{fig:ablation-para}
\end{figure}

\textbf{Ablation studies.}
As shown in Fig.~\ref{fig:ablation-para} (left), the term $\mathcal{L}_\text{smooth}$ helps to smooth the process of the two deformations,  while $\mathcal{L}_\text{Lap}$ further reduces the angle distortion. Our method can effectively reduce angle distortion by minimizing Laplacian loss and reduce area distortion by choosing an appropriate parametric domain. As demonstrated in Fig.~\ref{fig:ablation-para} (right), we calculate the angle distortion $\mathcal{E}_{\text{angle}}$ of parameterization as described in~\cite{degener2003adaptable, he2009divide}. We can observe that the Laplacian loss can effectively reduce angle distortion. 

\textbf{The choice of parametric domain.}
The different parametric domain influences the quality of parameterization as shown in Fig.~\ref{fig:comp-volsdf} (a), which is measured by area distortions. Generally, increasing the number of cubic primitives for better resembling the geometry of the target surface effectively reduces distortions but increases the complexity of the parametric domain. Finding an optimal balance between domain complexity and parameterization quality is challenging. 

\section{Conclusion}
We introduce a novel neural algorithm for parameterizing 3D surfaces represented by neural implicit functions to a user-defined parametric domain, such as spheres and polycubes. This parametric domain is learned given some user-specified hyperparameters. Utilizing bi-directional deformation, our method is capable of learning a nearly bijective mapping without relying on any prior knowledge, while controlling angle distortion through the usage of a Laplacian regularizer. Furthermore, our approach seamlessly integrates with the neural rendering pipeline, enabling the reconstruction of 3D objects from multi-view images and facilitating efficient volume rendering of modified textures and shadings — eliminating the necessity for network re-training. We demonstrated the efficacy of our method on human heads and man-made objects. 

\begin{figure}[htbp]
    \centering
    \begin{minipage}{0.45\linewidth}
    \includegraphics[width=0.23\linewidth]{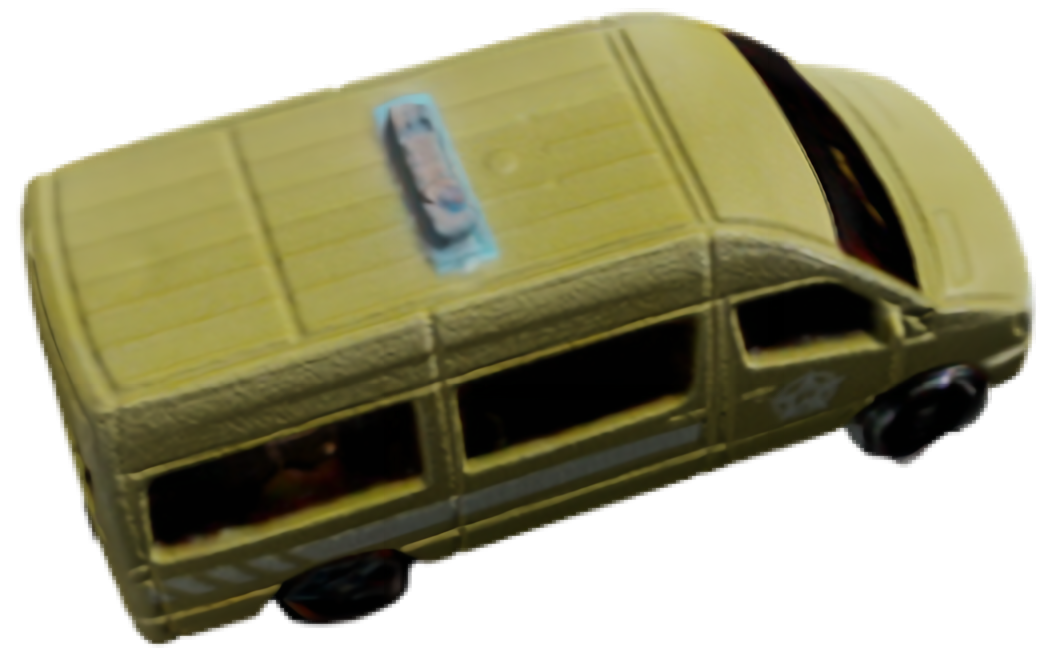}
    \includegraphics[width=0.23\linewidth]{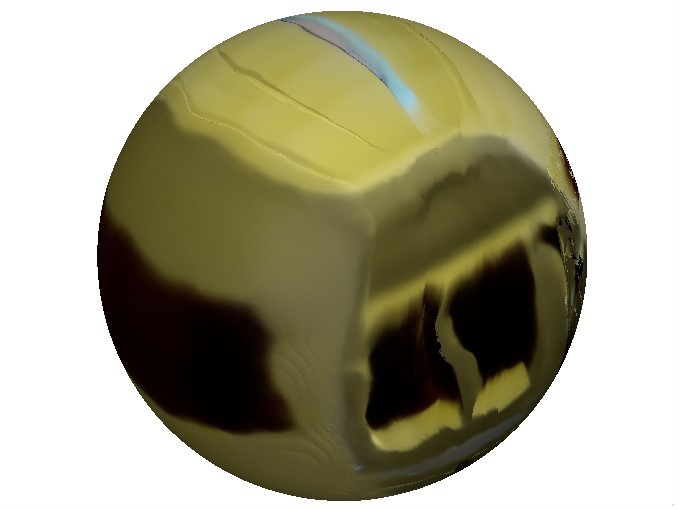}
    \includegraphics[width=0.23\linewidth]{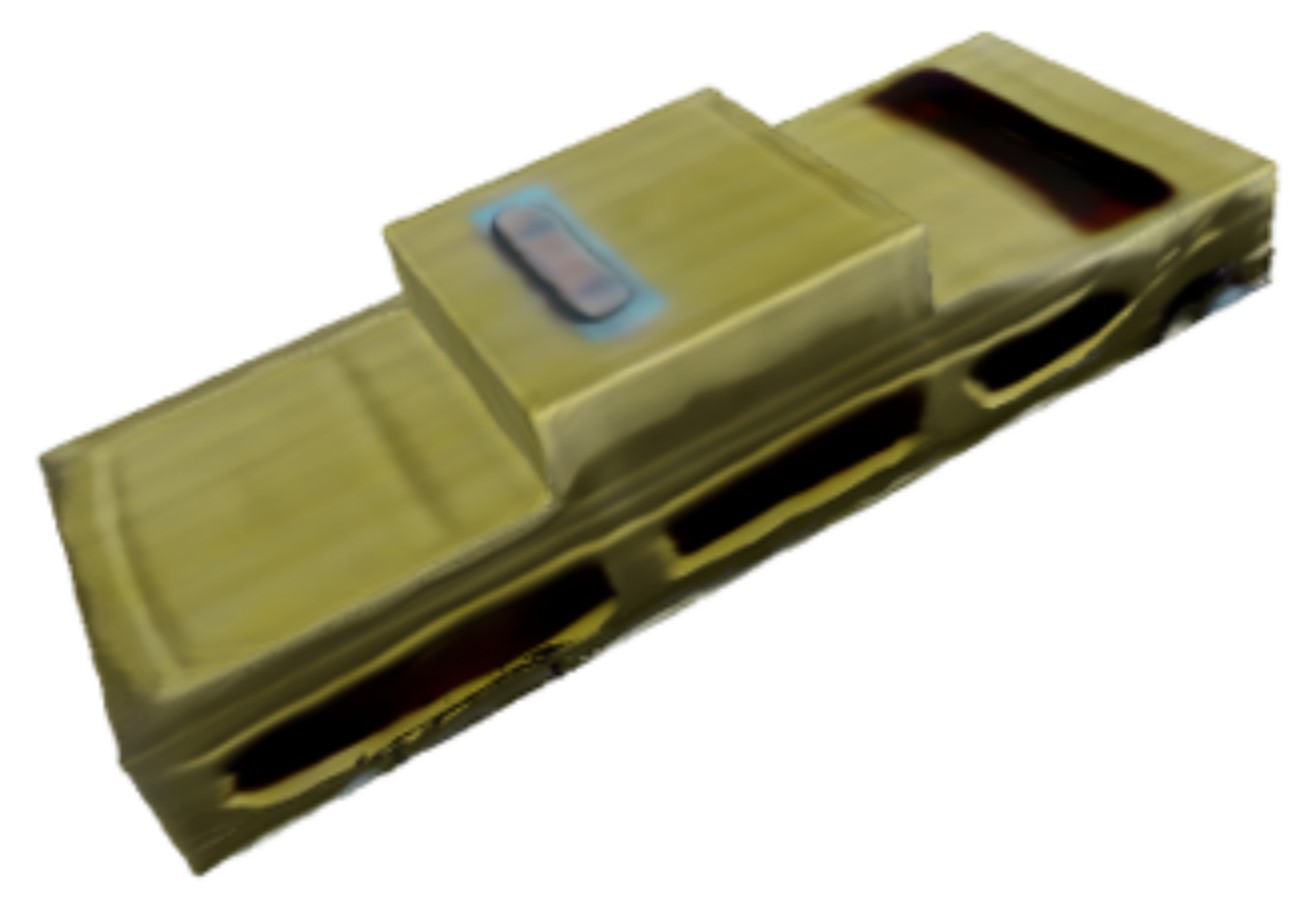}
    \includegraphics[width=0.23\linewidth]{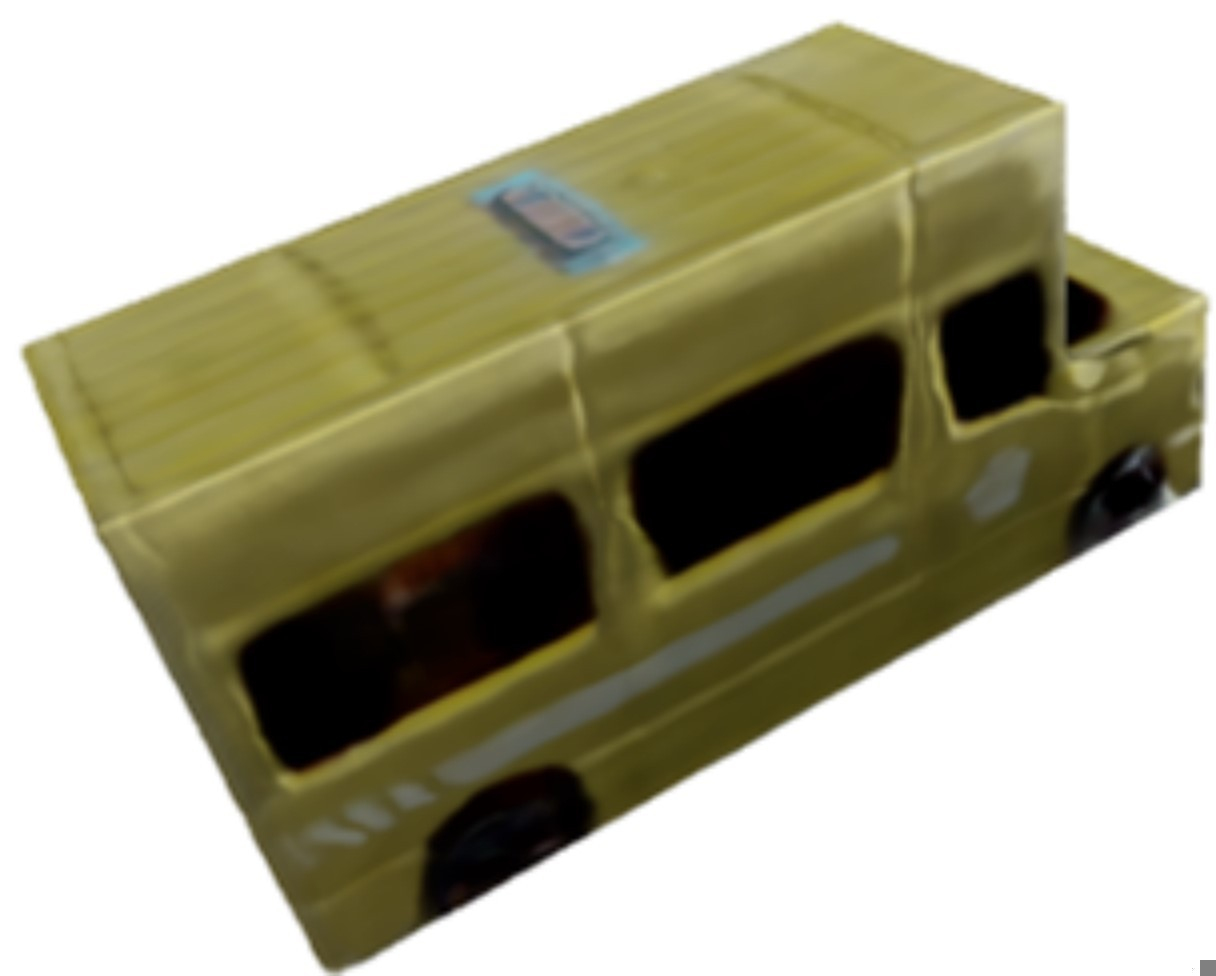}
    \\
    \makebox[0.23\linewidth]{\scriptsize $\mathcal{S}$}
    \makebox[0.23\linewidth]{\scriptsize $\mathcal{E}_\text{area}=
    2.505$}
    \makebox[0.23\linewidth]{\scriptsize $6.094$}
    \makebox[0.23\linewidth]{\scriptsize $1.633$}
    \end{minipage}
    \begin{minipage}{0.3\linewidth}
    \includegraphics[width=0.18\linewidth]{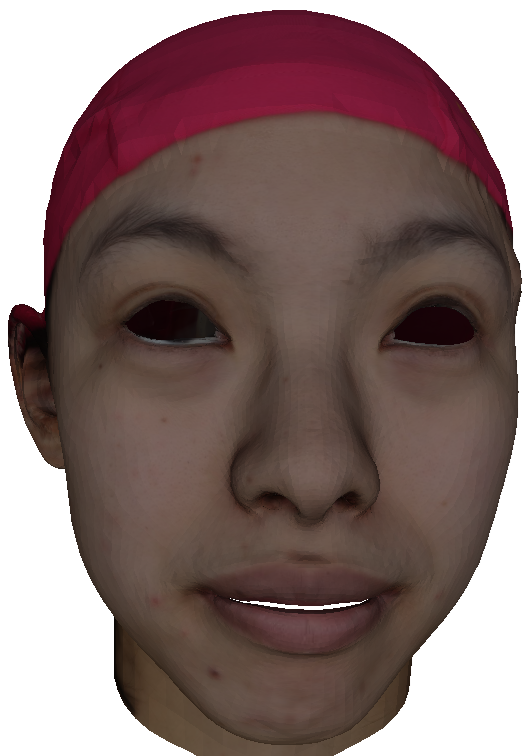}
    \includegraphics[width=0.18\linewidth]{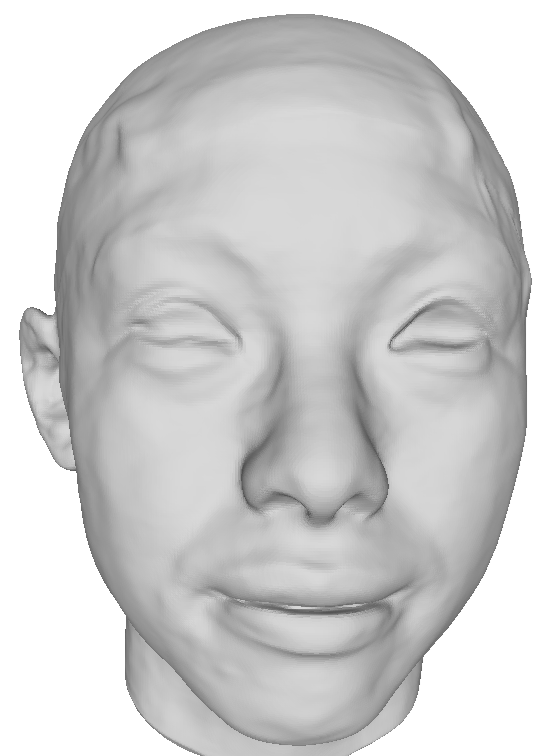}
    \includegraphics[width=0.18\linewidth]{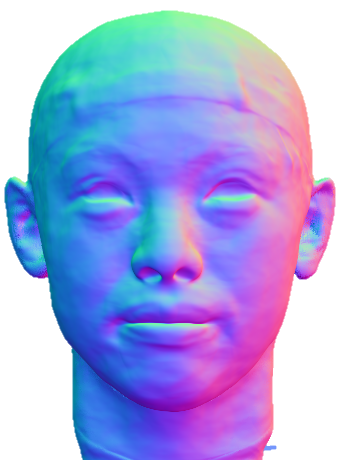}
    \includegraphics[width=0.18\linewidth]{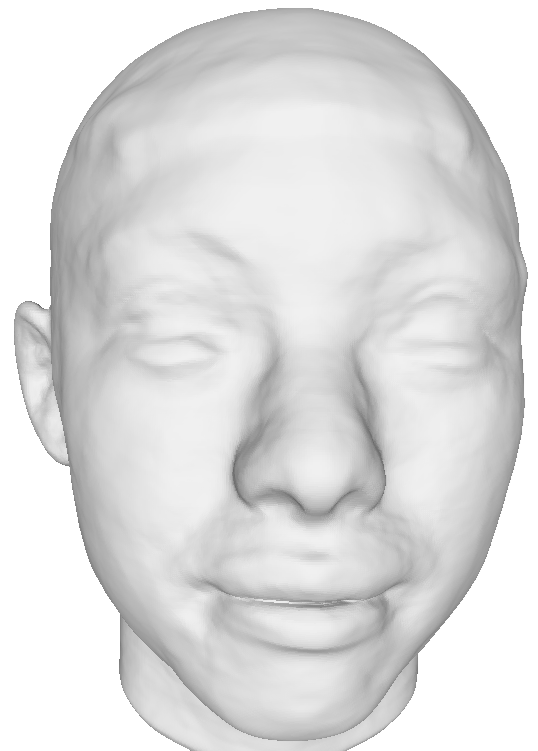}
    \includegraphics[width=0.18\linewidth]{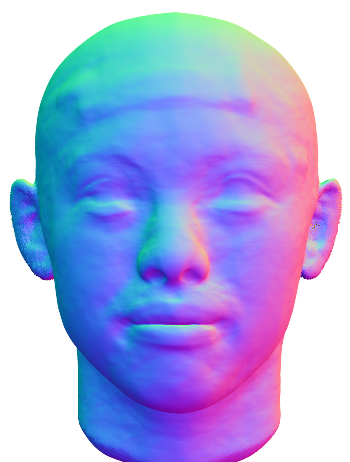}
    \makebox[0.18\linewidth]{\footnotesize GT}
    \makebox[0.36\linewidth]{\footnotesize VolSDF}
    \makebox[0.36\linewidth]{\footnotesize Ours}
    \end{minipage}
    \begin{minipage}{0.2\linewidth}
        \resizebox{0.95\linewidth}{!}{
\begin{tabular}{ccc}\\\toprule
Method &  PSNR & CD($10^{-3}$) \\ \midrule
VolSDF &31.15 & 3.270\\  \midrule
Ours &30.87 & 3.423\\ \bottomrule
\end{tabular}}
\label{tab:comp_volsdf}
    \end{minipage}
    \\
    \makebox[0.4\linewidth]{(a)}
    \makebox[0.5\linewidth]{(b)}
    \caption{We present area distortion $\mathcal{E}_\text{area}$ based on the different choices of the parametric domain in (a) and geometry reconstruction results of VolSDF~\cite{yariv2021volume} and our method in (b). Our method is capable of editing through parameterization, accompanied by an acceptable compromise in reconstruction quality. }
    \label{fig:comp-volsdf}
\end{figure}

Our method has a few limitations that require further improvement in the future. First, quantitative comparisons on the FS-Syn dataset show that the reconstruction quality of our method is slightly worse than the standard SDF-based neural rendering algorithms, such as VolSDF~\cite{yariv2021volume} and NeuS~\cite{wang2021neus}. 
For example, as illustrated in Fig.~\ref{fig:comp-volsdf} (b) and, We report PSNR and CD compared with VolSDF~\cite{yariv2021volume} on Fs-Syn dataset~\cite{yang2020facescape}.
This decline in visual quality arises from the incorporation of additional modules, such as bi-directional deformation and radiance decomposition, resulting in the cessation of updates to the SDF network within our neural parameterization pipeline. Therefore, computing a parameterization without compromising the reconstruction quality is highly desired. Second, our radiance decomposition operates under the assumptions of Lambertian reflectance and grayscale shading, which might hinder its effectiveness when dealing with intricate materials or sophisticated shadings. 

\section*{Acknowledgment}
This study is supported under the RIE2020 Industry Alignment Fund – Industry Collaboration Projects (IAF-ICP) Funding Initiative, as well as cash and in-kind contribution from the industry partner(s). This project is also partially supported by the Ministry of Education, Singapore, under its Academic Research Fund Grants (MOE-T2EP20220-0005 \& RT19/22).
\bibliographystyle{splncs04}
\bibliography{main}

\setcounter{page}{1}

\appendix
\section{Implementation Details and Ablation Studies}
\label{appendix:details}
In our framework, the bi-deformation networks $F_{\text{def}}$ and $F_{\text{inv-def}}$, the material network $F_{\text{mat}}$, and the shading network $F_{\text{shd}}$ are all multilayer perceptrons, consisting of 8, 8, 4, and 4 layers, respectively. Each of these hidden layers has 256 neurons. 
The latent shape codes $z_s$ and the appearance codes $z_a$ are 128-dimensional. 
Within the neural rendering process, we set the frequency of position encoding for positions and viewpoints as $6$ and $4$,  respectively. We trained our network using the Adam optimizer~\cite{kingma2014adam}. In each iteration, we randomly sample 1024 pixels from each input image for training. As for other parameter settings related to neural field reconstruction, we refer mainly to the work of VolSDF~\cite{yariv2021volume}.

\begin{figure}[htbp]
    \centering
    \includegraphics[width=0.45\linewidth]{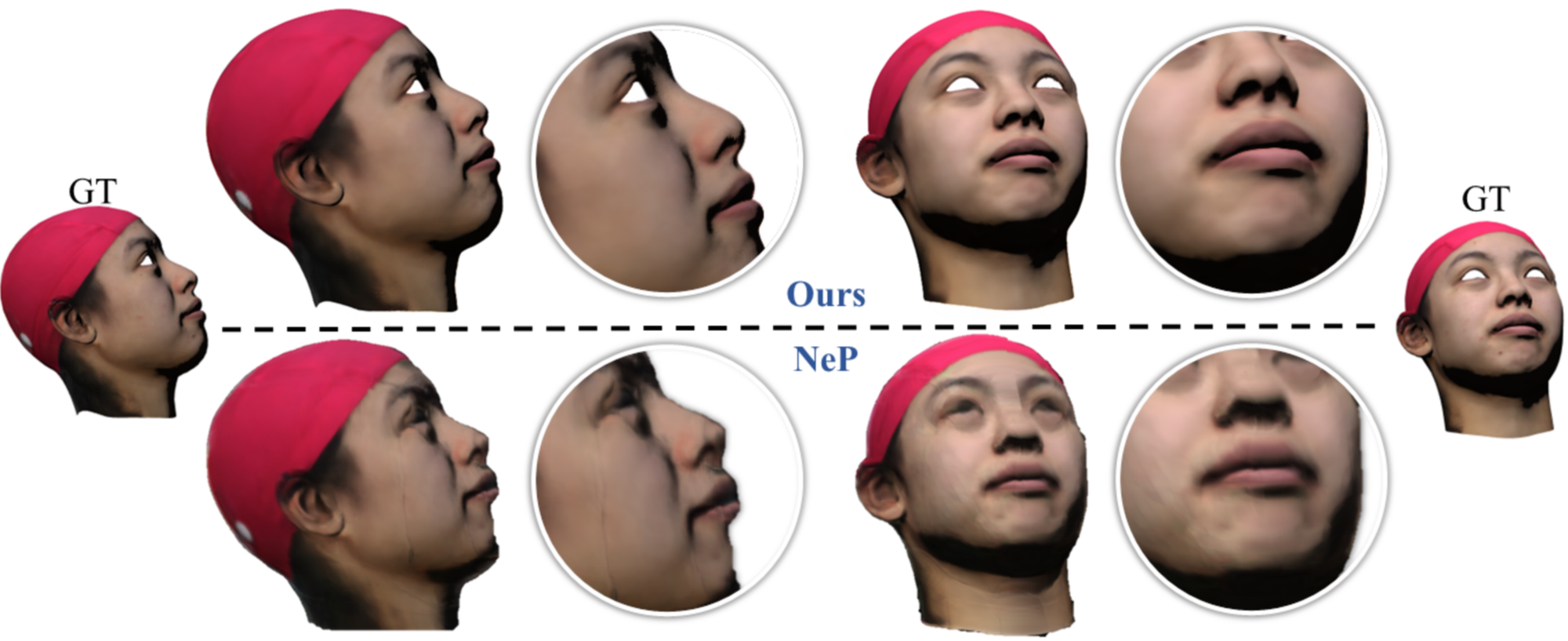}
    \includegraphics[width=0.45\linewidth]{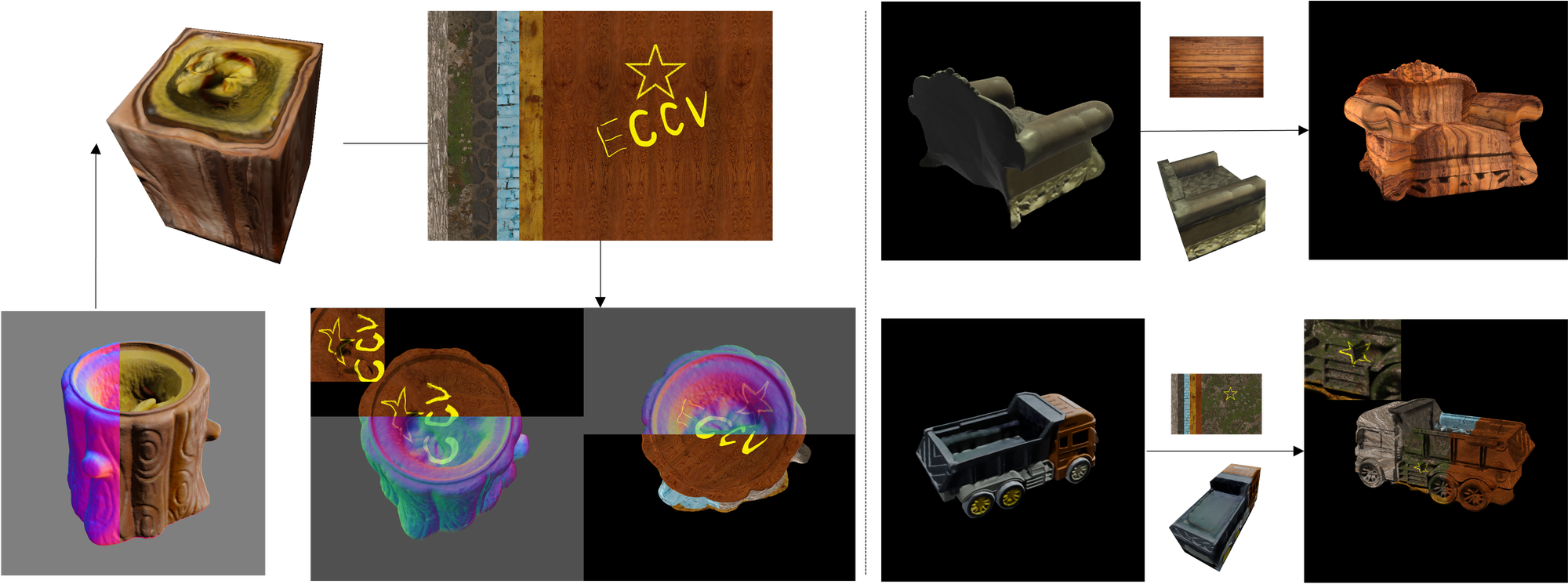}
    \caption{(Left) We compare the rendering results with NeP without prior information from UV mapping. (Right) We show texture editing results demonstrating that our method offers a more intuitive editing approach, capable of modifying the material without changing the shading.}
    \label{fig:compare-nep}
\end{figure}

\paragraph{Baseline implementation details.}
We remove the warp module in the NeP~\cite{ma2022neural} for static settings. Then we train NeP under two settings, one is under UV reference as supervision and the other is not (see Fig.~\ref{fig:compare-nep}). The parameter configurations for other settings can be found in the original paper~\cite{ma2022neural}.

\begin{figure}
    \centering
    \begin{minipage}{0.6\linewidth}
    \includegraphics[width=0.23\linewidth]{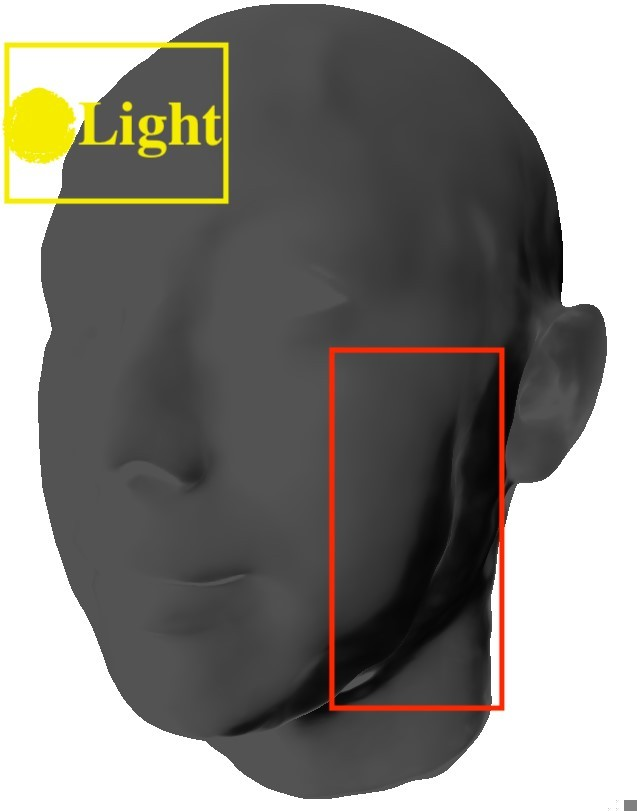}
    \includegraphics[width=0.23\linewidth]{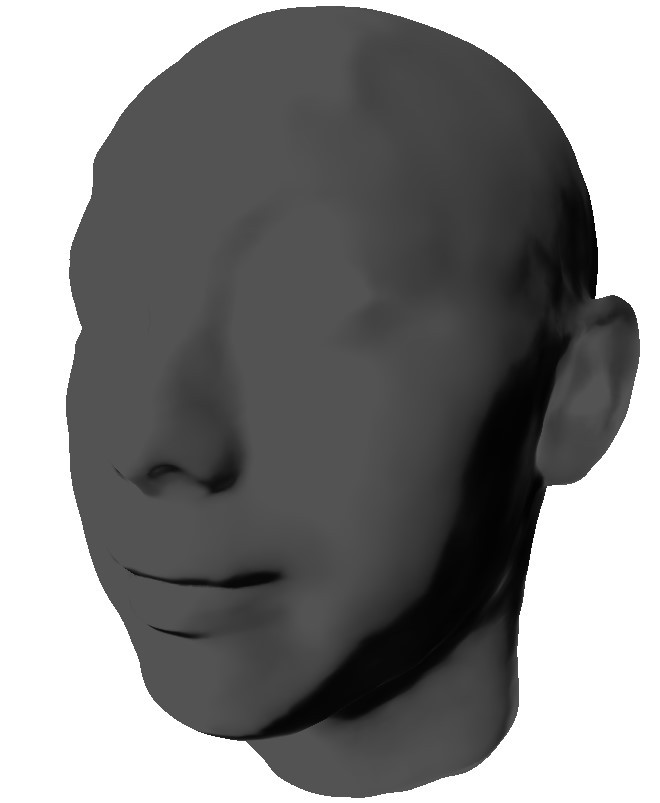}
    \includegraphics[width=0.23\linewidth]{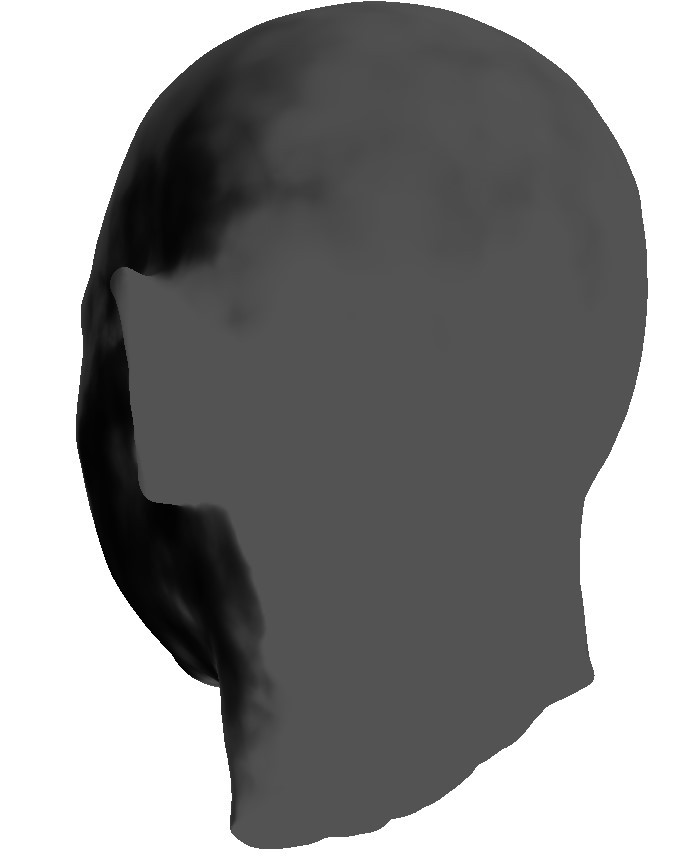}
    \includegraphics[width=0.23\linewidth]{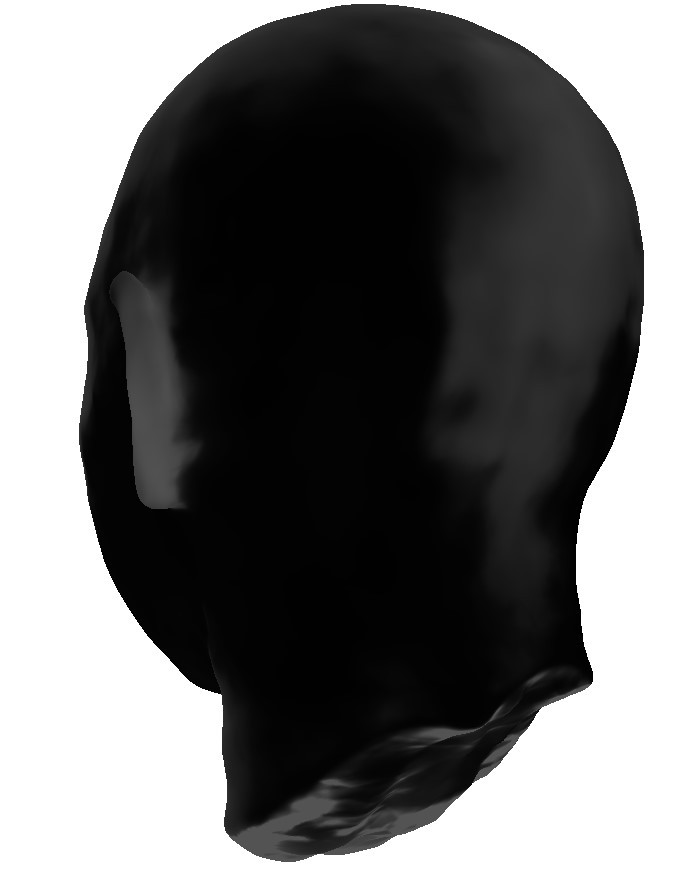} 
    \\
    \makebox[0.23\linewidth]{$\lambda_\text{shd}=0.1$}
    \makebox[0.23\linewidth]{
    $0.01$}
    \makebox[0.23\linewidth]{ 
    $0.01$}
    \makebox[0.23\linewidth]{$0$}
    \end{minipage}
    \begin{minipage}{0.35\linewidth}
    \includegraphics[width=0.95\linewidth]{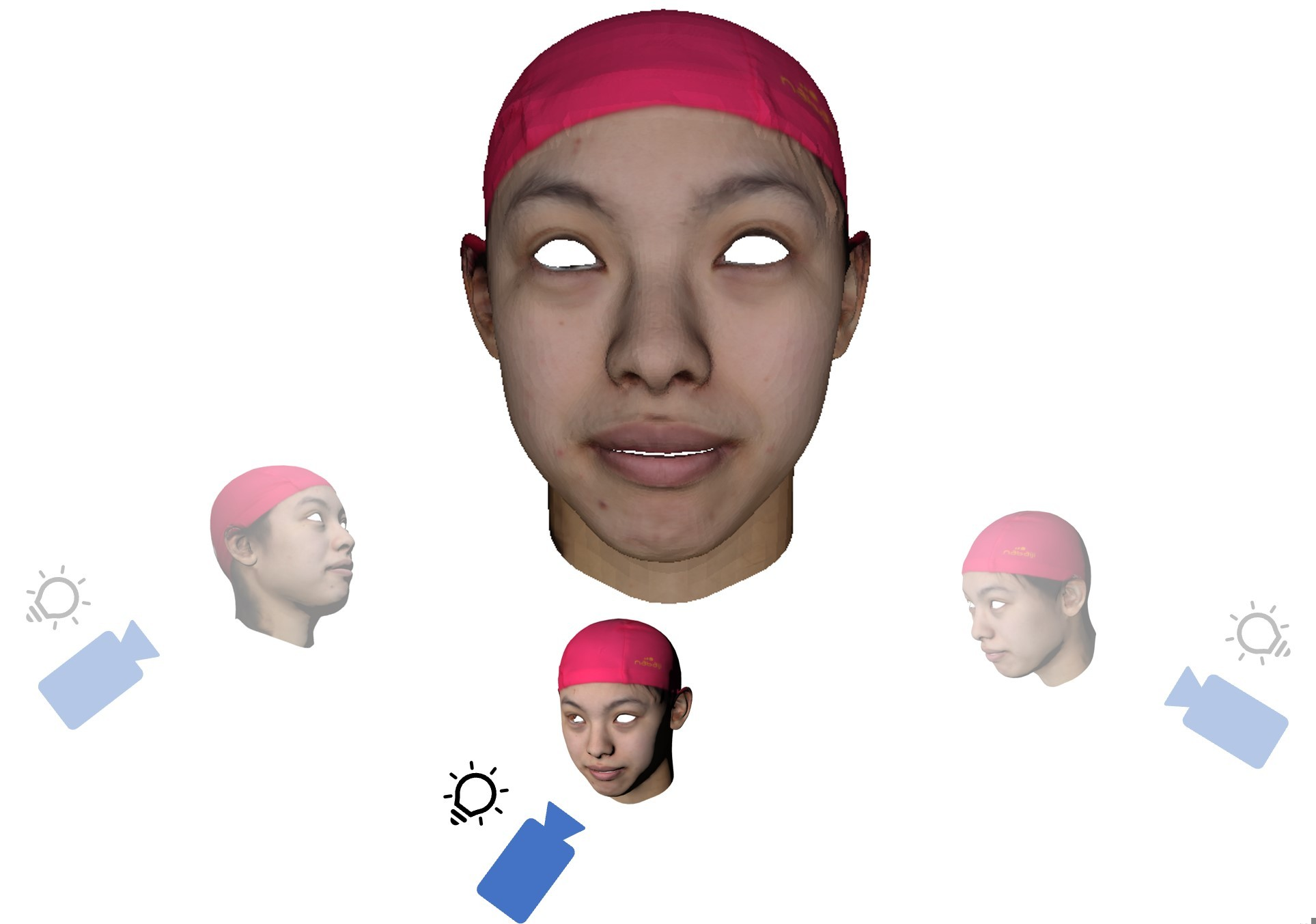}
    \end{minipage}
    \caption{Ablation studies on the shading module demonstrate how the learning of lighting varies with different values of $\lambda_\text{shd}$ (left). We created the FS-Syn dataset from the FaceScape dataset~\cite{yang2020facescape} by illuminating human heads while moving both the light source and the camera, thereby producing varying lighting effects (right).}
    \label{fig:ablation_shading_and_loss}
\end{figure}

\paragraph{Ablation studies on shading module.} 
As shown in Fig.~\ref{fig:ablation_shading_and_loss} (left), a larger value ($\lambda_\text{shd}=0.1$) leads to broken and inconsistent shading. Conversely, omitting $\mathcal{L}_\text{shd}$ results in the shading module learning material intensity that is not caused by the lighting.

\paragraph{FS-Syn dataset preparation.} We derived a customized dataset from Facescape~\cite{yang2020facescape}, as shown in Fig.~\ref{fig:ablation_shading_and_loss}.
We randomly chose 10 human heads with neutral expressions from uniformly topologized models in the FaceScape~\cite{yang2020facescape} dataset. Subsequently, we illuminated the human head model by moving both the light source and the camera, generating diverse lighting effects. Following this, we captured 30 images from different angles around the human head, each accompanied by its corresponding camera parameters.

\section{Geometry Editing via Modifying the Parametric Domain \texorpdfstring{$\mathcal{D}$}{} }
\label{appd:param-edit-results}

\begin{figure}[htbp]
    \centering
    \includegraphics[width=0.975\linewidth]{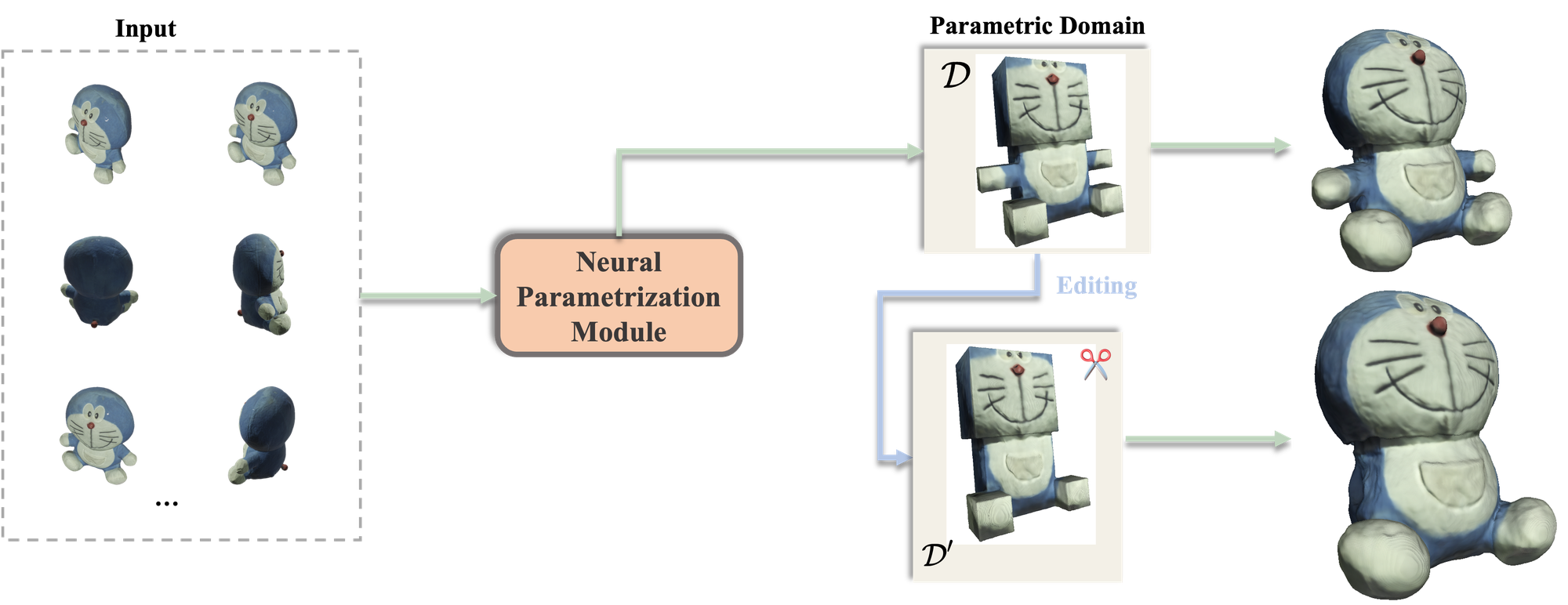}
    \caption{Pipeline of geometry editing on the parametric domain. We replace the original parametric domain with a modified one and immediately render the edited objects.}
    \label{fig:param_edit_pipeline}
\end{figure}

\begin{figure}[htbp]
    \centering
    \includegraphics[width=0.15\linewidth]{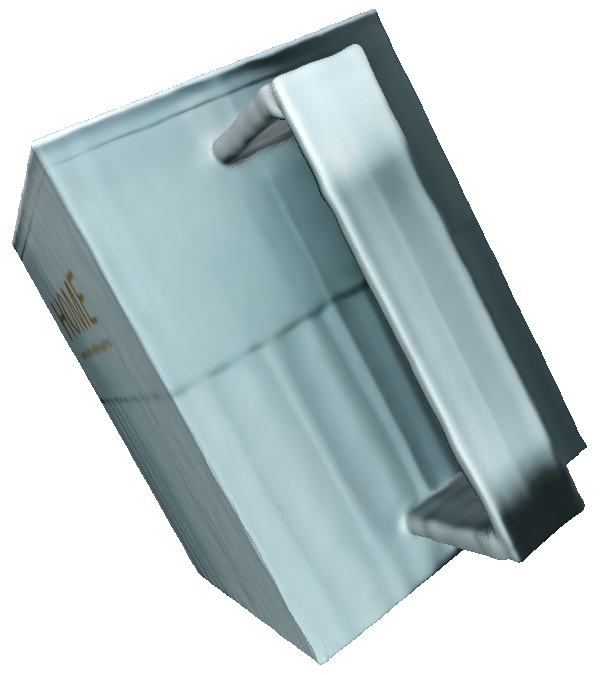}
    \includegraphics[width=0.15\linewidth]{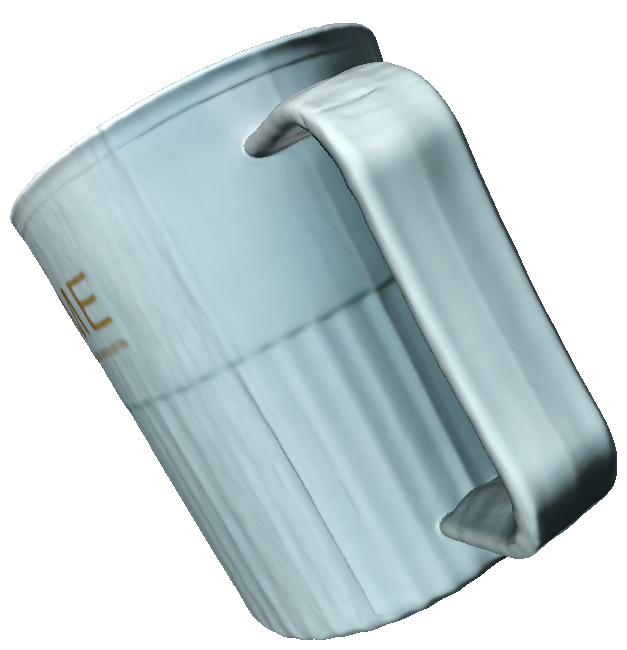}
    \includegraphics[width=0.15\linewidth]{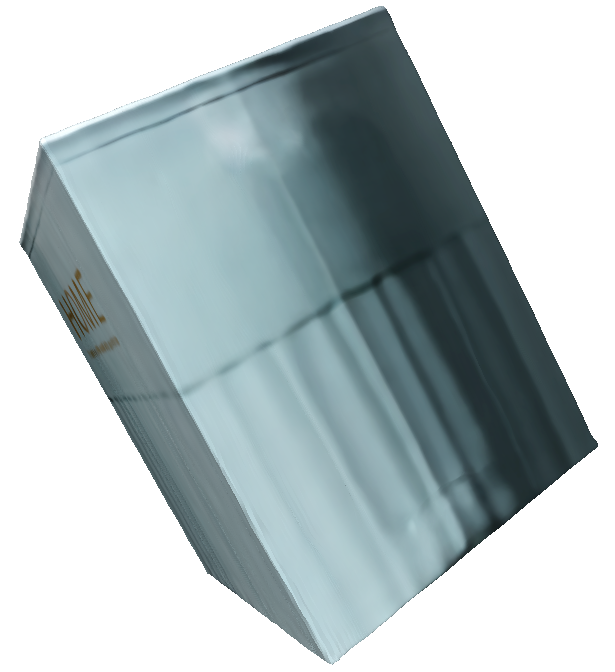}
    \includegraphics[width=0.15\linewidth]{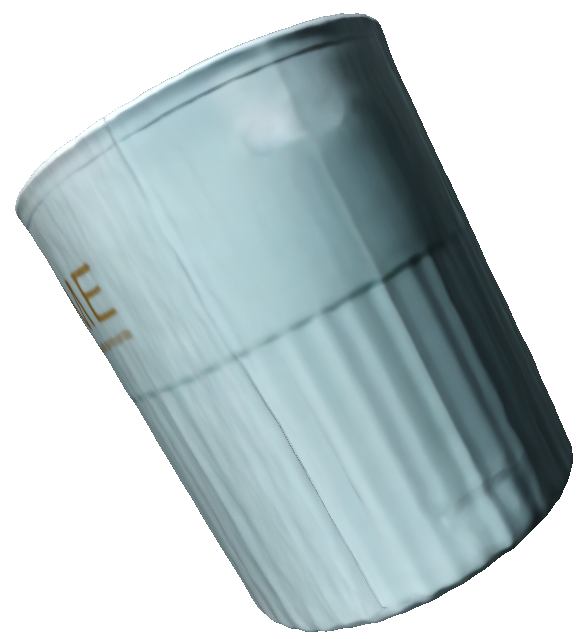}
    \includegraphics[width=0.15\linewidth]{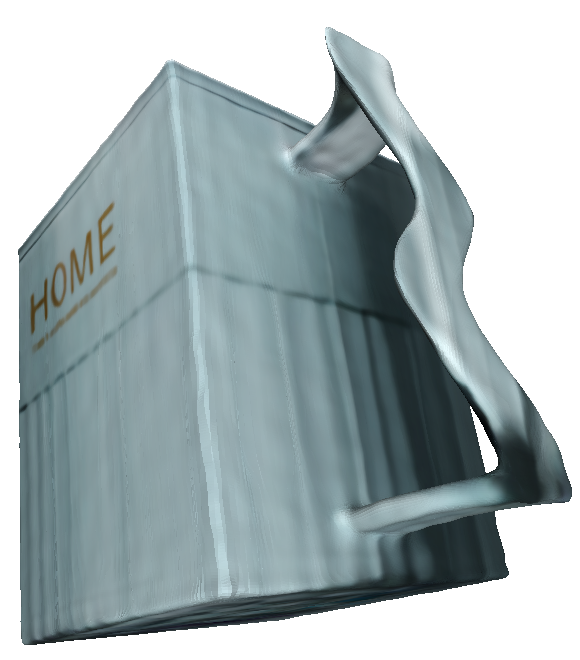}
    \includegraphics[width=0.15\linewidth]{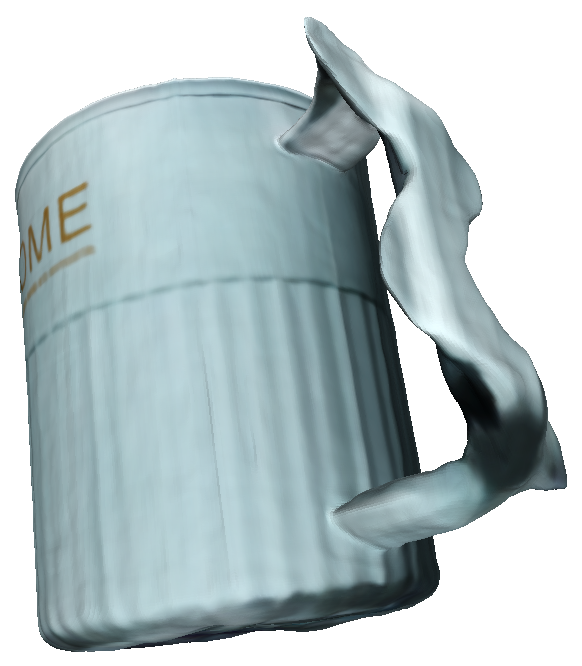}
    \\
    \makebox[0.15\textwidth]{\small{$\mathcal{D}$}}
    \makebox[0.15\textwidth]{\small{$\mathcal{S}$}}
    \makebox[0.15\textwidth]{\small{$\mathcal{D}'$}}
    \makebox[0.15\textwidth]{\small{$\mathcal{S}'$}}
    \makebox[0.15\textwidth]{\small{$\mathcal{D}'$}}
    \makebox[0.15\textwidth]{\small{$\mathcal{S}'$}}
    \caption{Editing the geometry of the parametric surface $\mathcal{D}$ naturally induces a modified neural implicit surface. We edit the geometry of the cup by removing and deforming the handle in $\mathcal{D}$ using Free Form Deformation~\cite{sederberg1986free} in the Blender~\cite{blender}. 
    The first two columns show the original parametric surface $\mathcal{D}$ and reconstructed surface $\mathcal{S}$. Then we give two modified parametric surfaces $\mathcal{D}'$ and the neural implicit surface $\mathcal{S}$ corresponded to each parametric surface. }
    \label{fig:param_edit}
\end{figure}

We also provide a geometry-editing method (See Fig.~\ref{fig:param_edit_pipeline}) based on the parametric domain. As shown in Fig.~\ref{fig:param_edit}, we remove the handle of the cup in the parametric domain, and then we render the cup model in volume rendering without any training again. The parametric shape $\mathcal{D}$ is manually created in the Blender~\cite{blender} and adjusted using techniques like cutting and the Free Form Deformation. We then fit the deformed parametric surface $\mathcal{D}'$
and replace the parametric domain SDF with $\mathcal{D}'$
, and then infer the deformed source geometry $\mathcal{S}'$ and render novel images in volume rendering.

\section{Qualitative Comparison}
We summarize some parametrization methods (refer to Tab.~\ref{tab:comp_method})
to illustrate the effectiveness and superiority of our approach.
As shown in Tab.~\ref{tab:comp_editing}, we categorize these texture editing methods with other editing methods into three groups: scene-level, object-level, and pixel-level editing. Scene-level editing methods focus on the entire appearance of a scene including lighting and material. Object-level editing methods employ distinct part-based latent codes to manipulate different attributes of a scene, such as hairstyles in i3DMM~\cite{yenamandra2021i3dmm}.
Pixel-level editing, on the other hand, offers a fine-grained editing result by taking into account precise user guidance.
\begin{figure}[htbp]
\begin{minipage}{0.9\linewidth}

\resizebox{0.95\linewidth}{!}{\begin{tabular}{l|c|c|c|c|c|c}
\hline
\textbf{Method}  & \textbf{Representation} & \textbf{Prior} & \textbf{Rendering} & \textbf{Regularization} & \textbf{Texture} & \textbf{Domain}\\
\hline
\hline
\textbf{NeuTex}~\cite{xiang2021neutex} & Density & Init. UV-inv & V.R. & NA & Entangled & Sphere\\ \hline
\textbf{ISO-UV}~\cite{DasISOUVfield} & SDF & Init. UV & D.R. & Jacobian & Entangled & $\mathbb{R}^2$\\ \hline
\textbf{NeP}~\cite{ma2022neural} & Density & Mesh \& UV & V.R. & Angle & Disentangled & $\mathbb{R}^2$\\
\hline
\textbf{Ours}  & SDF & None & V.R. & Laplace & Disentangled & Sphere \& polycube\\
\hline
\end{tabular}}
\captionof{table}{Qualitative comparison with other neural parameterization methods. V.R. and D.R. stand for volume rendering and differentiable rendering, respectively. }
\label{tab:comp_method}
\end{minipage}

\begin{minipage}{0.9\linewidth}

\resizebox{0.9\linewidth}{!}{\begin{tabular}{l|c|c|c}
\hline
\textbf{Method}  & \textbf{Level} & \textbf{Retraining} & \textbf{ 3D input } \\
\hline
\hline
\textbf{EditNeRF}~\cite{liu2021editing} & Object & Y & N\\ \hline
\textbf{StylizedNeRF}~\cite{Huang22StylizedNeRF} & Scene & N & N\\ \hline
\textbf{i3DMM}~\cite{yenamandra2021i3dmm} & Object & N & Y\\ \hline
\textbf{NeuMesh}~\cite{yang2022neumesh} & Pixel & Y & N\\ \hline
\textbf{Seal3D}~\cite{wang2023seal3d} & Pixel & Y & N\\ \hline
\textbf{NeuTex}~\cite{xiang2021neutex} & Pixel & N & Y\\ \hline
\textbf{NeP}~\cite{ma2022neural} & Pixel & N & Y \\
\hline
\textbf{Ours}  & Pixel+Object & N & N \\
\hline
\end{tabular}}
\captionof{table}{Qualitative comparison with other neural texture editing methods. }
\label{tab:comp_editing}
\end{minipage}

\end{figure}

\end{document}